\documentclass[11pt]{article}
\usepackage{lineno}

\usepackage[utf8]{inputenc}
\usepackage[T1]{fontenc}
\usepackage[margin=0.8in]{geometry}
\usepackage{amsmath,bm}
\usepackage{amssymb}
\usepackage{amsthm,amsfonts}
\usepackage{booktabs}
\usepackage{nicefrac,microtype}
\usepackage{tikz}
\usepackage{cancel}
\usepackage{wasysym}
\usepackage{graphicx}
\usepackage{float}
\usepackage{mathtools}
\usepackage{commath}
\usepackage{float}
\usepackage{enumitem}
\usepackage{physics}
\usepackage{braket}
\usepackage{hyperref,url}
\usepackage{pdfpages}
\usepackage{subfigure}
\usepackage[skins,breakable]{tcolorbox}
\usepackage{dcolumn}

\usepackage{pdfpages}

\DeclareMathOperator*{\argmin}{arg\,min}

\def\bPhi{\mathbf{\Phi}}

\def\bLambda{{\bm \Lambda}}
\def\bLambda{{\bf{H}}}
\def\bLambda{{\bm \Lambda}}

\def\1{{\bf 1}}

\def\w{{\mathbf w}}
\def\x{{\mathbf x}}
\def\y{{\mathbf y}}

\def\I{{\bf I}}

\def\K{{\mathbf K}}

\def\Y{{\mathbf Y}}

\usepackage{titling}
\thanksmarkseries{arabic}

\usepackage[symbol]{footmisc}
\renewcommand{\thefootnote}{\fnsymbol{footnote}}

\title{Spectral Bias and Task-Model Alignment Explain Generalization in Kernel Regression and Infinitely Wide Neural Networks}
\author{
  Abdulkadir Canatar $^{1,2}$ %
  \and
  Blake Bordelon $^{2,3}$%
  \and
  Cengiz Pehlevan $^{2,3,*}$}

\date{}

\begin{document}
\maketitle

\renewcommand*{\thefootnote}{\arabic{footnote}}
\footnotetext[1]{Department of Physics, Harvard University, Cambridge, MA 02138, USA.}
\footnotetext[2]{Center for Brain Science, Harvard University, Cambridge, MA 02138, USA.}
\footnotetext[3]{John A. Paulson  School  of  Engineering  and  Applied  Sciences, Harvard University, Cambridge, MA 02138, USA.}
\renewcommand*{\thefootnote}{\fnsymbol{footnote}}
\footnotetext[1]{Corresponding Author, E-mail: cpehlevan@seas.harvard.edu}

\renewcommand{\figurename}{Fig.}
\renewcommand{\figref}[2][]{{{\figurename\,\ref{#2}#1}}}

\begin{abstract}
A theoretical understanding of generalization remains an open problem for many machine learning models, including deep networks where overparameterization leads to better performance, contradicting the conventional wisdom from classical statistics. Here, we investigate generalization error for kernel regression, which, besides being a popular machine learning method, also describes certain infinitely overparameterized neural networks. We use techniques from statistical mechanics to derive an analytical expression for generalization error applicable to any kernel and data distribution. We present applications of our theory to real and synthetic datasets, and for many kernels including those that arise from training deep networks in the infinite-width limit. We elucidate an inductive bias of kernel regression to explain data with simple functions, characterize whether a kernel is compatible with a learning task, and show that more data may impair generalization when noisy or not expressible by the kernel, leading to non-monotonic learning curves with possibly many peaks. 
\end{abstract}

\newpage

\section*{Introduction}

Learning machines aim to find statistical patterns in data that generalize to previously unseen samples \cite{hastie2009elements}. How well they perform in doing so depends on factors such as the size and the nature of the training data set, the complexity of the learning task, and the inductive bias of the learning machine. Identifying precisely how these factors contribute to the generalization performance has been a theoretical challenge. In particular, a definitive theory should be able to predict generalization performance on real data. Existing theories fall short of this goal, often providing impractical bounds and inaccurate estimates \cite{zhang2016understanding,belkin2019reconciling}. 

The need for a new theory of generalization is exacerbated by recent developments in deep learning \cite{lecun2015deep}.
Experience in the field suggests that larger models perform better \cite{nakkiran2019deep,kaplan2020scaling,lepikhin2020gshard}, encouraging training of larger and larger networks with state-of-the-art architectures reaching hundreds of billions of parameters \cite{lepikhin2020gshard}. These networks work in an overparameterized regime \cite{belkin2019reconciling,nakkiran2019deep} with much more parameters than training samples, and are highly expressive to a level that they can even fit random noise \cite{zhang2016understanding}. Yet, they generalize well, contradicting the conventional wisdom from classical statistical learning theory \cite{hastie2009elements,belkin2019reconciling,Vapnik1998} according to which  overparameterization should lead to overfitting and  worse generalization. It must be that overparameterized networks have inductive biases that suit the learning task. Therefore, it is crucial for a theory of generalization to elucidate such biases.

While addressing the full complexity of deep learning is as of now beyond the reach of theoretical study, a tractable, yet practically-relevant limit was established by recent work pointing to a correspondence between training deep networks and performing regression with various rotation invariant kernels. In the limit where the width of a network is taken to infinity (network is thus overparameterized), neural network training with a certain random initialization scheme can be described by ridgeless kernel regression with the Neural Network Gaussian Process kernel (NNGPK) if only the last layer is trained \cite{neal1996bayesian,cho2009kernel,matthews2018gaussian,lee2017deep}, or the Neural Tangent Kernel (NTK) if all the layers are trained  \cite{jacot2018neural}. Consequently, studying the inductive biases of kernels arising from the infinite-width limit should give insight to the success of overparameterized neural networks. Indeed, key generalization phenomena in deep learning also occur in kernel methods, and it has been argued that understanding generalization in kernel methods is necessary for understanding generalization in deep learning \cite{belkin2018understand}.

Motivated by these connections to deep networks and also by its wide use, in this paper, we present a theory of generalization in kernel regression \cite{wahba1990spline,evgeniou2000regularization,shawe2004kernel,mohri2018foundations,jacot2020kernel}. Our theory is generally applicable to any kernel and contains the infinite-width limit of deep networks as a special case. Most importantly, our theory is applicable to real datasets.

We describe typical generalization performance of kernel regression shedding light onto practical uses of the algorithm, in contrast to the worst case bounds of statistical learning theory \cite{Vapnik1998,mohri2018foundations,cucker2002best,caponnetto2007optimal,liang2018just}. In the past, statistical mechanics provided a useful theoretical framework for such typical-case analyses for various algorithms \cite{gardner1988space,Hertz_1989, krogh, sompolinsky1992examples,optimalperceptron,statmechofrule,sollich1999learning,Malzahn_Opper,engel2001statistical,bahri2019statistical}. Here, using the replica method of statistical mechanics \cite{mezard1987spin}, we derive an analytical expression for the typical generalization error of kernel regression as a function of 1) the number of training samples, 2) the eigenvalues and eigenfunctions of the kernel, which define the inductive bias of kernel regression, and 3) the alignment of the target function with the kernel's eigenfunctions, which provides a notion of how compatible the kernel is for the task. We test our theory on various real datasets and kernels. Our analytical generalization error predictions fit experiments remarkably well.

Our theory sheds light onto the various generalization phenomena. We elucidate a strong inductive bias: as the size of the training set grows, kernel regression fits successively higher spectral modes of the target function, where the spectrum is defined by solving an eigenfunction problem \cite{bordelon2020spectrum, jacot2020implicit,jacot2020kernel, liang_isometry}. Consequently, our theory can predict which kernels or neural architectures are well suited to a given task by studying the alignment of top kernel eigenfunctions with the target function for the task. Target functions that place most power in the top kernel eigenfunctions can be estimated accurately at small sample sizes, leading to good generalization. Finally, when the data labels are noisy or the target function has components not expressible by the kernel, we observe that generalization error can exhibit non-monotonic behavior as a function of the number of samples, contrary to the common intuition that more data should lead to smaller error. This non-monotonic behavior is reminiscent of the recently described ``double-descent'' phenomenon \cite{belkin2019reconciling,nakkiran2019deep,loog2020brief,montanari2019generalization}, where generalization error is non-monotonic as a function of model complexity in many modern machine learning models.  We show that the non-monotonicity can be mitigated by increasing the implicit or explicit regularization.

To understand these phenomena better, we present a detailed analytical study of the application of our theory to rotation invariant kernels, motivated by their wide use and relevance for deep learning. Besides NNGPK and NTK, this class includes many other popular kernels such as the Gaussian, Exponential and Matern kernels \cite{genton2001classes,RasmussenWilliams}. When the data generating distribution is also spherically symmetric, our theory is amenable to further analytical treatment. Our analyses provide a mechanistic understanding of the inductive bias of kernel regression and the possible non-monotonic behavior of learning curves.

\section*{Results}
\subsection*{Generalization Error of Kernel Regression from Statistical Mechanics}

Kernel regression is a supervised learning problem where one estimates a function from a number of observations. For our setup, let $\mathcal{D} = \{\mathbf{x}^\mu, y^\mu\}_{\mu=1}^P$ be a sample of $P$ observations drawn from a probability distribution on $\mathcal{X} \times \mathbb{R}$, and $\mathcal X \subseteq \mathbb{R}^D$. The inputs $\x^{\mu}$ are drawn from a distribution $p(\x)$, and the labels $y^\mu$ are assumed to be generated by a noisy target $ y^\mu = \bar f(\mathbf{x}^\mu) + \epsilon^\mu$, where $\bar f$ is square integrable with respect to $p(\x)$, and $\epsilon^\mu$ represents zero-mean additive noise with covariance $\left< \epsilon^\mu \epsilon^\nu \right> = \delta_{\mu \nu} \sigma^2$. The kernel regression problem is
\begin{linenomath*}\begin{equation}\label{eq:regression}
    f^* = \argmin_{f \in \mathcal{H}} \frac{1}{2\lambda } \sum_{i=1}^P ( f(\mathbf{x}^\mu) - y^\mu)^2 + \frac{1}{2}\left< f,f \right>_{\mathcal{H}},
\end{equation}\end{linenomath*}
where $\lambda$ is the ``ridge'' parameter, $\mathcal{H}$ is a Reproducing Kernel Hilbert Space (RKHS) uniquely determined by its reproducing kernel $K(\mathbf{x},\mathbf{x}')$ and the input distribution $p(\x)$ \cite{scholkopf2002learning},  and  $\left< \cdot, \cdot \right>_{\mathcal{H}}$ is the RKHS inner product. The Hilbert norm penalty controls the complexity of $f$. The $\lambda\to 0$ limit is referred to as the kernel interpolation limit, where the dataset is exactly fit: $f^* = \argmin_{f \in \mathcal{H}} \left< f,f \right>_{\mathcal{H}},\,{\rm s.t.}\, f(\mathbf{x}^\mu) = y^\mu, \mu = 1,\ldots P$. We emphasize that in our setting the target function does not have to be in the RKHS. 

Our goal is to calculate generalization error, i.e. mean squared error between the estimator, $f^*$, and the ground-truth (target) $\bar f(\mathbf{x})$ averaged over the data distribution and datasets:
\begin{linenomath*}\begin{equation}\label{eq:genError}
    E_g = \left< \int d\x \, p(\x)\,\big(f^*(\x) - \bar f(\mathbf{x})\big)^2 \right>_\mathcal{D}.
\end{equation}\end{linenomath*}
$E_g$ measures, on average, how well the function learned agrees with the target on previously unseen (and seen) data sampled from the same distribution. 

This problem can be analyzed using the replica method from statistical physics of disordered systems \cite{mezard1987spin}, treating the training set as a quenched disorder. Our calculation is outlined in Methods and further detailed in the Supplementary Information. Here we present our main results.

Our results rely on the Mercer decomposition of the kernel in terms of orthogonal eigenfunctions $\{\phi_\rho\}$, 
%
\begin{linenomath*}\begin{equation}\label{eq:eigenvalue}
    \int  d \mathbf{x}'\,p(\mathbf{x}') K(\mathbf{x}, \mathbf{x}') \phi_\rho(\mathbf{x}')  = \eta_\rho \phi_\rho(\mathbf{x}), \qquad \rho = 1,\ldots, N,
\end{equation}\end{linenomath*}
which form a complete basis for the RKHS, and eigenvalues $\{\eta_\rho\}$. $N$ is typically infinite. For ease of presentation, we assume that all eigenvalues are strictly greater than zero. In Supplementary Note 1 and 2, we fully address the case with zero eigenvalues. Working with the orthogonal basis set $\psi_\rho(\mathbf{x}) \equiv \sqrt{\eta_\rho} \phi_\rho(\mathbf{x})$,
 also called a feature map, we introduce coefficients $\{\overline{w}_\rho\}$ and $\{w_\rho^*\}$ that represent the target and learned functions respectively $\bar f(\mathbf{x}) = \sum_\rho \overline{w}_\rho \psi_\rho(\mathbf{x})$, and $f^*(\mathbf{x}) = \sum_\rho w_\rho^* \psi_\rho(\mathbf{x})$.
    
With this setup, we calculate the generalization error of kernel regression for any kernel and data distribution to be (Methods and Supplementary Note 2):    %
\begin{linenomath*}\begin{equation}\label{eq:finitePgenErr}
\begin{gathered}
    E_g = \frac{1}{1-\gamma}\sum_{\rho}\frac{\eta_\rho }{\big(\kappa+P\eta_\rho\big)^2}\big(\kappa^2\bar w_\rho^2+\tilde{\sigma}^2 P\eta_\rho\big), \\
    \kappa = \lambda  + \sum_\rho \frac{\kappa\eta_\rho}{\kappa+{P}\eta_\rho},\quad
    \gamma = \sum_\rho \frac{P\eta_\rho^2}{(\kappa+P\eta_\rho)^2}.
\end{gathered}
\end{equation}\end{linenomath*}
We note that the generalization error is the sum of a $\sigma$-independent term and a $\sigma$-dependent term, the latter of which fully captures the effect of noise on generalization error.

Formally, this equation describes the typical behavior of kernel regression in a thermodynamic limit that involves taking $P$ to infinity. In this limit, variations in kernel regression's performance due to the differences in how the training set is formed, which is assumed to be a stochastic process, become negligible. The precise nature of the limit depends on the kernel and the data distribution. In this work, we consider two different analytically solvable cases and identify natural scalings of $N$ and $D$ with $P$, which in turn govern how the kernel eigenvalues $\eta_\rho$ scale inversely with $P$. We further give the infinite-$P$ limits of Eq. \eqref{eq:finitePgenErr} explicitly for these cases. In practice, however, we find that our generalization error formula describes average learning curves very well for finite $P$ for even as low as a few samples. We observe that the variance in learning curves due to stochastic sampling of the training set is significant for low $P$, but decays with increasing $P$ as expected.

We will demonstrate various generalization phenomena that arises from Eq. \eqref{eq:finitePgenErr} through simulations and analytical study. One immediate observation is the \textit{spectral bias}: faster rates of convergence of the error along eigenfunctions corresponding to higher eigenvalues in the noise free ($\sigma^2 = 0$) limit. The generalization error can be decomposed into a sum of modal errors $ E_g = \sum_\rho \eta_\rho \overline{w}_{\rho}^2 E_\rho$, where each normalized mode error $E_\rho$ represents the contribution of the mode error due to estimation of the coefficient for eigenfunction $\psi_\rho$ (Methods). The normalized mode errors are ordered according to their eigenvalues (Methods)
\begin{linenomath*}\begin{equation}
    \eta_\rho > \eta_{\rho'} \implies E_{\rho} < E_{\rho'},
\end{equation}\end{linenomath*}
which implies that modes $\rho$ with large eigenvalues $\eta_\rho$ are learned more rapidly as $P$ increases than modes with small eigenvalues. 

An important implication of this result is that target functions acting on the same data distribution with higher \textit{cumulative power distributions} $C(\rho)$, defined as the proportion of target power in the first $\rho$ modes
\begin{linenomath*}
\begin{equation}\label{eq:Ck}
    C(\rho) = \frac{\sum_{\rho' \leq \rho} \eta_{\rho'} \overline{w}_{\rho'}^2}{\sum_{\rho'} \eta_{\rho'} \overline{w}_{\rho'}^2},
\end{equation}
\end{linenomath*}
for all $k\geq 1$ will have lower generalization error normalized by total target power, $E_g(P)/E_g(0)$, for all $P$ (Methods). Therefore, $C(\rho)$ provides a measure of the compatibility between the kernel and the target, which we name {\it task-model alignment}. 

We further note that the target function enters normalized generalization error only through combinations $C(\rho)-C(\rho-1) = \eta_{\rho} \overline{w}_{\rho}^2/\sum_{\rho} \eta_\rho \overline{w}_\rho^2$. Hence, the kernel eigenvalues, the cumulative power distribution, and the noise parameter completely specify the normalized generalization error. Spectral bias, task-model alignment and noise explain generalization in kernel regression.

Generalization error can exhibit non-monotonicity which can be understood through the bias and variance decomposition \cite{montanari2019generalization,dascoli2020double,dascoli2020triple}, $E_g = B+V$, where  $B = \int d\x\, p(\x) \left(\left<f^*(\x) \right>_{\mathcal{D}} - \bar f(\x) \right)^2$ and $V = \left<\int d\x \, p(\x) \left(f^*(\x) - \left<f^*(\x) \right>_{\mathcal{D}} \right)^2 \right>_{\mathcal D}$. We found that the average estimator is given by 
 $\braket{f^*(\x;P)}_\mathcal{D}= \sum_\rho \frac{P\eta_\rho}{P\eta_\rho + \kappa}\bar w_\rho \psi_\rho(\x)$, which
 monotonically approaches the target function as $P$ increases, giving rise to a monotonically decreasing bias (Supplementary Note 2). However, the variance term arising from the variance of the estimator over possible sampled datasets $\mathcal{D}$ is potentially non-monotonic as the dataset size increases. Therefore, the total generalization error can exhibit local maxima.

\subsection*{Applications to Real Datasets}

Next, we evaluate our theory on realistic datasets and show that it predicts kernel regression learning curves with remarkable accuracy. We further elucidate various heuristic generalization principles.

To apply our theory, we numerically solve the eigenvalue problem Eq. \eqref{eq:eigenvalue} on the dataset (Methods) and obtain the necessary eigenvalues and eigenfunctions. When solved on a finite dataset, Eq. \eqref{eq:eigenvalue} is an uncentered kernel PCA problem (Methods).
We use these eigenfunctions (or eigenvectors for finite data) to express our target function, and the resulting coefficients and kernel eigenvalues to evaluate the generalization error.

\begin{figure}[h]
    \centering
    \includegraphics[scale = 1]{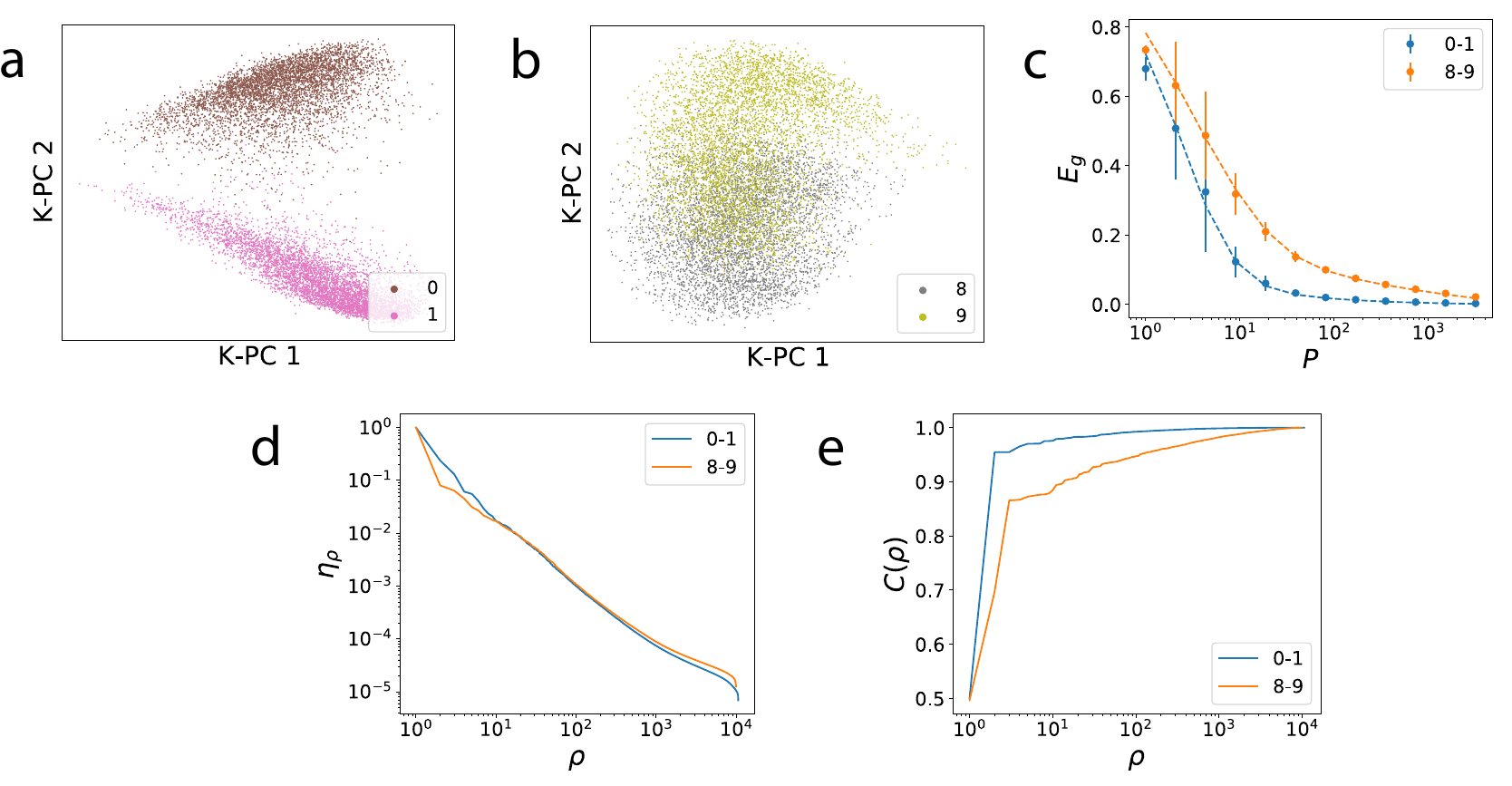}
    \caption{\textbf{Effect of task-model alignment on the generalization of kernel regression.} \textbf{a, b.} Projections of digits from MNIST along the top two (uncentered) kernel principal components of 2-layer NTK for 0s vs. 1s and 8s vs. 9s, respectively. \textbf{c.} Learning curves for both tasks. The theoretical learning curves (Eq. \eqref{eq:finitePgenErr}, dashed lines) show strong agreement with experiment (dots). \textbf{d.} The kernel eigenspectra for the respective datasets. \textbf{e.} The cumulative power distributions $C(\rho)$. Error bars show the standard deviation over $50$ trials.}\label{fig:easy_vs_hard_visualization}
\end{figure}

In our first experiment, we test our theory using a 2-layer NTK \cite{cho2009kernel,jacot2018neural} on two different tasks: discriminating between 0s and 1s, and between 8s and 9s from MNIST dataset \cite{MNIST}. We formulate each of these tasks as a kernel regression problem by considering a vector target function which takes in digits and outputs one-hot labels. Our kernel regression theory can be applied separately to each element of the target function vector (Methods), and a generalization error can be calculated by adding the error due to each vector component.

We can visualize the complexity of the two tasks by plotting the projection of the data along the top two kernel principal components (\figref{fig:easy_vs_hard_visualization}a,b). The projection for 0-1 digits appears highly separable compared to 8-9s, and thus simpler to learn to discriminate. 
Indeed, the generalization error for the 0-1 discrimination task falls more rapidly than the error for the 8-9 task (\figref{fig:easy_vs_hard_visualization}c). Our theory is in remarkable agreement with experiments. Why is 0-1 discrimination easier for this kernel? \figref{fig:easy_vs_hard_visualization}d shows that the eigenvalues of the NTK evaluated on the data are very similar for both datasets. To quantify the compatibility of the kernel with the tasks, we measure the cumulative power distribution $C(\rho)$. Even though in this case the data distributions are different, $C(\rho)$ is still informative. \figref{fig:easy_vs_hard_visualization}e illustrates that $C(\rho)$ rises more rapidly for the easier 0-1 task and more slowly for the 8-9 task, providing a heuristic explanation of why it requires a greater number of samples to learn.  

\begin{figure}
    \centering
    \includegraphics[scale = 1]{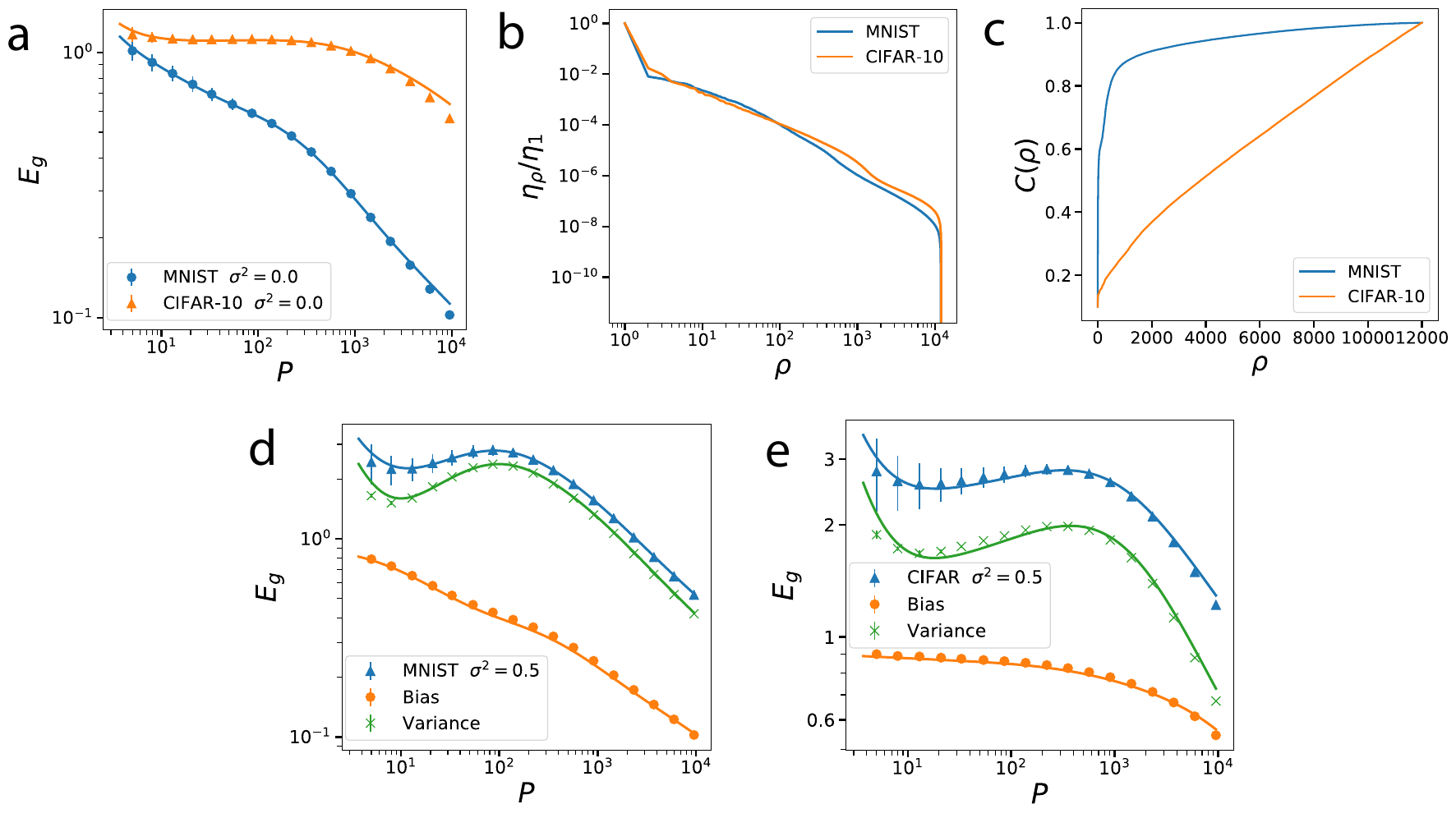}
    \caption{\textbf{Gaussian RBF kernel regression on MNIST and CIFAR-10 datasets.} Kernel is $K(\mathbf{x},\mathbf{x}')=e^{-\frac{1}{2 D \omega^2} ||x-x'||^2}$  with kernel bandwidth $\omega = 0.1$, ridge parameter $\lambda = 0.01$ and $D$ being the size of images. \textbf{a.} Generalization error $E_g(p)$ when $\sigma^2=0$: Solid lines are theory (Eq. \eqref{eq:finitePgenErr}), dots are experiments. \textbf{b.} Kernel eigenvalues and \textbf{c.} cumulative powers $C(\rho)$ for MNIST and CIFAR-10. \textbf{d, e.} Generalization error when $\sigma^2=0.5$ with its bias-variance decomposition for MNIST and CIFAR-10 datasets, respectively. Solid lines are theory, markers are experiments. Error bars represent standard deviation over $160$ trials. Bias and variance are obtained by calculating the mean and variance of the estimator over $150$ trials, respectively.} \label{fig:DoubleDescent_MNISTCIFAR}
\end{figure}

We next test our theory for Gaussian RBF kernel on the MNIST \cite{MNIST} and CIFAR \cite{CIFAR} datasets. \figref{fig:DoubleDescent_MNISTCIFAR}a shows excellent agreement between our theory and experiments for both. \figref{fig:DoubleDescent_MNISTCIFAR}b shows that the eigenvalues of the Gaussian RBF kernel evaluated on data are similar for MNIST and CIFAR-10. The cumulative powers $C(\rho)$ (\figref{fig:DoubleDescent_MNISTCIFAR}c), however, are very different. Placing more power in the first few modes makes learning faster. When the labels have nonzero noise $\sigma^2>0$ (\figref{fig:DoubleDescent_MNISTCIFAR}d,e), generalization error is non-monotonic with a peak, a feature that has been named ``double-descent
'' \cite{belkin2019reconciling,loog2020brief}. By decomposing $E_g$ into the bias and the variance of the estimator, we see that the non-monotonicity is caused solely by the variance (\figref{fig:DoubleDescent_MNISTCIFAR}d,e). Similar observations about variance were made in different contexts before \cite{montanari2019generalization, dascoli2020double,nakkiran2020optimal}.

These experiments and discussion in the previous section provide illustrations of the three main heuristics about how dataset, kernel, target function, and noise interact to produce generalization in kernel regression:
\begin{enumerate}
    \item \textit{Spectral Bias}: Kernel eigenfunctions $\phi_\rho$ with large eigenvalues $\eta_\rho$ can be estimated with kernel regression using a smaller number of samples. 
    \item \textit{Task-Model Alignment}: Target functions with most of their power in top kernel eigenfunctions can be estimated efficiently and are compatible with the chosen kernel. We introduce cumulative power distribution, $C(\rho)$, as defined in Eq. \eqref{eq:Ck}, as a measure off this alignment. 
    \item \textit{Non-monotonicity}: Generalization error may be non-monotonic with dataset size in the presence of noise (as in \figref{fig:DoubleDescent_MNISTCIFAR}), or when the target function is not expressible by the kernel (not in the RKHS). We provide a discussion of and examples for the latter kind in Supplementary Note 3 and 4. We show that modes of the target function corresponding to zero eigenvalues of the kernel act effectively as noise on the learning problem.
\end{enumerate}

To explore these phenomena further and understand their causes, we study several simplified models where the kernel eigenvalue problem and generalization error equations can be solved analytically.


\subsection*{Double-Descent Phase Transition in a Band-Limited RKHS} 

An explicitly solvable and instructive case is the white band-limited RKHS with $N$ equal nonzero eigenvalues, a special case of which is linear regression. Later, we will observe that the mathematical description of rotation invariant kernels on isotropic distributions reduces to this simple model in each learning stage.

In this model, the kernel eigenvalues are equal $\eta_\rho = \frac{1}{N}$ for a finite number of modes $\rho=1,...,N$ and truncate thereafter: $\eta_\rho = 0$ for $\rho>N$. Similarly, the target power $\overline {w}_\rho^2$ truncates after $N$ modes and satisfies the normalization condition $\sum_{\rho=1}^N \overline{w}_\rho^2 = N$. In Supplementary Note 3, we relax these constraints and discuss their implications. Linear regression (or linear perceptron) with isotropic data is a special case when $D=N$, $\phi_{\rho}(\mathbf{x}) = x_{\rho}$, and $\braket{x_{\rho} x_{\rho'}}_{\x\sim p(\x)} = \delta_{\rho\rho'}$ \cite{krogh}.

We study this model in the thermodynamic limit. We find that the natural scaling is to take $P \to \infty$ and $N \to \infty$ with  $\alpha = P/N \sim \mathcal{O}(1)$, and $D \sim O(1)$ (or $D=N\sim\mathcal{O}(P)$ in the linear regression case), leading to the generalization error:
\begin{linenomath*}\begin{equation}\label{Egband}
\begin{gathered}
E_g(\alpha,\lambda,\sigma^2) = \frac{\kappa_\lambda(\alpha)^2+\sigma^2\alpha}{(\kappa_\lambda(\alpha)+\alpha)^2 - \alpha}, \\ 
\kappa_\lambda(\alpha)  = \frac{1}{2} \left[(1+{\lambda}-\alpha) + \sqrt{(1+\lambda-\alpha)^2 + 4\lambda \alpha} \right].
\end{gathered}
\end{equation}\end{linenomath*}
Note that the result is independent of the teacher weights as long as they are properly normalized. The function $\kappa_\lambda(\alpha)$ appears in many contexts relevant to random matrix theory, as it is related to the resolvent, or Stieltjes transform, of a random Wishart matrix \cite{bai_silverstein, Advani_Ganguli_optimal} (Supplementary Note 3). This simple model shows interesting behavior, elucidating the role of regularization and under- vs. over-parameterization in learning machines.

\begin{figure}
    \centering
    \includegraphics[scale = 1]{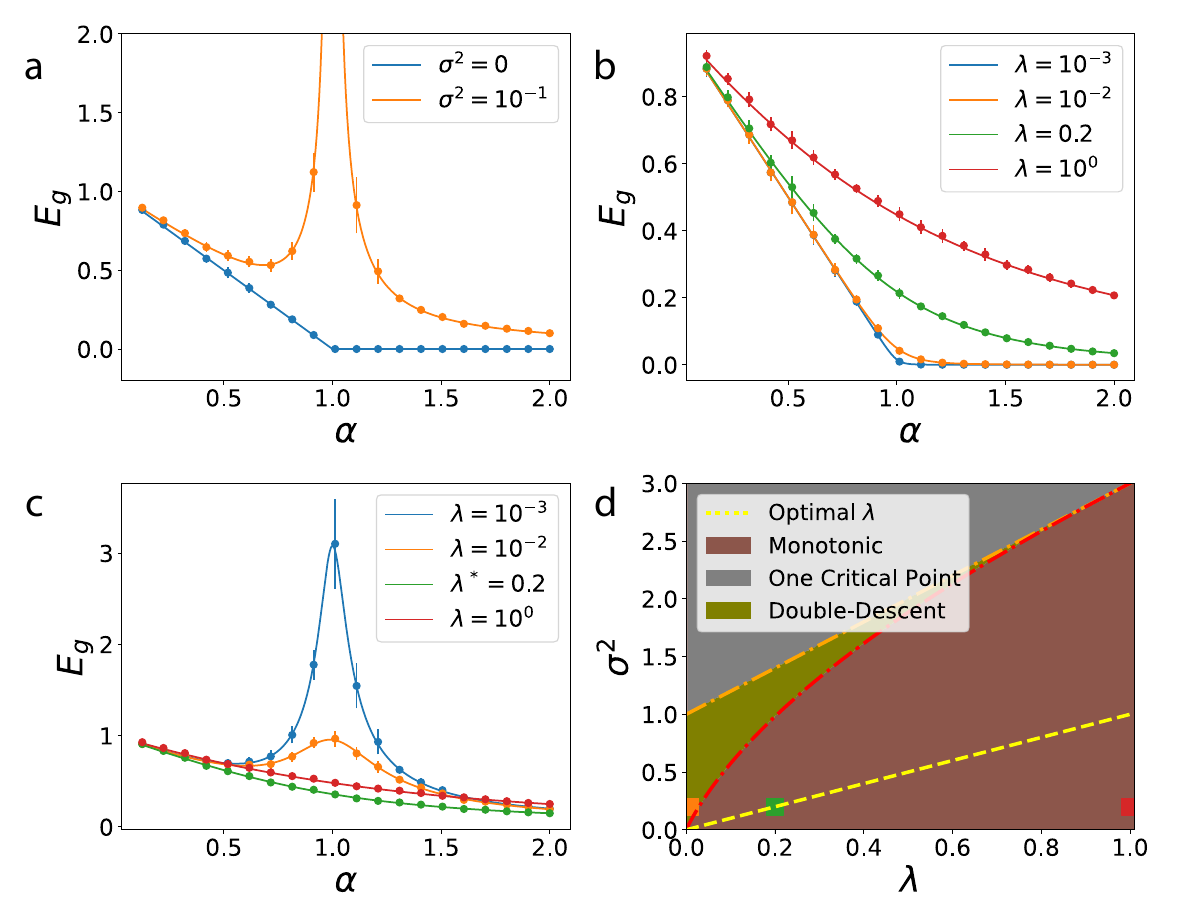}
    \caption{\textbf{Learning curves and double-descent phase diagram for kernels with white band-limited spectra.} We simulated $N=800$ dimensional uncorrelated Gaussian features $\bm\phi(\x) = \x \sim \mathcal{N}(0,\I)$ and estimated a linear function $\bar{f}(\x) = \bm\beta^\top \x$ with $||\bm\beta||^2 = N$. Error bars describe the standard deviation over $15$ trials. Solid lines are theory (Eq. \eqref{Egband}), dots are experiments. \textbf{a.} When $\lambda = 0$ and $\sigma^2=0$, $E_g$ linearly decreases with $\alpha$ and when $\sigma^2> 0$ it diverges as $\alpha \to 1$. \textbf{b.} When $\sigma^2=0$, explicit regularization $\lambda$ always leads to slower decay in $E_g$. \textbf{c.} For nonzero noise $\sigma^2>0$, there is an optimal regularization $\lambda^* = \sigma^2$ which gives the best generalization performance. \textbf{d.} Double-descent phase diagram where the colored squares correspond to the curves with same color in panel \textbf{c}. Optimal regularization ($\lambda^* = \sigma^2$) curve is shown in yellow dashed line which does not intersect the double-descent region above the curve defined by $g(\lambda)$ (Eq. \eqref{eq:phase_boundary_white}).}
    \label{fig:white}
\end{figure}

First we consider the interpolation limit ($\lambda = 0$, \figref{fig:white}a).  The generalization error simplifies to $E_g = (1-\alpha) \Theta(1-\alpha) + \frac{\sigma^2}{1-\alpha} \left[ \alpha \Theta(1-\alpha) -  \Theta(\alpha-1) \right]$. There is a first order phase transition at $\alpha_c = 1$, when the number of samples $P$ is equal to the number of non-zero modes $N$ and therefore to the number of parameters, $\lbrace \bar w_{\rho}\rbrace$, that define the target function. The phase transition is signaled by the non-analytic behavior of $E_g$ and verifiable by computing the first-derivative of free energy (Supplementary Note 3). When $\sigma = 0$, $E_g$ linearly falls with more data and at the critical point generalization error goes to zero. With noise present, the behavior at the critical point changes drastically, and there is a singular peak in the generalization error due to the noise term of the generalization error (\figref{fig:white}a). At this point the kernel machine is (over-)fitting exactly all data points, including noise. Then, as number of samples increase beyond the number of parameters ($\alpha > 1$), the machine is able to average over noise and the generalization error falls with asymptotic behavior $E_g \sim \sigma^2/\alpha$.  Our results are consistent with those previously obtained for the linear perceptron with a noisy target \cite{krogh,sollich1994finite}, which is a special case of kernel regression with a white band-limited spectrum.

When $\lambda > 0$ and $\sigma = 0$, $E_g$ decreases monotonically with $\alpha$ and is asymptotic to $E_g \sim {{\lambda}^2}/{\alpha^2}$ (\figref{fig:white}b). A sharp change at $\alpha =1$ is visible for small $\lambda$, reminiscent of the phase transition at $\lambda = 0$. When $\sigma > 0$ is sufficiently large compared to $\lambda$, non-monotonicity is again present, giving maximum generalization error at $\alpha \approx 1 + \lambda$ (\figref{fig:white}c), with an asymptotic fall $E_g \sim \frac{\sigma^2}{\alpha}$.

We find that $E_g(\alpha)$ is  non-monotonic when the noise level in target satisfies the following inequality (\figref{fig:white}d and Supplementary Note 3):
\begin{linenomath*}\begin{align}\label{eq:phase_boundary_white}
	\sigma^2 > \begin{cases} 
    g(\lambda) & \lambda< 1 \\
    2\lambda+1 & \lambda\geq 1
   \end{cases},
\end{align}\end{linenomath*}
where $g(\lambda) = 3\lambda\big[3\lambda+2-2\sqrt{1+\lambda}\sqrt{9\lambda+1}\cos\theta(\lambda)\big]$, and $ \theta(\lambda) = \frac{1}{3}\left(\pi+\tan^{-1}\frac{8\sqrt{\lambda}}{9\lambda(3\lambda+2)-1}\right)$. Although there is no strict phase transition (in the sense of non-analytic free energy) except at $\lambda=0$, Eq. \eqref{eq:phase_boundary_white} defines a phase boundary separating the monotonic and non-monotonic learning curve regions for a given regularization parameter and noise. For a given $\lambda$, double-descent occurs for sufficiently high $\sigma^2$.  In the non-monotonic region, there is a single local maximum when $\sigma^2 > 2\lambda + 1$, otherwise a local minima followed by a local maxima (we call only this kind of peak as the double-descent peak).

Based on this explicit formula, one could choose a large enough $\lambda$ to mitigate the peak to avoid overfitting for a given noise level (\figref{fig:white}d). However, larger $\lambda$ may imply slower learning (See \figref{fig:white}b and Supplementary Note 3) requiring more training samples to achieve the same generalization error. By inspecting the derivative $\frac{\partial E_g}{\partial \lambda}=0$, we find that $\lambda^* = \sigma^2$ (yellow dashed line in \figref{fig:white}d) is the optimal choice for ridge parameter, minimizing $E_g(\alpha)$ for a given $\sigma^2$ at all $\alpha$ (\figref{fig:white}c). For $\lambda > \lambda^*$ the noise-free error term increases from the optimum whereas $\lambda<\lambda^*$ gives a larger noise term. Our result agrees with a similar observation regarding the existence of an optimal ridge parameter in linear regression \cite{nakkiran2020optimal}.

\begin{figure}[h]
    \centering
    \includegraphics[scale = 1]{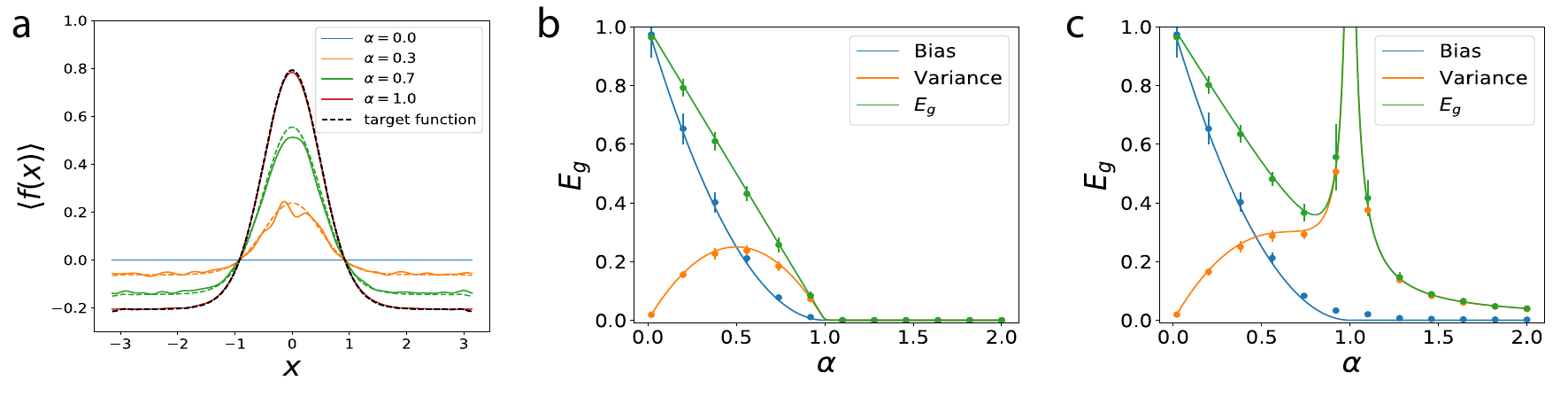}
    \caption{\textbf{Bias-variance decomposition of generalization error.} \textbf{a.}  Average estimator for kernel regression with $K(x,x') = \sum_{k=1}^N \cos(k(x-x'))$ on target function $\bar f(x) = e^{4(\cos x -1 )}$ with mean subtracted for different values of $\alpha = P/N$ when $\lambda = \sigma^2=0$. Estimator linearly approaches to the target function and estimates it perfectly when $\alpha = 1$. Dashed lines are theory. \textbf{b.} With the same setting in \figref{fig:white}, when $\lambda = 0$ and $\sigma^2=0$, the bias is a monotonically decreasing function of $\alpha$ while variance has a peak at $\alpha = 1/2$ yet it does not diverge. \textbf{c.} When $\lambda = 0$ and $\sigma^2=0.2$, we observe that the divergence of $E_g$ is only due to the diverging variance of the estimator. In \textbf{b.} and \textbf{c.}, solid lines are theory, dots are experiments. Error bars represent the standard deviation over $15$ trials.}
    \label{fig:bias_variance_estimator}
\end{figure}

Further insight to the phase transition can be gained by looking at the bias and the variance of the estimator \cite{montanari2019generalization,dascoli2020double,dascoli2020triple}. The average estimator learned by kernel regression linearly approaches the target function as $\alpha \to 1$ (Supplementary Note 2): $\left< f^*(\x) \right>_{\mathcal{D}} = \min\{\alpha,1\} \bar f(\x)$ (\figref{fig:bias_variance_estimator}a),
which indicates that the bias ($B$) and variance ($V$) contributions to generalization error have the forms
$B = \max\{0,1-\alpha\}^2$, $ V = \alpha(1-\alpha) \Theta(1-\alpha) + \frac{\sigma^2}{1-\alpha} \left[ \alpha \Theta(1-\alpha) -  \Theta(\alpha-1) \right]$.
In the absence of noise, $\sigma = 0$, variance is initially low at small $\alpha$, reaches its maximum at $\alpha = 1/2$ and then decreases as $\alpha \to 1$ as the learned function concentrates around $\bar f$ (\figref{fig:bias_variance_estimator}b). When there is noise, the phase transition at $\alpha = 1$ arises from the divergence in the variance $V$ of the learned estimator (\figref{fig:bias_variance_estimator}c).

\subsection*{Multiple Learning Episodes and Descents: Rotation Invariant Kernels and Measures}\label{Spherical} 

Next, we study kernel regression on high dimensional spheres focusing on rotation invariant kernels, which satisfy $K(\mathbf{O} \mathbf{x},\mathbf{O} \mathbf{x}') = K(\mathbf{x},\mathbf{x}')$, 
where $\mathbf{O}$ is an arbitrary orthogonal matrix. This broad class of kernels includes widely used radial basis function kernels $K(\mathbf{x},\mathbf{x}') = K(||\mathbf{x}-\mathbf{x}'||)$ (Gaussian, Laplace, Matern, rational quadratic, thin plate splines, etc) and dot product kernels $K(\mathbf{x},\mathbf{x}') = K(\mathbf{x} \cdot \mathbf{x}')$ (polynomial kernels, NNGPK and NTK) \cite{ cho2009kernel,genton2001classes,RasmussenWilliams}. 

When the data distribution is spherically isotropic $p(\mathbf{x}) = p(||\mathbf{x}||)$, we can separate Mercer eigenfunctions for rotation invariant kernels into radial and angular parts. The spherical parts depend on the radial distances of the data points $||\x||$, whereas the angular components admit a decomposition in terms of spherical harmonics of the unit vectors $Y_{km}(\hat x)$, where $k$ is the degree and $m$ is the order of the harmonic \cite{dai2013spherical}. A review of the basic properties of spherical harmonics are provided in the Supplementary Note 6. Utilizing this spherical symmetry, we obtain the following Mercer decomposition $K(\mathbf{x},\mathbf{x}') = \sum_{z k m} \eta_{z,k} R_{z,k}(||\mathbf{x}||) R_{z,k}(||\mathbf{x}'||) Y_{km}(\mathbf{\hat{x}}) Y_{km}(\mathbf{\hat{x}'})$.	
Since the eigenvalues are independent of the spherical harmonic order $m$, the minimal degeneracy of the RKHS spectrum is the number of degree $k$ harmonics: in the limit $D \to \infty$ given by $\sim {D^k}/{k!}$ \cite{bietti2019inductive,yang2020finegrained}. However, the degeneracy can be even larger if there are different $(z,k)$ indices with the same eigenvalue. 
For notational convenience, we denote degenerate eigenvalues as $\eta_{K}$ ($K\in \mathbb{Z}^+$) and corresponding eigenfunctions as $\phi_{K,\rho}$ where $\rho \in \mathbb{Z}^+$ indexes the degenerate indices. After finding the eigenvalues of a kernel on the basis $\phi_{K,\rho}$, one can evaluate  Eq. \eqref{eq:finitePgenErr} to predict the generalization error of the kernel machine.

We focus on the case where the degeneracy of $\eta_{K}$ is $N(D,K) \sim \mathcal{O}_D(D^K)$.  Correspondingly, for finite kernel power $\left< K(\mathbf{x},\mathbf{x}) \right>_{\mathbf{x} \sim p(\mathbf{x})}$ , the eigenvalues must scale with dimension like $\eta_{K} \sim \mathcal{O}_D(D^{-K})$ \cite{cohen2019learning,bordelon2020spectrum}. Examples include the widely-used Gaussian kernel and dot product kernels such as NTK, which we discuss below and in Supplementary Note 4.

\begin{figure}
    \centering
    \includegraphics[scale=1]{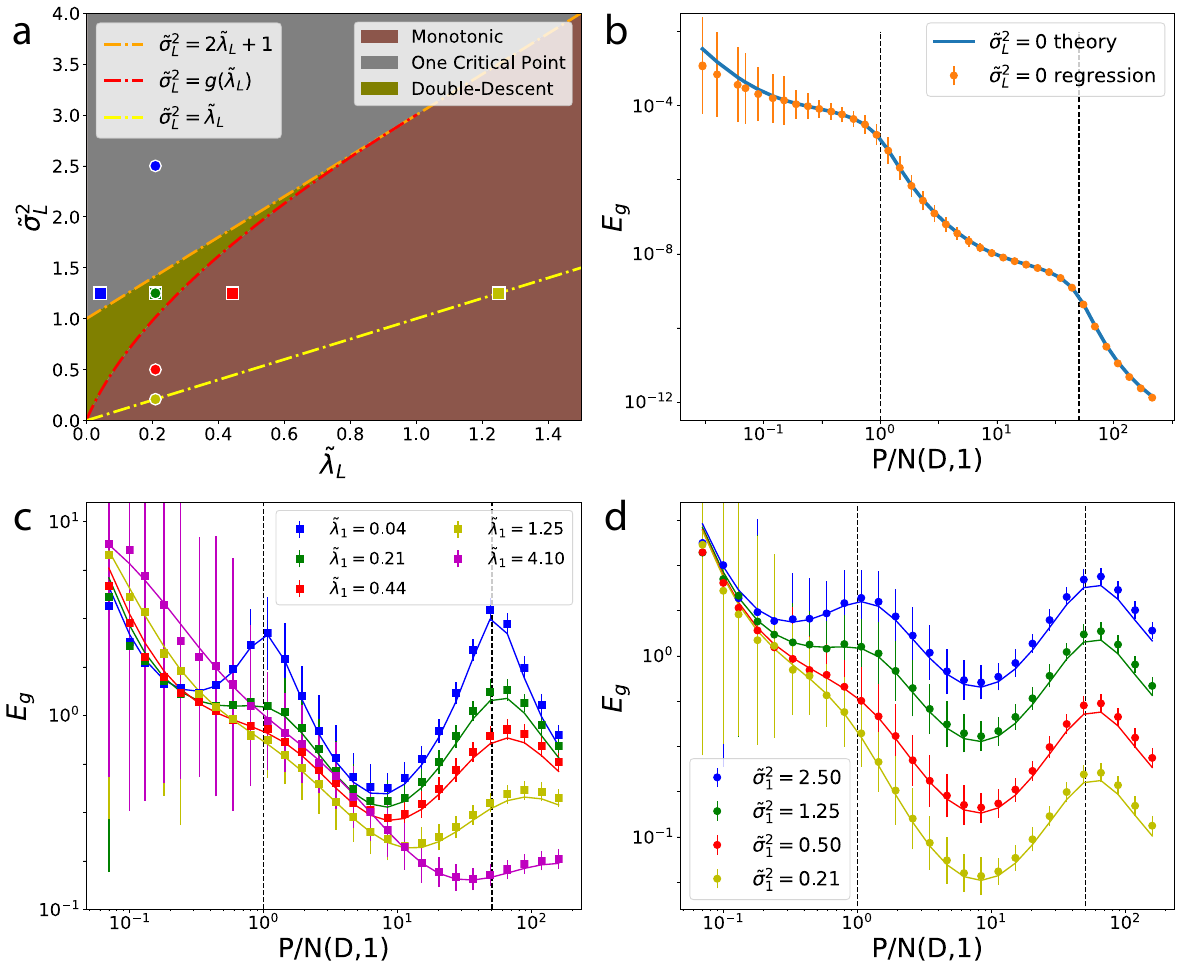}    \caption{\textbf{Gaussian RBF kernel regression on high-dimensional spherical data.} \textbf{a.} Phase diagram for non-monotonic learning curves obtained from the theory by counting the zeros of $\frac{\partial E_g}{\partial \alpha}$. Colored squares and colored circles correspond to curves in panel \textbf{c} and \textbf{d}, respectively. \textbf{b.} Kernel regression with Gaussian RBF $K(\mathbf{x},\mathbf{x}')=e^{-\frac{1}{2 D \omega^2} ||x-x'||^2}$ with $\omega = 3$, $D = 100$ and noise-free labels. Target is $\bar f(\x) = \sum_{k,m}\bar w_{km}\sqrt{\eta_{km}}Y_{km}(\x)$ with random and centered weights $\bar w_{km}$ such that $\braket{\bar w_{km}^2} = \eta_{km}$ (Supplementary Note 5). Dashed lines represent the locations of $N(D,1)$ and $N(D,2)$, showing different learning stages. \textbf{c, d.} Generalization error for Gaussian RBF kernel for various kernel widths $\omega$ corresponding to specific $\tilde\lambda_L$'s and noise variances $\tilde\sigma_L$ pointed in the phase diagram in $D = 100$. Solid lines - theory (Eq. \eqref{eq:finitePgenErr}). Larger regularization suppresses the descent peaks, which occur at $P^* \sim  N(D,L)$ shown by the vertical dashed lines. \textbf{c.} Varying $\tilde\lambda_L$ with fixed the $\tilde\sigma_L$. \textbf{d.} vice versa. For fixed noise, we observe an optimal $\tilde\lambda_1$ for up to $P/N(D,1)\sim 10$ after which the next learning stage starts. Error bars indicate standard deviation over $300$ trials for \textbf{b} and $100$ trials for \textbf{c, d}.}\label{fig:dot_prod_phase}
\end{figure}

This scaling from the degeneracy allows us to consider multiple $P, D \to \infty$ limits leading to different learning stages. We consider a separate limit for each degenerate eigenvalue $L$ while keeping $\alpha \equiv P/N(D,L)$ finite. With this setting, we evaluate Eq. \eqref{eq:finitePgenErr} with definitions $
    \bar\eta_K \equiv N(D,K)\eta_K$,  $\bar w_{K}^2 \equiv \frac{1}{N(D,K)}\sum_\rho \bar w_{K,\rho}^2$,
to obtain the generalization error in learning stage $L$:
%
\begin{linenomath*}\begin{equation}\label{eq:asympGenErr}
\begin{gathered}
E^{(L)}_g(\alpha)=\bar\eta_{L}\bar w_{L}^2 \frac{\tilde \kappa^2+\tilde\sigma_{L}^2\alpha}{(\tilde \kappa + \alpha)^2-\alpha}+\sum_{K>L}\bar\eta_{K}\bar w_{K}^2,\\
	\tilde \kappa(\alpha) = \frac{1}{2}(1+\tilde\lambda_{L}-\alpha)+\frac{1}{2}\sqrt{(\alpha+1+\tilde\lambda_{L})^2-4\alpha},\\
	\tilde \sigma_{L}^2 \equiv \frac {\sigma^2+E^{(L)}_g(\infty)}{\bar\eta_{L}\bar w_{L}^2},\quad \tilde\lambda_{L} \equiv  \frac{\lambda+\sum_{K>L}\bar\eta_{K}}{\bar\eta_{L}}.
\end{gathered}
\end{equation}\end{linenomath*}
%
Several observations can be made:
\begin{enumerate}[noitemsep,topsep=0pt,leftmargin=*]
\item We note that $E_g^{(L)}(0) = \bar\eta_L\bar w_{L}^2 + \sum_{K> L}\bar\eta_{K}\bar w_{K}^2 = \bar\eta_L\bar w_{L}^2+E^{(L)}_g(\infty)$.
In the learning stage $L$, generalization error due to all target modes with $K<L$ has already decayed to zero. As $\alpha \to \infty$, $K=L$ modes of the target function are learned, leaving $K>L$ modes. This illustrates an inductive bias towards learning target function modes corresponding to higher kernel eigenvalues.
\item  $E_g^{(L)}(\alpha)-E^{(L)}_g(\infty)$ reduces, up to a constant $\bar\eta_{L}\bar w_{L}^2$, to the generalization error in the band limited case,  Eq. \eqref{Egband}, with the identification of an \textit{effective noise parameter},  $\tilde \sigma_L$, and an \textit{effective ridge parameter}, $\tilde\lambda_{L}$. Inspection of $\tilde \sigma_L$ reveals that target modes with $K>L$ ($E^{(L)}_g(\infty)$) act as noise in the current stage. Inspection of $\tilde\lambda_{L}$ reveals that kernel eigenvalues with $K>L$ act as a regularizer in the current stage. The role of the number of eigenvalues in the white band limited case, $N$, is played here by the degeneracy $N(D,L)$.
\item Asymptotically, first term in $E_g^{(L)}(\alpha)$ is monotonically decreasing with $\alpha^{-2}$, while the second term shows non-monotonic behavior having a maximum at $\alpha = 1+\tilde\lambda_{L}$. Similar to the white band-limited case, generalization error diverges due to variance explosion at $\alpha = 1+\tilde\lambda_{L}$ when $\tilde\lambda_{L} = 0$ (a band-limited spectrum is possible) implying again a first order phase transition.  Non-monotonicity caused by the noise term implies a possible peak in the generalization error in each learning stage. A phase diagram can be drawn, where phase boundaries are again defined by Eq. \eqref{eq:phase_boundary_white} evaluated with the effective ridge and noise parameters, \figref{fig:dot_prod_phase}a.

\item Similar to the white band limited case,  optimal regularization happens when
\begin{linenomath*}\begin{equation}
    \tilde\lambda_{L}=\tilde \sigma_L^2,
\end{equation}\end{linenomath*}
minimizing $E_g^{(L)}(\alpha)$ for a given $\tilde \sigma_L$ for all $\alpha$. This result extends the previous findings on linear regression \cite{nakkiran2020optimal} to the large class of rotation invariant kernels.
\item When all stages are considered, it is possible to observe learning curves with \textit{multiple descents} with at most one peak per stage. The presence and size of the descent peak depends on the level of noise in the data and the effective regularization as shown in Eq. \eqref{eq:phase_boundary_white} and Eq. \eqref{eq:asympGenErr}. Similar observations of multiple peaks in the learning curves were made in \cite{liang_isometry} in the context of ridgeless regression on polynomial kernels.
\end{enumerate}

As an example of the effect of kernel spectrum on double-descent, consider a power law $\bar\eta_K = K^{-s}$ where $s\geq 1$. Then $\tilde\lambda_L=L^s\big(\zeta(s,L )+\lambda\big)-1 \approx \frac{L}{s-1}+\lambda L^s,\;\;(L\gg 1)$, where $\zeta(s,L)$ is Hurwitz-Zeta function. In the ridgeless $\lambda = 0$ case, faster decaying spectra (higher $s$, smaller $\tilde\lambda_L$) are more prone to double-descent than the slower ones (\figref{fig:dot_prod_phase}a). Furthermore, we also observe that higher modes (higher $L$, higher $\tilde\lambda_L$) are more immune to overfitting, signaled by non-monotonicity, than the lower modes.

We apply our theory to Gaussian RBF regression on synthetic data in \figref{fig:dot_prod_phase} where \figref{fig:dot_prod_phase}b demonstrates a perfect agreement with theory and experiment on Gaussian RBF with synthetic data when no label noise is present. The vertical dashed lines represent the locations where $P = N(D,1)$ and $P = N(D,2)$, respectively. \figref{fig:dot_prod_phase}c shows the regression experiment with the parameters $(\tilde\sigma_1^2,\tilde\lambda_1)$ indicated by colored squares on the phase diagram (\figref{fig:dot_prod_phase}a). When the parameters chosen on the yellow dashed line in \figref{fig:dot_prod_phase}a, corresponding to the optimal regularization for fixed effective noise, the lowest generalization error is achieved in the first learning episode without a double-descent. Finally, \figref{fig:dot_prod_phase}d shows the theory and experiment curves with the parameters $(\tilde\sigma_1^2,\tilde\lambda_1)$ shown by the colored circles in \figref{fig:dot_prod_phase}a. As expected, for fixed effective regularization, increasing noise hurts generalization. For further experiments see
Supplementary Note 4.

\subsection*{Dot Product Kernels, NTK and Wide Neural Networks} 

Our theory allows the study of generalization error for trained and wide feedforward neural networks by exploiting a correspondence with kernel regression. When weights in each layer are initialized from a Gaussian distribution $\mathcal{N}(0,\sigma^2_W)$ and the size of hidden layers tend to infinity, the function $f(\mathbf{x},\bm{\theta})$ learned by training the network parameters $\bm{\theta}$ with gradient descent on a squared loss to zero training error is equivalent to the function obtained from ridgeless ($\lambda=0$) kernel regression with the NTK: $ \mathbf{K}_{NTK} (\mathbf{x}_i, \mathbf{x}_j) =  \nabla_{\bm{\theta}} f(\mathbf{x}_i, \bm{\theta}_0) \cdot \nabla_{\bm{\theta}} f(\mathbf{x}_j, \bm{\theta}_0)$ \cite{jacot2018neural}. For fully connected neural networks, the NTK is a dot product kernel $K_{NTK}(\mathbf{x}, \mathbf{x}') = K_{NTK}(\mathbf{x}\cdot  \mathbf{x}')$ \cite{jacot2018neural,bordelon2020spectrum}. For such kernels and spherically symmetric data distributions $p(\mathbf{x})=p(\Vert\mathbf{x}\Vert)$, kernel eigenfunctions do not have a radial part, and consequently the eigenvalues are free of a $z$-index. Therefore, $k$-th eigenvalue has degeneracy of the degree $k$ spherical harmonics, $\mathcal{O}_D(D^k)$, ($K,L\to k,l$ and $\rho \to m$) \cite{bordelon2020spectrum}, allowing recourse to the same scaling we used to analyze rotation invariant kernels in the previous section. The learning curves for infinitely wide neural network will thus have the same form in  Eq. \eqref{eq:asympGenErr}, evaluated with NTK eigenvalues and with $\lambda = 0$.

\begin{figure}[ht]
    \centering
    \includegraphics[scale=1]{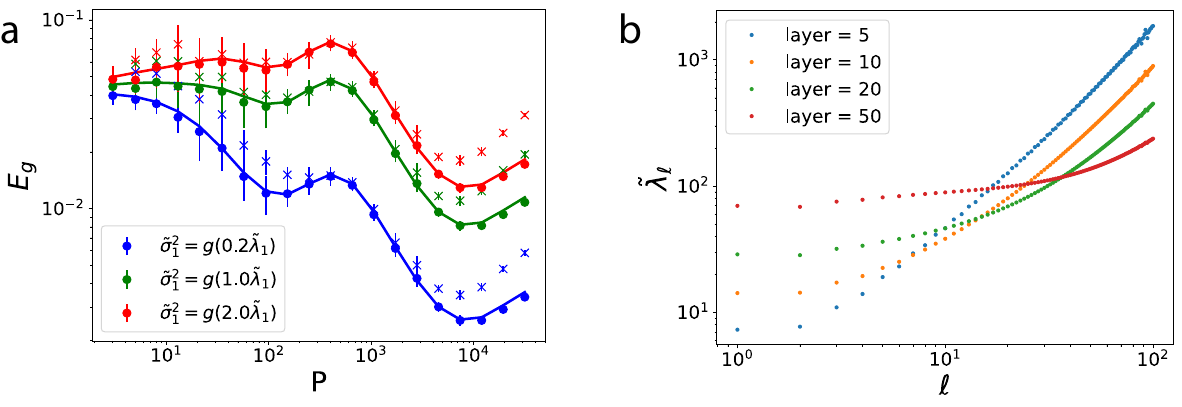}
    \caption{\textbf{Comparison of our theory with finite width neural network experiments.} \textbf{a.} 2-layer NTK regression and corresponding neural network training using NeuralTangents package \cite{neuraltangents2020} with 50000 hidden units for $D = 25$ with varying noise levels chosen according to $g(\lambda)$. Target function is a single degree mode $\bar f(\x) = c_k Q^{(D-1)}_k(\bm{\beta}\cdot\x)$, where $c_k$ is a constant, $\bm{\beta}$ is a random vector, and $Q_k^{(D-1)}$ is the $k$-th Gegenbauer polynomial (see Supplementary Note 5 and 6). Here we picked $k=1$ (linear target). Solid lines are the theory predicted learning curves (Eq. \eqref{eq:finitePgenErr}), dots represent NTK regression and $\times$ represents $E_g$ after neural network training. Correspondence between NN training and NTK regression breaks down at large sample sizes $P$ since the network operates in under-parameterized regime and finite-size effects become dominating in $E_g$. Error bars represent standard deviation of $15$ averages for kernel regression and $5$ averages for neural network experiments. \textbf{b.} $\tilde\lambda_l$ dependence to mode $l$ across various layer NTKs. The weight and bias variances for the neural network are chosen to be $\sigma^2_W = 1$ and $\sigma^2_b = 0$, respectively.}
    \label{fig:neural_network_exp}
\end{figure}

In \figref{fig:neural_network_exp}a, we compare the prediction of our theoretical expression for $E_g$, Eq. \eqref{eq:finitePgenErr}, to NTK regression and neural network training. The match to NTK training is excellent. We can describe neural network training up to a certain $P$ after which the correspondence to NTK regression breaks down due to the network's finite-width. For large $P$, the neural network operates in under-parameterized regime where the network initialization variance due to finite number of parameters starts contributing to the generalization error \cite{belkin2019reconciling, montanari2019generalization, montanari2019surprises, dascoli2020double}. A detailed discussion of these topics is provided in Supplementary Note 4.
	
Neural networks are thought to generalize well because of implicit regularization \cite{zhang2016understanding,bietti2019inductive,yang2020finegrained}. This can be addressed with our formalism. For spherical data, we see that the implicit regularization of a neural network for each mode $l$ is given by $\tilde\lambda_l = \frac{\sum_{k>l}\bar\eta_k}{\bar\eta_l}$. As an example, we calculate the spectrum of NTK for rectifier activations, and observe that the spectrum whitens with increasing depth \cite{yang2020finegrained}, corresponding to larger $\tilde\lambda_{l}$ and therefore more regularization for small learning stages $l$ (\figref{fig:neural_network_exp}b). The trend for small degree harmonics $l$ is especially relevant since, as we have shown, approximately $D^l$ samples are required to learn degree $l$ harmonics. In this small $l$ regime, we see that deep networks exhibit much higher effective regularization compared to shallow ones due to slower decay of eigenvalues.

\section*{Discussion}

We studied generalization in kernel regression using statistical mechanics and the replica method \cite{mezard1987spin}. We derived an analytical expression for the generalization error, Eq. \eqref{eq:finitePgenErr}, valid for any kernel and any dataset. We showed that our expression explains generalization on real datasets, and provided a detailed analysis of its application to band-limited kernels with white spectra and the widely used class of rotation invariant kernels \cite{genton2001classes,RasmussenWilliams} operating on spherical data. For the latter case, we defined an effective regularization and an effective noise which govern the generalization behavior, including non-monotonicity of learning curves. It will be interesting to see if analogues of these concepts can be defined for real datasets. Our results are directly applicable to infinite-width limit of neural networks that admit a kernel description (including feedforward, convolutional and recurrent neural networks) \cite{jacot2018neural, neuraltangents2020, alemohammad2020recurrent, arora2019exact, yang2019tensor1, yang2020tensor2}, and explain their inductive bias towards simple functions \cite{neyshabur_inductivebias, lee2019wide, jacot2020implicit, chizat2020implicit, bietti2019inductive, ghorbani_twolayer, soudry_implicitbias}. We also note a closely related recent study  \cite{jacot2020kernel}, which we discuss further in Supplementary Discussion, that utilizes random matrix theory to study generalization in kernel regression.

One goal of our present work is to provide a framework that incorporates structural information about the data distribution into a realistic prediction of generalization performance that holds for real data and any kernel.  Indeed, a recent study suggested that structure in data allows kernel methods to outperform pessimistic generalization expectations based on the high ambient dimension \cite{spigler2020asymptotic}. In a different setting, authors of \cite{liao2020random} calculate test error of random Fourier features model using random matrix theory techniques without strong assumptions on data distribution and obtain excellent agreement on real datasets. Overall, our results demonstrate how data and inductive biases of a model interact to shape generalization behavior, and in particular the importance of the compatibility of a learning task with the model for sample-efficient learning. Our findings elucidate three heuristic principles for generalization in kernel regression. 

First is the spectral bias. The eigendecomposition of the kernel provides a natural ordering of functions which are easiest to estimate. Decomposing generalization error into modal errors, we found that errors in spectral modes with large eigenvalues decrease more rapidly with increasing sample size than modes with small eigenvalues, also observed in \cite{bordelon2020spectrum}, illustrating a preference to fit certain functions over others.
Our findings are consistent with other experimental results and analytical ones which derive error bounds on test risk to elucidate the spectral or frequency bias of NTK and NNGP \cite{cao2019understanding,rahaman2018spectral,luo2019frequencybias,valleprez2020generalization}.

Consequently, how a given task decomposes in the eigenbasis, a heuristic that we name task-model alignment, determines the number of samples required to achieve good performance: tasks with most of their power in top eigenmodes can be learned in a sample efficient manner. We introduced cumulative power distribution as a metric for task-model alignment and proved that target functions with higher cumulative power distributions will have lower normalized generalization error for all $P$ under the same kernel and data distribution. A related notion of kernel compatibility with target was defined in \cite{cristianini2002kerneltarget,cortes2012algorithms}, which we discuss in detail in Supplementary Discussion.

The third phenomenon we explore is how non-monotonicity can appear in the learning curves when either labels are noisy, or the target function has modes that is not expressible with the kernel. Non-monotonicity is caused by the variance term in the bias-variance decomposition of the generalization error. In the analytically tractable models we considered, this is related to a phase transition appearing in separate learning stages for the rotation invariant kernels. Non-monotonicity can be mitigated with explicit or implicit regularization \cite{montanari2019generalization,dascoli2020double,nakkiran2019moredata}. We showed the existence of an optimal regularizer, independent of sample size, for our theoretical settings. When applied to linear regression, our optimal regularizer matches that previously given by \cite{nakkiran2020optimal}.

Non-monotonicity in generalization error gathered a lot of interest recently. Many studies pointed to absence of overfitting in overparameterized machine learning models, signaled by a peak and a subsequent descent in generalization error as the model complexity, or the number of parameters, increases, and the model transitions from an underparameterized to overparameterized (interpolating) regime \cite{belkin2019reconciling,nakkiran2019deep,jacot2020implicit,montanari2019surprises,montanari2019generalization,dascoli2020double,dascoli2020triple,spigler2019jamming,mezard2020generalisation, liao2020random}. Multiple peaks are also possible in this context \cite{adlam2020neural}. Our work provides an explanation for the lack of overfitting in overparameterized models by elucidating strong inductive biases of kernel regression, valid even in the interpolation limit, which includes infinitely overparameterized limits of neural networks. Sample-wise non-monotonicity has also been observed previously in many models \cite{nakkiran2019deep,krogh,Hertz_1989, montanari2019surprises,nakkiran2019moredata}, including ones that show multiple peaks \cite{nakkiran2020optimal,dascoli2020triple,liang_isometry,chen2020multipledesign}. A closely related study obtained an upper bound for test risk in ridgeless regression which shows non-monotonic behavior with increasing sample size whenever $P\sim \mathcal{O}(D^L)$, consistent with our results on rotation invariant kernels and isotropic data. 

An interesting comparison can be made between the multiple peaks we observed in our analytically solvable models and the multiple peaks observed in random features models \cite{dascoli2020triple,adlam2020neural}. In these models, one of the peaks (termed ``nonlinear'' in \cite{dascoli2020triple}) happens when the number of samples reaches the number of features, and thus the number of parameters of the model, crossing the interpolation threshold. While the peak we observed in the white band limited case with nonlinear features also happens at the interpolation threshold ($P=N$), the mechanisms causing double descent are different. In random features models, this peak is due to variance in the initialization of the random feature vectors. In our example, such variance does not exist. The peak is due to overfitting the noisy labels and disappears when there is no noise. The peaks observed for the rotationally invariant case has a more subtle connection. In each learning stage, peaks occur when number of samples reach the degeneracy of eigenfunctions in that stage. While kernel regression is non-parametric, one can think of this again as crossing an interpolation threshold defined by the dimensionality of the feature space in the large-$D$ limit. Like the white band limited case, these peaks are due to overfitting noise.

While our theory is remarkably successful in its predictions, it has limitations. First, the theory requires eigendecomposition of the kernel on the full dataset which is computationally costly. 
Second, its applicability to state-of-the-art neural networks is limited by the kernel regime of networks, which does not capture many interesting and useful deep learning phenomena \cite{fort2020deep,chizat2020implicit}.  Third, our theory uses a Gaussian approximation \cite{Dietrich_Sompolinsky} and a replica symmetric ans\"atz \cite{mezard1987spin}. While these assumptions were validated by the remarkable fit to experiments, it will be interesting to see if their relaxations reveal new insights.


\section*{Methods}\label{Methods}

\subsection*{Statistical Mechanics Formulation}

With the setting described in the main text, kernel regression problem reduces to minimization of the energy function
\begin{linenomath*}\begin{equation}\label{eq:energy_fn}
     H(\w) \equiv \frac{1}{2\lambda} \sum_{\mu=1}^P\left(\sum_{\rho=1}^N(\bar w_{\rho}-w_{\rho})\psi_{\rho}(\x^\mu)+\epsilon^\mu\right)^2 + \frac{1}{2}\Vert\w\Vert^2_2.
\end{equation}\end{linenomath*}
The quantity of interest is the generalization error in equation Eq. \eqref{eq:genError}, which in matrix notation is
\begin{linenomath*}\begin{equation}
    E_g =  \left<(\w^*-\bar\w)^\top\bLambda(\w^*-\bar\w)\right>_{\mathcal{D}},
\end{equation}\end{linenomath*}
where $\bLambda_{\rho\gamma} = \eta_\rho\delta_{\rho\gamma}$ represents a diagonal matrix with entries given by the RKHS eigenvalues $\eta_\rho$.

In order to calculate the generalization error, we introduce a Gibbs distribution $p_G(\w)\equiv \frac{1}{Z}e^{-\beta H(\w)}
$ with the energy function in Eq. \eqref{eq:energy_fn}. In the $\beta \to \infty$ limit, this Gibbs measure is dominated by the solution to the kernel regression problem. We utilize this fact to calculate the generalization error for kernel regression. This can be done by introducing a source term with strength $J$ to the partition function,
\begin{linenomath*}\begin{equation}
     Z(J) = \int d\w e^{-\beta H(\w,\mathcal{D})+{J \beta P} (\w-\bar\w)^\top\bLambda(\w-\bar\w)} \ , \ E_g(\mathcal{D}) = \left.\lim_{\beta\to \infty}\frac{1}{\beta P} \frac{d}{d J}\ln Z(J)\right |_{J=0},
\end{equation}\end{linenomath*}
where we recognize the free energy $\beta F \equiv -\ln Z(J)$ is the relevant quantity to compute generalization error for a given dataset, $E_g(\mathcal{D})$. In Supplementary Note 2, we introduce other source terms to calculate training error, average estimator and its variance.

The free energy depends on the sampled dataset $\mathcal{D}$, which can be thought of as a quenched disorder of the system. Experience from the study of physics of disordered systems suggests that the free energy concentrates around its mean (is self-averaging) for large $P$ \cite{mezard1987spin}. Therefore, we calculate the typical behavior of the system by performing the average free energy over all possible datasets: $\beta  F = \beta \left< F  \right >_{\mathcal D}= -  \braket{\ln Z(J)}_{\mathcal{D}}$ in the $P\to\infty$ limit. 

All calculations are detailed in Supplementary Note 1 and 2. Here we provide a summary. To perform averages over the quenched disorder, we resort to the replica trick \cite{spinGlassReview} using $\braket{\log Z(J)}_{\mathcal{D}} = \lim_{n\to 0}\frac{1}{n}(\braket{Z(J)^n}_{\mathcal{D}}-1)$.
A key step is a Gaussian approximation to the average over the dataset in the feature space \cite{sompolinsky1999statistical}, which exploits the orthogonality of the feature vectors with respect to the input distribution $p(\x)$. These averages are expressed in terms of order parameters defining the mean and the covariance of the Gaussian. The calculation proceeds by a replica symmetric ansatz \cite{mezard1987spin},
evaluating the saddle point equations and taking the $\beta\to\infty$ limit. 

\subsection*{Modal errors}

The generalization error can be written as a sum of modal errors arising from the estimation of the coefficient for eigenfunction $\psi_\rho$:
\begin{linenomath*}\begin{equation} 
E_g = \sum_\rho \eta_\rho \overline{w}_\rho^2 E_\rho,
\end{equation}\end{linenomath*}
where
\begin{linenomath*}\begin{equation}
E_\rho = \frac{1}{\overline{w}_\rho^2} \left< ( {w}^*_\rho - \overline{w}_\rho)^2 \right>_{\mathcal D} = \frac{1}{1-\gamma} \frac{\kappa^2 }{(\kappa+P\eta_\rho)^2} .
\end{equation}\end{linenomath*}
We now provide a proof that the mode error equation implies that the logarithmic derivatives of mode errors have the same ordering as the kernel eigenvalues when $\sigma=0$. Assuming that $\eta_\rho > \eta_{\rho'}$, and noting that $ \kappa'(P) = - \frac{\kappa \gamma}{P(1 + \gamma)} < 0$ since $\kappa, \gamma, P > 0$, we have
\begin{linenomath*}\begin{equation}
    \frac{d}{dP} \log \left( \frac{E_\rho}{E_{\rho'}} \right)  = - 2 \left[ \frac{\kappa'(P) + \eta_\rho}{\kappa+P\eta_\rho} - \frac{\kappa'(P) + \eta_{\rho'}}{\kappa + P\eta_{\rho'}} \right] <0.
\end{equation}\end{linenomath*} 
Thus, we arrive at the conclusion
\begin{linenomath*}\begin{equation}
    \frac{d}{dP} \log E_\rho  < \frac{d}{dP} \log E_{\rho'} \quad \implies \quad 
    \frac{1}{E_\rho} \frac{dE_\rho}{dP} < \frac{1}{E_{\rho'}} \frac{d E_{\rho'}}{dP}.
\end{equation}\end{linenomath*}
Let $u_{\rho,\rho'}(P) = \log\left( \frac{E_\rho}{E_{\rho'}} \right)$. The above derivation demonstrates that $\frac{d}{dP} u_{\rho,\rho'}(P) < 0$ for all $P$. Since $u_{\rho,\rho'}(0) = 0$,  this implies that $u_{\rho,\rho'}(P) < 0$ or equivalently $E_\rho < E_{\rho'}$ for all $P$. This result indicates that the mode errors have the opposite ordering of eigenvalues, summarizing the phenomenon of \textit{spectral bias} for kernel regression: the generalization error falls faster for the eigenmodes with larger eigenvalues. If the target function has norm $T = \left<\bar f(\x)^2\right>=\sum_{\rho} \eta_{\rho} \overline{w}_{\rho}^2$ then the generalization error is a convex combination of $\{ T E_\rho(P) \}_{\rho=1}^{\infty}$. The quantities $T E_\rho(P)$ maintain the same ordering as the normalized mode errors $E_\rho$ for all $P$, and we see that re-allocations of target function power that strictly increase the cumulative power distribution $C(\rho) = \frac{1}{T} \sum_{\rho' \leq \rho} \eta_{\rho'} \overline{w}_{\rho'}^2 $ curve must cause a reduction in generalization error. We emphasize that, for a fixed set of kernel eigenvalues, strictly higher $C(\rho)$ yields better generalization but provide a caveat: for a fixed target function, comparison of different kernels requires knowledge of both the change in the spectrum $\eta_\rho$ as well as changes in the $C(\rho)$ curve.

\subsection*{Diagonalizing the kernel on real datasets}

Calculation of $E_g$ requires two inputs: kernel eigenvalues $\eta_\rho$ and teacher weights $\bar w_\rho$, both calculated using the underlying data distribution. For a dataset with a finitely many samples, we assume a discrete uniform distribution over data $p(\mathbf{x}) = \frac{1}{M} \sum_{i=1}^{M} \delta(\mathbf{x}-\mathbf{x}_i)$ with $M$ being the size of the whole dataset (train+test).  Then, the corresponding eigenvalue problem reads:
\begin{linenomath*}\begin{equation}\label{eq:KPCA}
    \eta_k \phi_k(\x') = \int p(\x) K(\x,\x') \phi_k(\x) d\x = \frac{1}{M} \sum_{i=1}^M K(\x_i,\x') \phi_k(\x_i).
\end{equation}\end{linenomath*}
Given a kernel $K(\x,\x')$, one can evaluate the $M\times M$ kernel Gram matrix $\K_{ij} = K(\x_i,\x_j)$ and solve for the eigenvalues $\bLambda_{kl}=\eta_{k}\delta_{kl}$ and eigenfunctions $\bPhi_{ki} = \phi_k(\x_i)$ of $\K = N\bPhi\bLambda\bPhi^\top$. Note that both data indices and eigen-indices take values $i,k = 1,...,M$. For a target function with multiple classes $\bar f(\x):\mathbb{R}^D\to\mathbb{R}^C$, we denote the one-hot encoded labels $\Y = [\y_1,...,\y_C]\in\mathbb{R}^{M\times C}$ and obtain the teacher weights for each class with $\bar\w_c = \frac{1}{M} \mathbf{\Lambda}^{-1/2} \mathbf\Phi^\top \mathbf y_c $. For solving kernel regression, each of the $C$ one-hot output channels can be treated as an individual target function $f_{t,c}(\x)$ where $f_{t,c}(\x^\mu) = y_c^\mu$ for one-hot labels $y^\mu_c$. The weights $\overline{\w}_c$ obtained above allows the expansion of each teacher channel in the kernel's eigenbasis $f_{t,c}(\x) = \sum_{k=1}^M \overline{w}_{c,k} \sqrt{\eta_k} \phi_k(\x)$.  The total generalization error for the entire task is simply $E_g = \sum_{c=1}^C \left< (f_c^*(\x) - f_{t,c}(\x))^2 \right>_{\x,\mathcal{D}}$ where $f_c^*$ is the kernel regression solution for output channel $c$. Note that, with the choice of one-hot labels, the total target power is normalized to $1$. After computing learning curves for each channel $c$, which requires plugging in $\{(\eta_k,\overline{w}_{c,k}^2)\}_{k=1}^M$ into our theory, the learning curves for each channel are simply summed to arrive at the final generalization error.

In other cases, when we do not generally possess a priori knowledge about $p(\x)$, the underlying data distribution, solving the kernel eigenvalue problem in Eq. \eqref{eq:eigenvalue} appears intractable. However, when we are provided with a large number of samples from the dataset, we may approximate the kernel eigenvalue problem by using a Monte-Carlo estimate of the data density i.e. $p(\mathbf{x}) \approx \frac{1}{M} \sum_{i=1}^{M} \delta(\mathbf{x}-\mathbf{x}_i)$ with $M$ being the size of the dataset. Then Eq. \eqref{eq:KPCA} approximates the eigenvalues and eigenvectors where one obtains the exact eigenvalues when the number of samples is large, $M\to\infty$ \cite{RasmussenWilliams}.

\section*{Data Availability}

All data generated and/or analyzed in this study, along with the analysis code, are available in the Github repository  \cite{codeanddata2021canatar}, \href{https://github.com/Pehlevan-Group/kernel-generalization}{https://github.com/Pehlevan-Group/kernel-generalization}.

\section*{Code Availability}

All code used to perform numerical experiments, analyze data and generate figures are available in the Github repository \cite{codeanddata2021canatar}, \href{https://github.com/Pehlevan-Group/kernel-generalization}{https://github.com/Pehlevan-Group/kernel-generalization}.




\section*{Acknowledgements}

We thank the Harvard Data Science Initiative and Harvard Dean’s Competitive Fund for Promising Scholarship for their support. We also thank the reviewers for their detailed and constructive feedback, and Jan Drugowitsch and Guillermo Valle-Perez for comments on the manuscript.

\section*{Author Contributions}

A.C., B.B. and C.P. contributed to the planning of the study. A.C. and B.B. performed the analytical calculations and numerical simulations. A.C., B.B. and C.P. contributed to the interpretation of the results and writing of the manuscript.

\section*{Competing Interests}

The authors declare no competing interests.



\section*{Supplementary material}

# Supplementary Information for "Spectral Bias and Task-Model Alignment Explain Generalization in Kernel Regression and Infinitely Wide Neural Networks"

Abdulkadir Canatar [1,2]        Blake Bordelon [2,3]        Cengiz Pehlevan [2,3,*]

# Contents

[1]Department of Physics, Harvard University, Cambridge, MA 02138, USA.

[2]Center for Brain Science, Harvard University, Cambridge, MA 02138, USA.

[3]John A. Paulson School of Engineering and Applied Sciences, Harvard University, Cambridge, MA 02138, USA.

[*]Corresponding Author, E-mail: cpehlevan@g.harvard.edu

# Supplementary Note 1 - Problem Setup

A reproducing kernel Hilbert space [1] $\mathcal{H}$ living on $\mathcal{X} \subset \mathbb{R}^D$ is a subset of square integrable functions $L_2(\mathcal{X}, p)$ for measure $p$, equipped with an inner product $\langle \cdot, \cdot \rangle_{\mathcal{H}}$ and a kernel satisfying the following reproducing property:

$$\langle f(\cdot), K(\cdot, \mathbf{x}) \rangle_{\mathcal{H}} = f(\mathbf{x}), \quad \forall f(\cdot) \in \mathcal{H}, \ \forall \mathbf{x} \in \mathcal{X}, \tag{1}$$

with $K(., \mathbf{x})$ is itself being an element of $\mathcal{H}$. We assume data is drawn from a finite measure with density $p(\mathbf{x})$ and that the kernel has finite trace on this measure $\int K(\mathbf{x}, \mathbf{x}) p(\mathbf{x}) d\mathbf{x} < \infty$. We introduce the integral operator $T_K : L_2(\mathcal{X}, p) \to L_2(\mathcal{X}, p)$ which is a linear map from functions to functions

$$T_K[f](\mathbf{x}') = \int p(\mathbf{x}) K(\mathbf{x}, \mathbf{x}') f(\mathbf{x}) d\mathbf{x}. \tag{2}$$

*Mercer's Theorem* allows the diagonalization of the kernel in terms of the eigenfunctions of $T_K$, $\{\phi_\rho(\mathbf{x})\}|_{\rho=0}^\infty$ where $T_K[\phi_\rho] = \eta_\rho \phi_\rho$ with $\eta_\rho \geq 0$ [1, 2, 3]. This set of eigenfunctions forms an orthonormal basis with respect to $L_2(\mathcal{X}, p)$. The resulting Mercer decomposition of the kernel is

$$K(\mathbf{x}, \mathbf{x}') = \sum_{\rho=0}^\infty \eta_\rho \phi_\rho(\mathbf{x}) \phi_\rho(\mathbf{x}'). \tag{3}$$

We refer to $\{\eta_\rho\}$ as the spectrum of RKHS, or the kernel eigenvalues. The number of nonzero kernel eigenvalues defines the dimension of the RKHS, which is infinite in the general setting. By the orthonormality of the functions $\phi_\rho(\mathbf{x})$ on $p(\mathbf{x})$, we can verify that the eigenfunction property is satisfied

$$T_K[\phi_\rho](\mathbf{x}') = \int p(\mathbf{x}) K(\mathbf{x}, \mathbf{x}') \phi_\rho(\mathbf{x}) d\mathbf{x} = \sum_{\rho'} \eta_{\rho'} \phi_{\rho'}(\mathbf{x}') \int p(\mathbf{x}) \phi_{\rho'}(\mathbf{x}) \phi_\rho(\mathbf{x}') d\mathbf{x} = \eta_\rho \phi_\rho(\mathbf{x}'). \tag{4}$$

As a consequence of the basis property of $\{\phi_\rho\}_{\rho=0}^\infty$, a square integrable function $f(\mathbf{x}) \in L_2(\mathcal{X}, p)$ can be expanded as:

$$f(\mathbf{x}) = \sum_{\rho=0}^\infty a_\rho \phi_\rho(\mathbf{x}) \ , \ a_\rho = \int p(\mathbf{x}) f(\mathbf{x}) \phi_\rho(\mathbf{x}) d\mathbf{x}. \tag{5}$$

By the reproducing property of the kernel, we find the identity $\langle \phi_\rho, \phi_\gamma \rangle_{\mathcal{H}} = \frac{\delta_{\rho,\gamma}}{\eta_\rho}$. Therefore the RKHS norm of a function $f$ is

$$\|f\|_{\mathcal{H}}^2 = \langle f, f \rangle_{\mathcal{H}} = \sum_{\rho, \gamma} a_\rho a_\gamma \langle \phi_\rho, \phi_\gamma \rangle_{\mathcal{H}} = \sum_{\rho=0}^\infty \frac{a_\rho^2}{\eta_\rho}. \tag{6}$$

A function $f$ is said to be a member of the RKHS $\mathcal{H}$ if and only if $\|f\|_{\mathcal{H}}^2 < \infty$.

Given a set of training examples $\mathcal{D} \equiv \{\mathbf{x}^\mu, y^\mu\}|_{\mu=1}^P$, the problem of interest is the minimization of the energy function $H[f; \mathcal{D}]$ with respect to functions $f \in \mathcal{H}$:

$$f^*(\mathbf{x}) = \mathrm{argmin}_{f \in \mathcal{H}} H[f; \mathcal{D}], \qquad H[f; \mathcal{D}] \equiv \frac{1}{2\lambda} \sum_{\mu=1}^P \left(y^\mu - f(\mathbf{x}^\mu)\right)^2 + \frac{1}{2}\|f\|_{\mathcal{H}}^2. \qquad (7)$$

Here, we explicitly denote dataset dependence $\mathcal{D}$. Labels $y^\mu$ are generated from a noisy target function:

$$y^\mu = \bar{f}(\mathbf{x}^\mu) + \epsilon^\mu, \quad \langle \epsilon^\mu, \epsilon^\nu \rangle = \sigma^2 \delta^{\mu\nu}. \qquad (8)$$

We define the null-space indices of the kernel as $\mathcal{N} = \{\rho | \eta_\rho = 0\}$ and its complement as $\mathcal{N}^\perp$. Since the learned function is always in the RKHS (the energy $H$ includes a penalty on $\|f\|_{\mathcal{H}}^2$), it cannot place any power in the null modes indexed by $\mathcal{N}$ and therefore can only optimize over linear combinations of eigenfunctions indexed by $\mathcal{N}^\perp$. We introduce rescaled features $\psi_\rho = \sqrt{\eta_\rho} \phi_\rho$ for all $\rho \in \mathcal{N}^\perp$. The square integrable target function admits the decomposition

$$\bar{f}(\mathbf{x}) = \sum_{\rho \in \mathcal{N}^\perp} \overline{w}_\rho \psi_\rho(\mathbf{x}) + \sum_{\rho \in \mathcal{N}} a_\rho \phi_\rho(\mathbf{x}) = \overline{\mathbf{w}} \cdot \boldsymbol{\Psi}(\mathbf{x}) + \overline{\mathbf{a}} \cdot \boldsymbol{\Phi}(\mathbf{x}), \qquad (9)$$

where we have introduced vectors $\overline{\mathbf{w}}, \overline{\mathbf{a}}, \boldsymbol{\Phi}, \boldsymbol{\Psi}$ of the appropriate dimensions. We stress that the assumption of a square integrable target $\bar{f}$ is identical to the condition that the target possess finite variance under the data density $p(\mathbf{x})$, a very mild assumption which gives our theory significant generality. The kernel regression task reduces to minimization of the energy function over the learned *weights* $w_\rho$:

$$\mathbf{w}^* = \mathrm{argmin}_{\mathbf{w}} H(\mathbf{w}; \mathcal{D}), \quad H(\mathbf{w}; \mathcal{D}) \equiv \frac{1}{2\lambda} \sum_{\mu=1}^P \left(\boldsymbol{\Psi}(\mathbf{x}^\mu) \cdot (\overline{\mathbf{w}} - \mathbf{w}) + \overline{\mathbf{a}} \cdot \boldsymbol{\Phi}(\mathbf{x}^\mu) + \epsilon^\mu\right)^2 + \frac{1}{2}\|\mathbf{w}\|_2^2. \quad (10)$$

where $\overline{\mathbf{a}}$ contains the coefficients of the teacher for null modes indexed by $\mathcal{N}$. Generalization error and training error are defined as:

$$E_g(\mathbf{w}^*, \mathcal{D}) \equiv \left\langle \left(f^*(\mathbf{x}) - \bar{f}(\mathbf{x})\right)^2 \right\rangle_{\mathbf{x}} = (\mathbf{w}^* - \overline{\mathbf{w}})^\top \boldsymbol{\Lambda}(\mathbf{w}^* - \overline{\mathbf{w}}) + \|\overline{\mathbf{a}}\|^2$$

$$E_{tr}(\mathbf{w}^*, \mathcal{D}) \equiv \frac{1}{2P} \sum_{\mu=1}^P \left(\boldsymbol{\Psi}(\mathbf{x}^\mu) \cdot (\overline{\mathbf{w}} - \mathbf{w}^*) + \overline{\mathbf{a}} \cdot \boldsymbol{\Phi}(\mathbf{x}^\mu) + \epsilon^\mu\right)^2, \qquad (11)$$

where we introduced the diagonal matrix of the spectrum $\boldsymbol{\Lambda}_{\rho\gamma} \equiv \eta_\rho \delta_{\rho\gamma}$. The last term in $E_g(\mathcal{D})$, $\|\overline{\mathbf{a}}\|^2$, represents the inability of the kernel to place any power in modes indexed by $\mathcal{N}$ and is thus an irreducible estimation error, present even in the $P \to \infty$ limit, which we denote as $E_g(\infty)$. Our main goal is to average $E_g(\mathcal{D})$ and $E_{tr}(\mathcal{D})$ over all possible realizations of $\mathcal{D}$ of fixed size $P$.

## Supplementary Note 2 - Replica Calculation

We perform replica calculation to find the expected value of training error, estimator and its RKHS norm as well as the generalization error. To set up our statistical mechanics problem, we first introduce the following partition function:

$$Z[J, K, \boldsymbol{\xi}, \boldsymbol{\chi}] = \int d\mathbf{w} \, e^{-\beta H(\mathbf{w}; \mathcal{D}) + J\frac{\beta P}{2} E_g(\mathbf{w}, \mathcal{D}) + K\frac{\beta P}{2\lambda} E_{tr}(\mathbf{w}, \mathcal{D}) + \beta \boldsymbol{\xi} \cdot \mathbf{w} + \frac{\beta}{2}\boldsymbol{\chi}^\top \mathbf{w}\mathbf{w}^\top \boldsymbol{\chi}}, \qquad (12)$$

such that

$$E_g(\mathbf{w}^*, \mathcal{D}) = \lim_{\beta \to \infty} \frac{2}{\beta P} \frac{\partial}{\partial J} \log Z[J, K, \boldsymbol{\xi}, \boldsymbol{\chi}] \Big|_{J,K,\boldsymbol{\xi},\boldsymbol{\chi}=0} + \|\bar{\mathbf{a}}\|^2$$

$$E_{tr}(\mathbf{w}^*, \mathcal{D}) = \lim_{\beta \to \infty} \frac{\lambda}{\beta P} \frac{\partial}{\partial K} \log Z[J, K, \boldsymbol{\xi}, \boldsymbol{\chi}] \Big|_{J,K,\boldsymbol{\xi},\boldsymbol{\chi}=0} \tag{13}$$

since the integral in Supplementary Equation 12 concentrates around the weights $\mathbf{w}^*$ of the kernel regression solution in the $\beta \to \infty$ limit. Last two terms in the exponent of Supplementary Equation 12 is to calculate the expected estimator $f(\mathbf{x})$ and its correlation function via:

$$\sqrt{\eta_\alpha} \, \langle w_\alpha(\mathcal{D}) \rangle_{\mathbf{w}} = \frac{1}{\beta} \frac{\partial}{\partial \xi'_\alpha} \log Z[J, K, \boldsymbol{\xi}, \boldsymbol{\chi}] \Big|_{J,K,\boldsymbol{\xi},\boldsymbol{\chi}=0}$$

$$\sqrt{\eta_\alpha \eta_\beta} \, \langle w_\alpha(\mathcal{D}) w_\beta(\mathcal{D}) \rangle_{\mathbf{w}} = \frac{1}{\beta} \frac{\partial^2}{\partial \chi'_\alpha \partial \chi'_\beta} \log Z[J, K, \boldsymbol{\xi}, \boldsymbol{\chi}] \Big|_{J,K,\boldsymbol{\xi},\boldsymbol{\chi}=0}, \tag{14}$$

where the primed quantities are defined as $\boldsymbol{\xi} = \boldsymbol{\Lambda}^{1/2} \boldsymbol{\xi}'$ and $\boldsymbol{\chi} = \boldsymbol{\Lambda}^{1/2} \boldsymbol{\chi}'$. Hence one can read out:

$$\langle f(\mathbf{x}, \mathcal{D}) \rangle_{\mathbf{w}} = \sum_\alpha \sqrt{\eta_\alpha} \, \langle w_\alpha(\mathcal{D}) \rangle_{\mathbf{w}} \, \phi_\alpha(\mathbf{x})$$

$$\langle f(\mathbf{x}, \mathcal{D}) f(\mathbf{x}', \mathcal{D}) \rangle_{\mathbf{w}} = \sum_{\alpha\beta} \sqrt{\eta_\alpha \eta_\beta} \, \langle w_\alpha(\mathcal{D}) w_\beta(\mathcal{D}) \rangle_{\mathbf{w}} \, \phi_\alpha(\mathbf{x}) \phi_\beta(\mathbf{x}') \tag{15}$$

Here the $\langle .. \rangle_{\mathbf{w}}$ refers to ensemble average over all possible $\mathbf{w}$'s with probability weights given by $e^{-\beta H(\mathbf{w};\mathcal{D})}$. We will eventually consider the $\beta \to \infty$ limit to obtain the quantities above corresponding to the kernel regression solution.

In order to perform the training set average $\langle \mathcal{O}(\mathcal{D}) \rangle_{\mathcal{D}}$ for the quantities above, we must average $\log Z$ over all possible training samples and noises. Resorting to the replica trick, averaging $\log Z$ reduces to averaging n-times replicated partition function $Z^n$:

$$\langle \log Z \rangle_{\mathcal{D}} = \lim_{n \to 0} \frac{\langle Z^n \rangle_{\mathcal{D}} - 1}{n}$$

$$\langle Z^n \rangle_{\mathcal{D}} = e^{-\frac{n\beta}{2}(\bar{\mathbf{W}} - \boldsymbol{\xi})^\top \bar{\mathbf{w}}} \int \left( \prod_{a=1}^n d\mathbf{w}^a \right) e^{-\frac{\beta}{2} \sum_{a=1}^n \mathbf{w}^{a\top}(\mathbf{I} - JP\boldsymbol{\Lambda} - \boldsymbol{\chi}\boldsymbol{\chi}^\top)\mathbf{w}^a - \beta \bar{\mathbf{W}}^\top \sum_{a=1}^n \mathbf{w}^a}$$

$$\times \left\langle e^{-\frac{\beta(1-K)}{2\lambda} \sum_{a=1}^n \left( \mathbf{w}^a \cdot \boldsymbol{\Psi}(\mathbf{x}) + \bar{\mathbf{a}} \cdot \boldsymbol{\Phi}(\mathbf{x}) + \epsilon^a \right)^2} \right\rangle_{\mathbf{x}, \{\epsilon^a\}}^P, \tag{16}$$

where we shifted $\mathbf{w}^a \to \mathbf{w}^a + \bar{\mathbf{w}}$ and defined $\bar{\mathbf{W}} = (\bar{\mathbf{w}} - \boldsymbol{\xi}) - (\boldsymbol{\chi}^\top \bar{\mathbf{w}})\boldsymbol{\chi}$ for notational simplicity. Now we can average over quenched disorder introduced due to the training samples and noise before integrating out the thermal degrees of freedom.

## Averaging over Quenched Disorder

The quantity of interest is the following:

$$\left\langle e^{-\frac{\beta(1-K)}{2\lambda} \sum_{a=1}^n \left( \mathbf{w}^a \cdot \boldsymbol{\Psi}(\mathbf{x}) + \bar{\mathbf{a}} \cdot \boldsymbol{\Phi}(\mathbf{x}) + \epsilon^a \right)^2} \right\rangle_{\mathbf{x}, \{\epsilon^a\}}. \tag{17}$$

Rather than integrating over $\mathbf{x}$, we integrate over $q^a = \mathbf{w}^a \cdot \boldsymbol{\Psi}(\mathbf{x}) + \bar{\mathbf{a}} \cdot \boldsymbol{\Phi}(\mathbf{x}) + \epsilon^a$, which is itself a random variable with mean and covariance:

$$\boldsymbol{\mu}^a \equiv \langle q^a \rangle = \langle \mathbf{w}^a \cdot \boldsymbol{\Psi}(\mathbf{x}) \rangle + \langle \bar{\mathbf{a}} \cdot \boldsymbol{\Phi}(\mathbf{x}) \rangle + \langle \epsilon^a \rangle = \sqrt{\eta_0} w_0^a,$$

$$\mathbf{C}^{ab} \equiv \langle q^a q^b \rangle = \mathbf{w}^{a\top} \langle \boldsymbol{\Psi}(\mathbf{x}) \boldsymbol{\Psi}(\mathbf{x})^\top \rangle \mathbf{w}^b + \bar{\mathbf{a}}^\top \left\langle \boldsymbol{\Phi}(\mathbf{x}) \boldsymbol{\Phi}(\mathbf{x})^\top \right\rangle \bar{\mathbf{a}} + \langle \epsilon^a \epsilon^b \rangle$$

$$= \mathbf{w}^{a\top} \boldsymbol{\Lambda} \mathbf{w}^b + \boldsymbol{\Sigma}^{ab}, \tag{18}$$

where $\boldsymbol{\Sigma} = \left( \|\bar{\mathbf{a}}\|^2 + \sigma^2 \right) \mathbf{1}\mathbf{1}^\top$ is the covariance matrix of noise across replicas. Computation of this mean and covariance relied on the orthogonality of the eigenfunctions. Notice that the target modes on the null space of the kernel act as noise on training labels, quantifying the effect of out-of-RKHS target functions. From here on, we take $\|\bar{\mathbf{a}}\|^2 = 0$ for simplicity. At the end of the calculation it can be recovered by shifting $\sigma^2 \to \sigma^2 + \|\bar{\mathbf{a}}\|^2$ and $E_g \to E_g + \|\bar{\mathbf{a}}\|^2$. Note that the noise-free part of the diagonal elements represents the generalization error in a single replica i.e. $\mathbf{C}^{aa} = E_g^a + \sigma^2$, while off-diagonal elements give the overlap of the weights across different replicas. In the limit $\beta \to \infty$, we expect these two quantities to be equal as the optimal weights averaged over training samples across different replicas will be the same due to the convexity of the problem.

First, we omit the zero mode $\mu$ by considering only kernels with zero constant terms (such that $\eta_0 = 0$), however it is straightforward to include it in the rest of the calculation and we give the contribution from zero mode in our final expression. Next, by observing that $q^a$ is a summation of many uncorrelated random variables ($\langle \psi_\rho(\mathbf{x}) \psi_{\rho'}(\mathbf{x}) \rangle_{\mathbf{x} \sim p(\mathbf{x})} = \eta_\rho \delta_{\rho\rho'}$) and a Gaussian noise, we approximate the probability distribution of $q^a$ by a multivariate Gaussian with its mean and covariance given by Supplementary Equation 18:

$$P(\{q^a\}) = \frac{1}{\sqrt{(2\pi)^n \det(\mathbf{C})}} \exp \left( -\frac{1}{2} \sum_{a,b} q^a (\mathbf{C}^{ab})^{-1} q^b \right). \tag{19}$$

This approximation is further validated with the excellent match of our theory to simulations. Then the average over quenched disorder reduces to:

$$\left\langle e^{-\frac{\beta(1-K)}{2\lambda} \sum_{a=1}^n \left( (\mathbf{w}^a - \bar{\mathbf{w}}) \cdot \boldsymbol{\Psi}(\mathbf{x}) + \epsilon^a \right)^2} \right\rangle_{\mathbf{x}, \{\epsilon^a\}} \approx \int \{dq^a\} P(\{q^a\}) \exp \left( -\frac{\beta_K}{2\lambda} \sum_{a=1}^n (q^a)^2 \right)$$

$$= \exp \left( -\frac{1}{2} \log \det \left( \mathbf{I} + \frac{\beta_K}{\lambda} \mathbf{C} \right) \right), \tag{20}$$

where we defined $\beta_K \equiv \beta(1-K)$ for notational convenience. Combining everything together, the dataset averaged replicated partition function becomes:

$$\langle Z^n \rangle_{\mathcal{D}} = e^{-\frac{n\beta}{2}(\bar{\mathbf{W}} - \boldsymbol{\xi})^\top \bar{\mathbf{w}}} \int \left( \prod_{a=1}^n d\mathbf{w}^a \right) e^{-\frac{\beta}{2} \sum_{a=1}^n \mathbf{w}^{a\top} (\mathbf{I} - JP\boldsymbol{\Lambda} - \boldsymbol{\chi}\boldsymbol{\chi}^\top) \mathbf{w}^a - \beta \bar{\mathbf{W}}^\top \sum_{a=1}^n \mathbf{w}^a}$$

$$\times e^{-\frac{P}{2} \log \det \left( \mathbf{I} + \frac{\beta_K}{\lambda} \mathbf{C} \right)} \tag{21}$$

Using the definitions Supplementary Equation 18, we insert conjugate variables through the following identity:

$$1 = \left( \frac{iP}{2\pi} \right)^{\frac{n(n+1)}{2}} \int \left( \prod_{a \geq b} d\mathbf{C}^{ab} d\hat{\mathbf{C}}^{ab} \right) \exp \left[ -P \sum_{a \geq b} \hat{\mathbf{C}}^{ab} \left( \mathbf{C}^{ab} - \mathbf{w}^{a\top} \boldsymbol{\Lambda} \mathbf{w}^b - \boldsymbol{\Sigma}^{ab} \right) \right] \tag{22}$$

5

Here, integral over $\hat{\mathbf{C}}$ runs over the imaginary axis and we explicitly scaled it by $P$. Then defining:

$$G_E = \frac{1}{2}\log\det\left(\mathbf{I} + \frac{\beta_K}{\lambda}\mathbf{C}\right) \tag{23}$$

$$e^{-G_S} = \int\left(\prod_{a=1}^{n}d\mathbf{w}^a\right)\exp\left(-\frac{\beta}{2}\sum_{a\geq b}\mathbf{w}^{a\top}\left(\left(\mathbf{I} - JP\mathbf{\Lambda} - \mathbf{\chi}\mathbf{\chi}^{\top}\right)\mathbf{I}^{ab} - \frac{2P}{\beta}\mathbf{\Lambda}\hat{\mathbf{C}}^{ab}\right)\mathbf{w}^b - \beta\bar{\mathbf{W}}^{\top}\sum_{a=1}^{n}\mathbf{w}^a\right),$$

we obtain:

$$\langle Z^n\rangle = e^{\frac{n(n+1)}{2}\log\left(\frac{iP}{2\pi}\right) - \frac{n\beta}{2}(\bar{\mathbf{W}} - \boldsymbol{\xi})^{\top}\bar{\mathbf{w}}}\int\left(\prod_{a\geq b}d\mathbf{C}^{ab}d\hat{\mathbf{C}}^{ab}\right)\exp\left[-P\sum_{a\geq b}\hat{\mathbf{C}}^{ab}(\mathbf{C}^{ab} - \mathbf{\Sigma}^{ab}) - PG_E - G_S\right]. \tag{24}$$

Therefore, we only need to evaluate the integral in $G_S$. First we would like to express the ordered sum $\sum_{a\geq b}\mathbf{w}^{a\top}\left(\left(\mathbf{I} - JP\mathbf{\Lambda} - \mathbf{\chi}\mathbf{\chi}^{\top}\right)\mathbf{I}^{ab} - \frac{2P}{\beta}\mathbf{\Lambda}\hat{\mathbf{C}}^{ab}\right)\mathbf{w}^b$ as an unordered sum over $a, b$. Note that

$$\sum_{a,b}\mathbf{w}^{a\top}\left(\left(\mathbf{I} - JP\mathbf{\Lambda} - \mathbf{\chi}\mathbf{\chi}^{\top}\right)\mathbf{I}^{ab} - \frac{2P}{\beta}\mathbf{\Lambda}\hat{\mathbf{C}}^{ab}\right)\mathbf{w}^b$$

$$= 2\sum_{a\geq b}\mathbf{w}^{a\top}\left(\left(\mathbf{I} - JP\mathbf{\Lambda} - \mathbf{\chi}\mathbf{\chi}^{\top}\right)\mathbf{I}^{ab} - \frac{2P}{\beta}\mathbf{\Lambda}\hat{\mathbf{C}}^{ab}\right)\mathbf{w}^b - \sum_{a,b}\mathbf{w}^{a\top}\left(\left(\mathbf{I} - JP\mathbf{\Lambda} - \mathbf{\chi}\mathbf{\chi}^{\top}\right)\mathbf{I}^{ab} - \frac{2P}{\beta}\mathbf{\Lambda}\mathrm{diag}(\hat{\mathbf{C}})^{ab}\right)\mathbf{w}^b$$

Hence, we obtain:

$$\sum_{a\geq b}\mathbf{w}^{a\top}\left(\left(\mathbf{I} - JP\mathbf{\Lambda} - \mathbf{\chi}\mathbf{\chi}^{\top}\right)\mathbf{I}^{ab} - \frac{2P}{\beta}\mathbf{\Lambda}\hat{\mathbf{C}}^{ab}\right)\mathbf{w}^b = \sum_{a,b}\mathbf{w}^{a\top}\mathbf{X}^{ab}\mathbf{w}^b,$$

where we defined:

$$\mathbf{X}^{ab} = \left(\mathbf{I} - JP\mathbf{\Lambda} - \mathbf{\chi}\mathbf{\chi}^{\top}\right)\mathbf{I}^{ab} - \frac{P}{\beta}\mathbf{\Lambda}\left(\hat{\mathbf{C}} + \mathrm{diag}(\hat{\mathbf{C}})\right)^{ab}. \tag{25}$$

In order to evaluate the Gaussian integral, we will cast the function and target weights into an $nM$ dimensional vector:

$$\mathbf{w} = \begin{pmatrix}\mathbf{w}^1 & \mathbf{w}^2 & .. & \mathbf{w}^a & .. & \mathbf{w}^n\end{pmatrix}_{nM\times 1}$$

$$\bar{\mathbf{W}}_{\otimes n} = \begin{pmatrix}\bar{\mathbf{W}} & \bar{\mathbf{W}} & \ldots & \bar{\mathbf{W}}\end{pmatrix}_{nM\times 1} \tag{26}$$

Furthermore, we introduce the $nM \times nM$ matrix $\mathbf{X}$ as:

$$\mathbf{X} = \begin{pmatrix}\mathbf{X}^{11} & \mathbf{X}^{12} & \ldots & \ldots & \mathbf{X}^{1n}\\ \mathbf{X}^{21} & \mathbf{X}^{22} & \ldots & \ldots & \mathbf{X}^{2n}\\ \vdots & \vdots & \ddots & & \vdots\\ \vdots & \ldots & \mathbf{X}^{ab} & \ddots & \vdots\\ \mathbf{X}^{n1} & \ldots & \ldots & \ldots & \mathbf{X}^{nn}\end{pmatrix}_{nM\times nM} \tag{27}$$

Finally we denote the integration measure as $\mathcal{D}\mathbf{w} = \prod_{a,\rho} dw^a_\rho$. With these definitions, $G_S$ becomes:

$$e^{-G_S} = \int \mathcal{D}\mathbf{w}\, e^{-\frac{\beta}{2}\mathbf{w}^\top \mathbf{X}\mathbf{w} - \beta \bar{\mathbf{W}}^\top_{\otimes n}\mathbf{w}} \tag{28}$$

Hence, we turned the integral in $G_S$ to a simple Gaussian integral. The result is:

$$e^{-G_S} = \left(\frac{2\pi}{\beta}\right)^{\frac{nM}{2}} \left(\det \mathbf{X}\right)^{-\frac{1}{2}} \exp\left(\frac{\beta}{2}\bar{\mathbf{W}}^\top_{\otimes n}\mathbf{X}^{-1}\bar{\mathbf{W}}_{\otimes n}\right). \tag{29}$$

Now the integral in Supplementary Equation 24 can be evaluated using the method of steepest descent. In Supplementary Equation 24, we see that all the terms in the exponent are $\mathcal{O}(n)$. Furthermore, we will use $P$ as the saddle point parameter going to infinity with a proper scaling. Therefore, defining the following function:

$$S[\mathbf{C}, \hat{\mathbf{C}}, \boldsymbol{\mu}, \hat{\boldsymbol{\mu}}] = \frac{1}{n}\sum_{a \geq b} \hat{\mathbf{C}}^{ab}(\mathbf{C}^{ab} - \boldsymbol{\Sigma}^{ab}) + \frac{1}{nP}\left(PG_E + G_S + \frac{n\beta}{2}(\bar{\mathbf{W}} - \boldsymbol{\xi})^\top \bar{\mathbf{w}}\right)$$

$$G_E = \frac{1}{2}\log\det\left(\mathbf{I} + \frac{\beta_K}{\lambda}\mathbf{C}\right)$$

$$G_S = \frac{1}{2}\log\det \mathbf{X} - \frac{\beta}{2}\bar{\mathbf{W}}^\top_{\otimes n}\mathbf{X}^{-1}\bar{\mathbf{W}}_{\otimes n}, \tag{30}$$

we obtain:

$$\langle \log Z \rangle = \lim_{n \to 0} \frac{1}{n}\left(\langle Z^n \rangle - 1\right),$$

$$\langle Z^n \rangle = e^{\frac{n(n+1)}{2}\log\left(\frac{iP}{2\pi}\right) + \frac{nM}{2}\log\frac{2\pi}{\beta}} \int \left(\prod_{a \geq b} d\mathbf{C}^{ab}d\hat{\mathbf{C}}^{ab}\right)e^{-nPS[\mathbf{C}, \hat{\mathbf{C}}]}. \tag{31}$$

The reader may question the dependence of various quantities in $S$ on $P$, since we are taking a $P \to \infty$ limit. This is because we want to keep our treatment general. Depending on the kernel and data distribution, there are other quantities here that can scale with $P$. Specific examples will be given.

## Replica Symmetry and Saddle Point Equations

In order to proceed with the saddle point integration, we further assume replica symmetry relying on the convexity of the problem:

$$C^0 = \mathbf{C}^{aa}, \qquad\qquad\qquad \hat{C}^0 = \hat{\mathbf{C}}^{aa},$$
$$C = \mathbf{C}^{a \neq b}, \qquad\qquad\qquad \hat{C} = \hat{\mathbf{C}}^{a \neq b}. \tag{32}$$

Therefore, we have $\mathbf{C} = (C_0 - C)\mathbf{I} + C\mathbf{1}\mathbf{1}^\top$ and $\hat{\mathbf{C}} = (\hat{C}_0 - \hat{C})\mathbf{I} + \hat{C}\mathbf{1}\mathbf{1}^\top$. In this case, the matrix $\mathbf{X}$ has the form:

$$\mathbf{X} = \begin{pmatrix} \mathbf{X}_0 & \mathbf{X}_1 & \mathbf{X}_1 & \dots & \mathbf{X}_1 \\ \mathbf{X}_1 & \mathbf{X}_0 & \mathbf{X}_1 & \dots & \mathbf{X}_1 \\ \mathbf{X}_1 & \mathbf{X}_1 & \mathbf{X}_0 & \dots & \vdots \\ \vdots & \dots & \dots & \ddots & \vdots \\ \mathbf{X}_1 & \dots & \dots & \dots & \mathbf{X}_0 \end{pmatrix} = \mathbf{I}_{n \times n} \otimes (\mathbf{X}_0 - \mathbf{X}_1)_{M \times M} + \mathbf{1}_{n \times n} \otimes (\mathbf{X}_1)_{M \times M}, \tag{33}$$

where:

$$\mathbf{X}_0 \equiv \mathbf{X}^{aa} = \left(\mathbf{I} - JP\mathbf{\Lambda} - \chi\chi^\top\right) - \frac{2P\hat{C}_0}{\beta}\mathbf{\Lambda}$$

$$\mathbf{X}_1 \equiv \mathbf{X}^{a\neq b} = -\frac{P\hat{C}}{\beta}\mathbf{\Lambda}.$$

It is straightforward to calculate the inverse of this matrix using Sherman-Morrison-Woodbury formula $(A + B)^{-1} = A^{-1} - A^{-1}BA^{-1}(I + BA^{-1})^{-1}$:

$$\mathbf{X}^{-1} = \mathbf{I}_n \otimes (\mathbf{X}_0 - \mathbf{X}_1)^{-1} - \left(\mathbf{1}_n \otimes (\mathbf{X}_0 - \mathbf{X}_1)^{-1}\mathbf{X}_1(\mathbf{X}_0 - \mathbf{X}_1)^{-1}\right)\left(\mathbf{I}_n \otimes \mathbf{I}_M + \mathbf{1}_n \otimes \mathbf{X}_1(\mathbf{X}_0 - \mathbf{X}_1)^{-1}\right)^{-1}$$
$$= \mathbf{I}_n \otimes \mathbf{G}^{-1} - \mathbf{1}_n \otimes \mathbf{G}^{-1}\mathbf{X}_1\mathbf{G}^{-1} + \mathcal{O}(n),$$

where we defined,

$$\mathbf{G} \equiv \mathbf{X}_0 - \mathbf{X}_1 = \mathbf{I} - \frac{P(2\hat{C}_0 - \hat{C}) + J\beta P}{\beta}\mathbf{\Lambda} - \chi\chi^\top, \tag{34}$$

for shorthand notation. Hence, we get:

$$\bar{\mathbf{W}}_{\otimes n}^\top \mathbf{X}^{-1}\bar{\mathbf{W}}_{\otimes n} = n\bar{\mathbf{W}}^\top \mathbf{G}^{-1}\bar{\mathbf{W}} + \mathcal{O}(n^2). \tag{35}$$

We also need to calculate the determinant of this matrix which can be done by using Gaussian elimination method to bring it into a block-triangular form. The result is:

$$\det \mathbf{X} = \det(\mathbf{X}_0 - \mathbf{X}_1)^{n-1}\det\left(\mathbf{X}_0 + (n-1)\mathbf{X}_1\right) = \det(\mathbf{X}_0 - \mathbf{X}_1)^{n-1}\det(\mathbf{X}_0 - \mathbf{X}_1 + n\mathbf{X}_1). \tag{36}$$

Taylor expanding the last term using $\det(\mathbf{I} + n\mathbf{C}) = 1 + n\operatorname{Tr}\mathbf{C} + \mathcal{O}(n^2)$, we obtain:

$$\log\det \mathbf{X} = n\log\det \mathbf{G} + n\operatorname{Tr}\left(\mathbf{X}_1\mathbf{G}^{-1}\right) + \mathcal{O}(n^2) = n\log\det \mathbf{G} - n\frac{P\hat{C}}{\beta}\operatorname{Tr}\mathbf{\Lambda}\mathbf{G}^{-1} + \mathcal{O}(n^2). \tag{37}$$

Next, using the matrix determinant lemma $\det\left(A + uv^T\right) = \det(A)(1 + v^TA^{-1}u)$, we obtain:

$$\det\left(\mathbf{I} + \frac{\beta_K}{\lambda}\mathbf{C}\right) = \left[1 + \frac{\beta_K}{\lambda}(C_0 - C)\right]^n\left(1 + n\frac{\beta_K C}{\lambda + \beta_K(C_0 - C)}\right),$$
$$\Rightarrow \log\det\left(\mathbf{I} + \frac{\beta_K}{\lambda}\mathbf{C}\right) = n\log\left(1 + \frac{\beta(1-K)}{\lambda}(C_0 - C)\right) + n\frac{\beta(1-K)C}{\lambda + \beta(1-K)(C_0 - C)}, \tag{38}$$

where we recovered $\beta_K = \beta(1 - K)$. Finally, we need to simplify $\sum_{a\geq b}\hat{\mathbf{C}}^{ab}(\mathbf{C}^{ab} - \mathbf{\Sigma}^{ab})$ under the replica symmetry up to leading order in $n$:

$$\sum_{a\geq b}\hat{\mathbf{C}}^{ab}(\mathbf{C}^{ab} - \mathbf{\Sigma}^{ab}) = n\left(\hat{C}_0(C_0 - \sigma^2) - \frac{1}{2}\hat{C}(C - \sigma^2)\right). \tag{39}$$

Therefore, under replica symmetry, the function $S$ given in Supplementary Equation 30 simplifies to:

$$S[\mathbf{C}, \hat{\mathbf{C}}] = \hat{C}_0(C_0 - \sigma^2) - \frac{1}{2}\hat{C}(C - \sigma^2) + \frac{1}{2}\log\left(1 + \frac{\beta(1-K)}{\lambda}(C_0 - C)\right) + \frac{1}{2}\frac{\beta(1-K)C}{\lambda + \beta(1-K)(C_0 - C)}$$
$$+ \frac{1}{2P}\left(\log\det \mathbf{G} - \frac{P\hat{C}}{\beta}\operatorname{Tr}\mathbf{\Lambda}\mathbf{G}^{-1}\right) - \frac{\beta}{2P}\bar{\mathbf{W}}^\top \mathbf{G}^{-1}\bar{\mathbf{W}} + \frac{\beta}{2P}(\bar{\mathbf{W}} - \boldsymbol{\xi})^\top\bar{\mathbf{w}}, \tag{40}$$

8

where we recall that $\mathbf{G} = \mathbf{I} - \frac{P(2\hat{C}_0 - \hat{C}) + J\beta P}{\beta}\mathbf{\Lambda} - \boldsymbol{\chi}\boldsymbol{\chi}^\top$. The saddle point equations of $S$ with respect to $C_0$ and $C$ are simple:

$$\frac{\partial S}{\partial C} = 0 \Rightarrow \boxed{\hat{C} = \frac{\beta^2(1-K)^2 C}{\left(\lambda + \beta(1-K)(C_0 - C)\right)^2}},$$

$$\frac{\partial S}{\partial C_0} = 0 \Rightarrow \boxed{\hat{C}_0 = \frac{1}{2}\hat{C} - \frac{1}{2}\frac{\beta(1-K)}{\lambda + \beta(1-K)(C_0 - C)}}. \tag{41}$$

The equation $\partial S/\partial \hat{C} = 0$ yields:

$$C = \frac{P\hat{C}}{\beta^2}\operatorname{Tr}\mathbf{\Lambda}\mathbf{G}^{-1}\mathbf{\Lambda}\mathbf{G}^{-1} + \bar{\mathbf{W}}^\top\mathbf{G}^{-1}\mathbf{\Lambda}\mathbf{G}^{-1}\bar{\mathbf{W}} + \sigma^2, \tag{42}$$

and the equation $\partial S/\partial \hat{C}_0 = 0$ yields:

$$C_0 = C + \frac{1}{\beta}\operatorname{Tr}\mathbf{\Lambda}\mathbf{G}^{-1}. \tag{43}$$

Two commonly appearing forms are:

$$\kappa \equiv \lambda + \beta(1-K)(C_0 - C) = \lambda + (1-K)\operatorname{Tr}\mathbf{\Lambda}\mathbf{G}^{-1},$$

$$\frac{2\hat{C}_0 - \hat{C}}{\beta} = -\frac{(1-K)}{\lambda + \beta(1-K)(C_0 - C)} = -\frac{(1-K)}{\kappa}. \tag{44}$$

Plugging second equation to the expression for $\mathbf{G}$, we get:

$$\mathbf{G} = \mathbf{I} + \frac{P}{\kappa}(1 - K - J\kappa)\mathbf{\Lambda} - \boldsymbol{\chi}\boldsymbol{\chi}^\top, \tag{45}$$

hence we obtain the following implicit equation:

$$\kappa = \lambda + (1-K)\operatorname{Tr}\mathbf{\Lambda}\left(\mathbf{I} + \frac{P}{\kappa}(1 - K - J\kappa)\mathbf{\Lambda} - \boldsymbol{\chi}\boldsymbol{\chi}^\top\right)^{-1}. \tag{46}$$

In terms of $\kappa$, final saddle point equations reduce to:

$$\hat{C}_0^* = \frac{1}{2}\hat{C}^* - \frac{1}{2}\frac{\beta(1-K)}{\kappa},$$

$$\hat{C}^* = \frac{\beta^2(1-K)^2 C^*}{\kappa^2},$$

$$C_0^* = C^* + \frac{\kappa - \lambda}{(1-K)\beta},$$

$$C^* = C^*(1-K)^2\frac{P}{\kappa^2}\operatorname{Tr}\mathbf{\Lambda}\mathbf{G}^{-1}\mathbf{\Lambda}\mathbf{G}^{-1} + \bar{\mathbf{W}}^\top\mathbf{G}^{-1}\mathbf{\Lambda}\mathbf{G}^{-1}\bar{\mathbf{W}} + \sigma^2. \tag{47}$$

Here * indicates the quantities that give the saddle point. Finally, solving for $C^*$ in the last equation, we obtain:

$$C^* = \frac{1}{1 - (1-K)^2\frac{P}{\kappa^2}\operatorname{Tr}\mathbf{\Lambda}\mathbf{G}^{-1}\mathbf{\Lambda}\mathbf{G}^{-1}}\left(\bar{\mathbf{W}}^\top\mathbf{G}^{-1}\mathbf{\Lambda}\mathbf{G}^{-1}\bar{\mathbf{W}} + \sigma^2\right). \tag{48}$$

Having obtained the saddle points, we can evaluate the saddle point integral. In the limit $P \to \infty$, the dominant contribution is:

$$\langle Z^n \rangle \approx e^{-nPS[\mathbf{C}^*, \hat{\mathbf{C}}^*]}. \tag{49}$$

Taking the $n \to 0$ limit and plugging in the saddle point solutions to the expression Supplementary Equation 40, we obtain the free energy $\langle \log Z \rangle = -PS$ to be:

$$\langle \log Z \rangle = \frac{P}{2} \frac{\kappa - \lambda}{\kappa} - \frac{P}{2} \log \frac{\kappa}{\lambda} - \frac{P}{2} \log \det \mathbf{G} - \frac{\beta P (1 - K)}{2} \frac{\sigma^2}{\kappa} + \frac{\beta}{2} \bar{\mathbf{W}}^\top \mathbf{G}^{-1} \bar{\mathbf{W}} - \frac{\beta}{2} (\bar{\mathbf{W}} - \boldsymbol{\xi})^\top \bar{\mathbf{w}},$$

$$\kappa = \lambda + (1 - K) \operatorname{Tr} \boldsymbol{\Lambda} \mathbf{G}^{-1},$$

$$\mathbf{G} = \mathbf{I} + \frac{P}{\kappa} (1 - K - J\kappa) \boldsymbol{\Lambda} - \boldsymbol{\chi} \boldsymbol{\chi}^\top,$$

$$\bar{\mathbf{W}} = (\bar{\mathbf{w}} - \boldsymbol{\xi}) - (\boldsymbol{\chi}^\top \bar{\mathbf{w}}) \boldsymbol{\chi}. \tag{50}$$

## Generalization Error

Now, we can calculate $E_g = \lim_{\beta \to \infty} \frac{2}{\beta P} \frac{\partial}{\partial J} \langle \log Z \rangle |_{J, K, \boldsymbol{\xi}, \boldsymbol{\chi} = 0}$. Recall that $\kappa$ is itself a function of $J$. Explicit calculation and $\beta \to \infty$ limit yields:

$$E_g = \left( \partial_J \kappa(0) + \kappa^2(0) \right) \sum_{\rho=1}^{N} \frac{\eta_\rho \bar{w}_\rho^2}{\left( \kappa(0) + P\eta_\rho \right)^2} + \sigma^2 \frac{\partial_J \kappa(0)}{\kappa^2(0)}, \tag{51}$$

where

$$\frac{\partial_J \kappa(0)}{\kappa^2(0)} = \frac{1}{1 - \sum_\rho \frac{P\eta_\rho^2}{(\kappa + P\eta_\rho)^2}} \sum_\rho \frac{P\eta_\rho^2}{(\kappa + P\eta_\rho)^2} \equiv \frac{\gamma}{1 - \gamma},$$

$$\kappa \equiv \kappa(0) = \lambda + \sum_\rho \frac{\kappa \eta_\rho}{P\eta_\rho + \kappa}, \tag{52}$$

where we defined $\gamma = \sum_\rho \frac{P\eta_\rho^2}{(\kappa + P\eta_\rho)^2}$. In terms of these quantities, averaged generalization error becomes:

$$\boxed{ E_g = \frac{1}{1 - \gamma} \sum_{\rho=1}^{N} \frac{\eta_\rho}{(\kappa + P\eta_\rho)^2} \left( \kappa^2 \bar{w}_\rho^2 + (\sigma^2 + \|\bar{\mathbf{a}}\|^2) P\eta_\rho \right) + \|\bar{\mathbf{a}}\|^2 + \frac{1 + \gamma}{1 - \gamma} \kappa^2 \frac{\eta_0 \bar{w}_0^2}{(\kappa + 2P\eta_0)^2}. } \tag{53}$$

For completeness, we recovered $\|\bar{\mathbf{a}}\|^2 \neq 0$ (Supplementary Note 1) to account for the case where target function $\bar{f}(\mathbf{x}) = \bar{\mathbf{w}} \cdot \boldsymbol{\Psi}(\mathbf{x}) + \sum_{\rho \in \mathcal{N}} a_\rho \Phi_\rho(\mathbf{x})$ has components out-of-RKHS. The last term accounts for the contribution to $E_g$ from the zero mode when $\eta_0 \neq 0$. Note that at this point we have already taken the $P \to \infty$ limit. $P$ still appears in these expressions to consider different scaling limits for kernel eigenvalues.

This expression generalizes Eq. (4) in main text to cases where the target function has modes not expressible by the kernel. These modes contribute an irreducible component to the generalization error, as well as act as noise on expressible modes. Based on this expression, we can define another effective noise $\tilde{\sigma} = \sigma^2 + \|\bar{\mathbf{a}}\|^2$, which can lead to non-monotonic learning curves even in the absence of label noise ($\sigma = 0$). An example of this phenomenon is provided in Supplementary Figure 2.

## Training Error

Next, we calculate the expected training error defined as:

$$E_{tr} = \lim_{\beta \to \infty} \frac{\lambda}{\beta P} \frac{\partial}{\partial K} \langle \log Z \rangle \bigg|_{J,K,\boldsymbol{\xi},\boldsymbol{\chi}=0}.$$ (54)

First we shall calculate:

$$\partial_K \kappa(0) = -\frac{\kappa^2 \delta}{1 - \gamma}, \quad \kappa = \lambda + \kappa \sum_\rho \frac{\eta_\rho}{\kappa + P\eta_\rho}, \quad \gamma = \sum_\rho \frac{P\eta_\rho^2}{(\kappa + P\eta_\rho)^2}, \quad \delta = \sum_\rho \frac{\eta_\rho}{(\kappa + P\eta_\rho)^2}$$ (55)

A cumbersome calculation yields:

$$\frac{\partial}{\partial K} \langle \log Z \rangle \big|_{J,K,\boldsymbol{\xi}=0} = -\frac{P\beta}{2} \left( \frac{\kappa\delta + \gamma - 1}{\kappa} \right) \left( E_g(P) + \sigma^2 \right) + \frac{P}{2} (\kappa\delta + \gamma),$$ (56)

where $E_g$ is defined above Supplementary Equation 53. By using the summation forms of $\kappa$, $\gamma$ and $\delta$, one can show that:

$$\kappa\delta + \gamma = 1 - \frac{\lambda}{\kappa}.$$ (57)

Plugging this back in, we obtain:

$$\frac{\partial}{\partial K} \langle \log Z \rangle \big|_{J,K,\boldsymbol{\xi}=0} = \frac{P\beta}{2} \frac{\lambda}{\kappa^2} \left( E_g(P) + \sigma^2 \right) + \frac{P}{2} \left( 1 - \frac{\lambda}{\kappa} \right).$$ (58)

Hence average case training error becomes:

$$E_{tr}(P) = \lim_{\beta \to \infty} \frac{\lambda}{\beta P} \frac{\partial}{\partial K} \langle \log Z \rangle \big|_{J,K,\boldsymbol{\xi},\boldsymbol{\chi}=0} = \frac{\lambda^2}{\kappa^2} \left( E_g(P) + \sigma^2 \right).$$ (59)

This is consistent with the expectation that in the $\beta \to \infty$ limit, kernel machine can fit all training samples perfectly when $\lambda = 0$. We also note that the same relation between training and generalization errors were also obtained in other works [4, 5] using random matrix theory.

## Expected Estimator and the Correlation Function

Finally, we calculate the RKHS weights of the expected function and its variance:

$$\sqrt{\eta_\alpha} \langle w_\alpha^* \rangle_{\mathcal{D}} = \lim_{\beta \to \infty} \frac{1}{\beta} \frac{\partial}{\partial \xi_\alpha'} \langle \log Z \rangle \bigg|_{J,K,\boldsymbol{\xi},\boldsymbol{\chi}=0}$$

$$\sqrt{\eta_\alpha \eta_\beta} \langle w_\alpha^* w_\beta^* \rangle_{\mathcal{D}} = \lim_{\beta \to \infty} \frac{1}{\beta} \frac{\partial^2}{\partial \chi_\alpha' \partial \chi_\beta'} \langle \log Z \rangle \bigg|_{J,K,\boldsymbol{\xi},\boldsymbol{\chi}=0},$$ (60)

where the derivatives are with respect to $\xi_\alpha' = \xi_\alpha / \sqrt{\eta_\alpha}$ and $\chi_\alpha' = \chi_\alpha / \sqrt{\eta_\alpha}$, respectively. Taking derivatives for each entry of $\boldsymbol{\xi}'$, we obtain:

$$\frac{1}{\beta} \frac{\partial}{\partial \xi_\alpha'} \langle \log Z \rangle = \sqrt{\eta_\alpha} \bar{w}_\alpha - \frac{\kappa \sqrt{\eta_\alpha} (\bar{w}_\alpha - \xi_\alpha)}{P\eta_\alpha + \kappa} = \frac{P\eta_\alpha \sqrt{\eta_\alpha} \bar{w}_\alpha + \kappa \sqrt{\eta_\alpha} \xi_\alpha}{P\eta_\alpha + \kappa}.$$ (61)

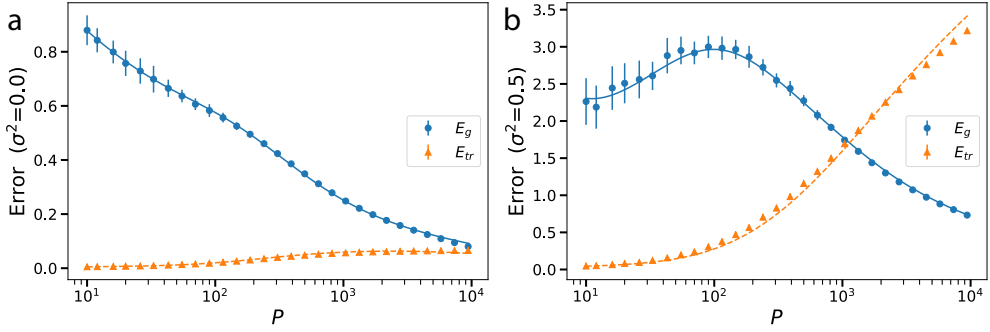

Supplementary Figure 1: Kernel regression with Gaussian RBF $K(\mathbf{x}, \mathbf{x}') = e^{-\frac{1}{2D\omega^2}||x-x'||^2}$ for MNIST with kernel bandwidth $\omega = 0.1$, ridge parameter $\lambda = 0.01$ and noise variance $\sigma^2 = 0, 0.5$. Theoretical predictions for $E_g$ (solid blue line) and $E_{tr}$ (dashed orange line) are in excellent agreement with experiment (blue dots and orange triangles, respectively). Error bars represent standard deviation over 30 trials. **a.** When no label noise is present, $E_{tr}$ is always smaller than $E_g$. **b.** When noise is non-zero, $E_{tr}$ may exceed $E_g$ due to explicit presence of label noise in the loss function.

Hence the *average estimator* has the following form:

$$\langle f^*(\mathbf{x}; P) \rangle_{\mathcal{D}} = \sum_\rho \frac{P\eta_\rho}{P\eta_\rho + \kappa} \bar{w}_\rho \psi_\rho(\mathbf{x}), \tag{62}$$

which approaches to the target function as $P \to \infty$. Note that the learned function can only express the components which span the RKHS. If the target function has out-of-RKHS components, those will never be learned.

A related result was given in [6] which used a similar technique to calculate the expected estimator. With our notation, their perturbative result for uniform spherical datasets and dot product kernels reads (Eq. (24) in [6]):

$$\langle f^*(\mathbf{x}; P) \rangle_{\mathcal{D}} = \sum_\rho \left( \frac{P\eta_\rho}{P\eta_\rho + \lambda} - \frac{P\eta_\rho}{(P\eta_\rho + \lambda)^2} \lambda \sum_\rho \frac{\eta_\rho}{P\eta_\rho + \lambda} + \dots \right) \bar{w}_\rho \psi_\rho(\mathbf{x}), \tag{63}$$

where $\lambda$ is the ridge parameter and $\dots$ represent the higher order corrections in their perturbative setting. We obtain the same result result if we expand our expression for $\langle f^*(\mathbf{x}; P) \rangle_{\mathcal{D}}$ in a power series of $\epsilon = (\kappa - \lambda)$ assuming $\epsilon$ is small, and use $\kappa - \lambda = \kappa \sum_\rho \frac{\eta_\rho}{P\eta_\rho + \kappa}$:

$$\frac{P\eta_\rho}{P\eta_\rho + \kappa} = \frac{P\eta_\rho}{P\eta_\rho + \lambda} - \frac{P\eta_\rho}{(P\eta_\rho + \lambda)^2} \lambda \sum_\rho \frac{\eta_\rho}{P\eta_\rho + \lambda} + \dots \tag{64}$$

Hence, our result includes corrections not captured in [6].

Finally, we want to calculate the correlation function of the estimator. Given the partition function $Z = \int d\mathbf{w} e^{-\beta H(\mathbf{w}; \mathcal{D}) + \beta \boldsymbol{\xi} \cdot \mathbf{w} + \frac{\beta}{2} \boldsymbol{\chi}^\top \mathbf{w} \mathbf{w}^\top \boldsymbol{\chi}}$, notice that the variance:

$$\frac{1}{\beta^2} \frac{\partial^2}{\partial \xi_\alpha \partial \xi_\beta} \langle \log Z \rangle = \left\langle \langle w_\alpha w_\beta \rangle_{\mathbf{w}} - \langle w_\alpha \rangle_{\mathbf{w}} \langle w_\beta \rangle_{\mathbf{w}} \right\rangle_{\mathcal{D}} = \frac{1}{\beta} \frac{\kappa}{P\eta_\alpha + \kappa} \delta_{\alpha\beta} \tag{65}$$

vanishes as $\beta \to \infty$ since there is a unique solution. However, there is variance to the estimator due to averaging over different training sets which is given by:

$$\langle w_\alpha w_\beta \rangle_{\mathbf{w}, \mathcal{D}} - \langle w_\alpha \rangle_{\mathbf{w}, \mathcal{D}} \langle w_\beta \rangle_{\mathbf{w}, \mathcal{D}}, \tag{66}$$

12

and it is finite as $\beta \to \infty$. The first term, the eigenfunction expansion coefficients of the correlation function of the estimator $\langle f(\mathbf{x})f(\mathbf{x}')\rangle$, can be calculated by taking two derivatives of $\langle \log Z \rangle$ with respect to $\boldsymbol{\chi}'$. To simplify the calculation, we first redefine $\mathbf{G} \equiv \left(\mathbf{I} + \frac{P}{\kappa} - \boldsymbol{\Lambda}^{1/2}\boldsymbol{\chi}'\boldsymbol{\chi}'^{\top}\boldsymbol{\Lambda}^{1/2}\right)^{-1}$ by setting $J = K = 0$ and introduce the notation $\partial_\alpha \equiv \frac{\partial}{\partial \chi'_\alpha}$ for notational simplicity. First, we calculate the derivatives of $\kappa$:

$$
\begin{aligned}
\partial_\alpha \kappa &= -\operatorname{Tr}\boldsymbol{\Lambda}\mathbf{G}^{-1}\left(-\frac{P}{\kappa^2}\boldsymbol{\Lambda}\partial_\alpha\kappa - \boldsymbol{\Lambda}^{1/2}\partial_\alpha(\boldsymbol{\chi}'\boldsymbol{\chi}'^{\top})\boldsymbol{\Lambda}^{1/2}\right)\mathbf{G}^{-1} \\
&= \partial_\alpha\kappa\frac{P}{\kappa^2}\operatorname{Tr}\boldsymbol{\Lambda}^2\mathbf{G}^{-2} + \operatorname{Tr}\boldsymbol{\Lambda}\mathbf{G}^{-1}\boldsymbol{\Lambda}^{1/2}\partial_\alpha(\boldsymbol{\chi}'\boldsymbol{\chi}'^{\top})\boldsymbol{\Lambda}^{1/2}\mathbf{G}^{-1} \\
&= \partial_\alpha\kappa\frac{P}{\kappa^2}\operatorname{Tr}\boldsymbol{\Lambda}^2\mathbf{G}^{-2} + 2\sum_\rho \eta_\rho\mathbf{G}_{\rho\alpha}^{-1}\sqrt{\eta_\alpha}\left(\sum_\sigma \sqrt{\eta_\sigma}\chi'_\sigma\mathbf{G}_{\sigma\rho}^{-1}\right).
\end{aligned}
\tag{67}
$$

Hence, we find that:

$$
\partial_{\boldsymbol{\chi}'}\kappa|_{\boldsymbol{\chi}'=0} = \partial_{\boldsymbol{\chi}'}\mathbf{G}|_{\boldsymbol{\chi}'=0} = 0.
\tag{68}
$$

This greatly simplifies the second derivative of $\kappa$:

$$
\begin{aligned}
\partial_\alpha\partial_\beta\kappa|_{\boldsymbol{\chi}'=0} &= \left[\partial_\alpha\partial_\beta\kappa\frac{P}{\kappa^2}\operatorname{Tr}\boldsymbol{\Lambda}^2\mathbf{G}^{-2} + 2\sum_\rho \eta_\rho\mathbf{G}_{\rho\alpha}^{-1}\sqrt{\eta_\alpha\eta_\beta}\mathbf{G}_{\beta\rho}^{-1}\right]\Bigg|_{\boldsymbol{\chi}'=0} \\
&= (\partial_\alpha\partial_\beta\kappa|_{\boldsymbol{\chi}'=0})\sum_\rho \frac{P\eta_\rho^2}{(P\eta_\rho+\kappa)^2} + \frac{2\kappa^2}{P}\frac{P\eta_\alpha^2}{(P\eta_\alpha+\kappa)^2}\delta_{\alpha\beta} \\
&= \frac{2\kappa^2}{P}\frac{1}{1-\gamma}\frac{P\eta_\alpha^2}{(P\eta_\alpha+\kappa)^2}\delta_{\alpha\beta},
\end{aligned}
\tag{69}
$$

where $\gamma = \sum_\rho \frac{P\eta_\rho^2}{(P\eta_\rho+\kappa)^2}$ as defined before. Now we calculate the variance of the expected function:

$$
\begin{aligned}
\sqrt{\eta_\alpha\eta_\beta}\left\langle w_\alpha^* w_\beta^*\right\rangle_{\mathcal{D}} &= \lim_{\beta\to\infty}\frac{1}{\beta}\partial_\alpha\partial_\beta\left\langle\log Z\right\rangle\big|_{\boldsymbol{\xi}',\boldsymbol{\chi}'=0} \\
&= \frac{P}{2}\frac{\sigma^2}{\kappa^2}\partial_\alpha\partial_\beta\kappa + \sqrt{\eta_\alpha\eta_\beta}\bar{w}_\alpha\bar{w}_\beta - \kappa\frac{\sqrt{\eta_\alpha\eta_\beta}\bar{w}_\alpha\bar{w}_\beta}{P\eta_\beta+\kappa} - \kappa\frac{\sqrt{\eta_\alpha\eta_\beta}\bar{w}_\alpha\bar{w}_\beta}{P\eta_\alpha+\kappa} \\
&\quad - \frac{1}{2}\bar{\mathbf{W}}^{\top}\mathbf{G}^{-1}\left(-\frac{P}{\kappa^2}\boldsymbol{\Lambda}\partial_\alpha\partial_\beta\kappa - \boldsymbol{\Lambda}^{1/2}\partial_\alpha\partial_\beta(\boldsymbol{\chi}'\boldsymbol{\chi}'^{\top})\boldsymbol{\Lambda}^{1/2}\right)\mathbf{G}^{-1}\bar{\mathbf{W}} \\
&= \frac{1}{1-\gamma}\left(\sigma^2 + \kappa^2\sum_\rho\frac{\eta_\rho\bar{w}_\rho^2}{(P\eta_\rho+\kappa)^2}\right)\frac{P\eta_\alpha^2}{(P\eta_\alpha+\kappa)^2}\delta_{\alpha\beta} + \kappa^2\frac{\sqrt{\eta_\alpha\eta_\beta}\bar{w}_\alpha\bar{w}_\beta}{(P\eta_\alpha+\kappa)(P\eta_\beta+\kappa)} \\
&\quad + \frac{P\eta_\beta\sqrt{\eta_\alpha\eta_\beta}\bar{w}_\alpha\bar{w}_\beta}{P\eta_\beta+\kappa} - \kappa\frac{\sqrt{\eta_\alpha\eta_\beta}\bar{w}_\alpha\bar{w}_\beta}{P\eta_\alpha+\kappa}.
\end{aligned}
\tag{70}
$$

Now we can calculate the coefficients of covariance of the estimator:

$$\sqrt{\eta_\alpha \eta_\beta} \left\langle w_\alpha^* w_\beta^* \right\rangle_\mathcal{D} - \sqrt{\eta_\alpha \eta_\beta} \left\langle w_\alpha^* \right\rangle_\mathcal{D} \left\langle w_\beta^* \right\rangle_\mathcal{D} \tag{71}$$

$$= \frac{1}{1-\gamma} \left( \sigma^2 + \kappa^2 \sum_\rho \frac{\eta_\rho \bar{w}_\rho^2}{(P\eta_\rho + \kappa)^2} \right) \frac{P\eta_\alpha^2}{(P\eta_\alpha + \kappa)^2} \delta_{\alpha\beta} + \kappa^2 \frac{\sqrt{\eta_\alpha \eta_\beta} \bar{w}_\alpha \bar{w}_\beta}{(P\eta_\alpha + \kappa)(P\eta_\beta + \kappa)}$$

$$+ \frac{(P\eta_\alpha + \kappa) P\eta_\beta \sqrt{\eta_\alpha \eta_\beta} \bar{w}_\alpha \bar{w}_\beta}{(P\eta_\alpha + \kappa)(P\eta_\beta + \kappa)} - \kappa \frac{(P\eta_\beta + \kappa) \sqrt{\eta_\alpha \eta_\beta} \bar{w}_\alpha \bar{w}_\beta}{(P\eta_\alpha + \kappa)(P\eta_\beta + \kappa)} - \frac{P^2 \eta_\alpha \eta_\beta \sqrt{\eta_\alpha \eta_\beta} \bar{w}_\alpha \bar{w}_\beta}{(P\eta_\alpha + \kappa)(P\eta_\beta + \kappa)}$$

$$\boxed{= \frac{1}{1-\gamma} \left( \sigma^2 + \kappa^2 \sum_\rho \frac{\eta_\rho \bar{w}_\rho^2}{(P\eta_\rho + \kappa)^2} \right) \frac{P\eta_\alpha^2}{(P\eta_\alpha + \kappa)^2} \delta_{\alpha\beta} .}$$

Hence the covariance of the estimator is:

$$Cov[\left\langle f^*(\mathbf{x}; P) f^*(\mathbf{x}'; P) \right\rangle_\mathcal{D}] = \sum_{\alpha\beta} \left( \left\langle w_\alpha^* w_\beta^* \right\rangle_\mathcal{D} - \left\langle w_\alpha^* \right\rangle_\mathcal{D} \left\langle w_\beta^* \right\rangle_\mathcal{D} \right) \psi_\alpha(\mathbf{x}) \psi_\beta(\mathbf{x}'). \tag{72}$$

Note that the generalization error can be decomposed as:

$$\left\langle E_g \right\rangle_\mathcal{D} = \int d\mathbf{x} \left\langle f^{*2}(\mathbf{x}) \right\rangle_\mathcal{D} - 2 \int d\mathbf{x} \left\langle f^*(\mathbf{x}) \right\rangle_\mathcal{D} \bar{f}(\mathbf{x}) + \int d\mathbf{x} \bar{f}(\mathbf{x})^2 \tag{73}$$

From the calculation above, we can find:

$$\int d\mathbf{x} \left\langle f^{*2}(\mathbf{x}) \right\rangle_\mathcal{D} - \int d\mathbf{x} \left\langle f^*(\mathbf{x}) \right\rangle_\mathcal{D} \left\langle f^*(\mathbf{x}) \right\rangle_\mathcal{D} = \frac{\gamma}{1-\gamma} \left( \sigma^2 + \kappa^2 \sum_\rho \frac{\eta_\rho \bar{w}_\rho^2}{(P\eta_\rho + \kappa)^2} \right)$$

$$\int d\mathbf{x} \left\langle f^*(\mathbf{x}) \right\rangle_\mathcal{D} \left\langle f^*(\mathbf{x}) \right\rangle_\mathcal{D} = \sum_\rho \frac{P^2 \eta_\rho^3 \bar{w}_\rho^2}{(P\eta_\rho + \kappa)^2}$$

$$\int d\mathbf{x} \left\langle f^*(\mathbf{x}) \right\rangle_\mathcal{D} \bar{f}(\mathbf{x}) = \sum_\rho \frac{P\eta_\rho^2 \bar{w}_\rho^2}{P\eta_\rho + \kappa}$$

$$\int d\mathbf{x} \bar{f}(\mathbf{x})^2 = \sum_\rho \eta_\rho \bar{w}_\rho^2, \tag{74}$$

where the first line is the contribution to generalization error due to estimator variance. Hence generalization error is:

$$\left\langle E_g \right\rangle_\mathcal{D} = \frac{\gamma}{1-\gamma} \left( \sigma^2 + \kappa^2 \sum_\rho \frac{\eta_\rho \bar{w}_\rho^2}{(P\eta_\rho + \kappa)^2} \right) + \sum_\rho \frac{P^2 \eta_\rho^3 \bar{w}_\rho^2}{(P\eta_\rho + \kappa)^2} - 2 \sum_\rho \frac{P\eta_\rho^2 \bar{w}_\rho^2}{P\eta_\rho + \kappa} + \sum_\rho \eta_\rho \bar{w}_\rho^2$$

$$= \frac{\gamma}{1-\gamma} \sigma^2 + \frac{\kappa^2}{1-\gamma} \sum_\rho \frac{\eta_\rho \bar{w}_\rho^2}{(P\eta_\rho + \kappa)^2}, \tag{75}$$

which is what we obtained before. In terms of the variance of the estimator, $E_g$ is also equal to:

$$\left\langle E_g \right\rangle_\mathcal{D} = \underbrace{\frac{\gamma}{1-\gamma} \left( \sigma^2 + \kappa^2 \sum_\rho \frac{\eta_\rho \bar{w}_\rho^2}{(P\eta_\rho + \kappa)^2} \right)}_{\textbf{variance } V} + \underbrace{\kappa^2 \sum_\rho \frac{\eta_\rho \bar{w}_\rho^2}{(P\eta_\rho + \kappa)^2}}_{\textbf{bias } B} . \tag{76}$$

This is the bias-variance decomposition of generalization error in our setting where the bias term is monotonically decreasing while the variance term is solely responsible for any non-monotonicity appearing in the generalization error.

14

# Supplementary Note 3 - White Band-limited RKHS Spectrum

As a simple but illuminating example, we consider a kernel with band-limited spectrum: $\eta_\rho = 0$ for $\rho > N$. For simplicity, we study the case where the spectrum is white $\eta_\rho = \frac{1}{N}$ for all $\rho = 1, ..., N$ and study this system in the large $N$, large $P$ limit with $\alpha = P/N \sim \mathcal{O}(1)$. We normalize the target power in the first $N$ modes $\sum_{\rho=1}^{N} \overline{w}_\rho^2 = N$. Furthermore, the coefficients for the target function are $a_\rho$ for all $\rho > N$: $f^*(\mathbf{x}) = \sum_{\rho=1}^{N} \overline{w}_\rho \psi_\rho(\mathbf{x}) + \sum_{\rho=N+1}^{\infty} a_\rho \phi_\rho(\mathbf{x})$. At the saddle point, the implicit equation $\kappa$ can be solved explicitly in the $P, N \to \infty$ limit with $\alpha = P/N \sim \mathcal{O}(1)$.

$$\kappa_\lambda(\alpha) = \frac{1}{2} \left[ (\lambda + 1 - \alpha) + \sqrt{(\lambda + 1 - \alpha)^2 + 4\lambda\alpha} \right],$$ (77)

The generalization error Supplementary Equation 53 becomes:

$$E_g = \frac{\kappa_\lambda(\alpha)^2 + \tilde{\sigma}^2 \alpha}{(\kappa_\lambda(\alpha) + \alpha)^2 - \alpha} + E_g(\infty)$$ (78)

where $E_g(\infty) = \sum_{\rho > N} a_\rho^2$ is the asymptotic value of the generalization error and $\tilde{\sigma}^2 = \sigma^2 + E_g(\infty)$ is the effective noise. The first term is the noiseless contribution to $E_g$ while second term is only due to the noise in target coming from both the explicit noise $\sigma^2$ and the variance of modes in the null-space of the kernel $E_g(\infty)$. In the following we will simply denote $\tilde{\sigma}^2$ with $\sigma^2$ since it is only relevant noise appearing in the generalization error formula. The generalization error asymptotically falls faster in the absence of noise:

$$E_g - E_g(\infty) \sim \frac{\sigma^2}{\alpha}, \ \alpha \to \infty, \ (\sigma > 0),$$
$$E_g - E_g(\infty) \sim \frac{\lambda^2}{\alpha^2}, \ \alpha \to \infty, \ (\sigma = 0).$$ (79)

Furthermore, explicit calculation reveals that the noiseless term monotonically decreases with $\alpha$, while the noise term has a maximum at $\alpha = 1 + \lambda$ and its maximum is given by:

$$\left. \frac{\gamma}{1 - \gamma} \right|_{\alpha = 1 + \lambda} = \frac{1}{2\sqrt{\lambda}} \frac{1}{\sqrt{\lambda} + \sqrt{\lambda + 1}}$$ (80)

In the presence of noise, generalization error diverges when $\lambda \to 0$, while finite $\lambda$ smoothes out the learning curve. In machine learning, this non-monotonic behavior of generalization error is called "double-descent", and signals overfitting of the noise in the data [7, 8, 9]. Diverging generalization error further implies a first order phase transition when $\alpha = 1 + \lambda = 1$. This can be seen by examining the first derivative of the free energy Supplementary Equation 50 in $\beta \to \infty$ limit:

$$\frac{1}{\beta} \frac{\partial S}{\partial \alpha} = \frac{\sigma^2}{2\kappa} \frac{\alpha \gamma}{1 - \gamma} \sim \frac{\sigma^2}{2\lambda} \theta(\alpha - 1) + \mathcal{O}(\lambda) \ , \ \lambda \to 0,$$ (81)

where the approximation is valid for $\lambda \ll 1$. We observe that, in the absence of noise, there is no phase transition while in the noisy case, there is a sharp discontinuity and divergence when $\lambda = 0$. Although there is no phase transition in the strict sense of a non-analytic free energy except for the case $\lambda = 0$, we describe whether there is non-monotonicity or not as separate phases of the kernel machine.

We would like to understand what combinations of $(\lambda, \sigma^2)$ leads to non-monotonicity in generalization error. One can obtain the exact phase boundary for non-monotonicity by studying the zeros

15

of $\partial E_g / \partial \alpha$ given by:

$$\frac{\partial E_g(\alpha)}{\partial \alpha} = -\frac{1}{2} + \frac{(\alpha + \lambda - 5)(\alpha + \lambda + 1)^2 + 2(\lambda + 2)(3\alpha + 1 + \lambda) - 2\sigma^2(\alpha - 1 - \lambda)}{2\left((\alpha + 1 + \lambda)^2 - 4\alpha\right)^{3/2}} = 0 \quad (82)$$

Explicit calculation yields:

$$\sigma_{\text{critical}}^2 > \begin{cases} g(\lambda) & \lambda < 1 \\ 2\lambda + 1 & \lambda \geq 1 \end{cases}, \quad (83)$$

where $g(\lambda)$ is:

$$g(\lambda) = 3\lambda(3\lambda + 2 - 2\sqrt{1 + \lambda}\sqrt{9\lambda + 1}\cos\theta),$$
$$\theta = \frac{1}{3}\left(\pi + \tan^{-1}\frac{8\sqrt{\lambda}}{9\lambda(3\lambda + 2) - 1}\right). \quad (84)$$

In the non-monotonic region, we further observe that the curve $\sigma_{\text{critical}}^2 = 2\lambda + 1$ for $\lambda < 1$ separates two regions with a single and double local extrema. Above this curve, there is a single local maximum corresponding to a peak while below there is a local minimum followed by a local maximum (double-descent). As a side note, we emphasize that double-descent might occur when target functions have out-of-RKHS components even without label corruption. In Supplementary Figure 2, the target function is chosen such that $E_g(\infty) = 0.2$ in Supplementary Equation 79. Since this causes non-zero effective noise, we observe double-descent even in the absence of label noise.

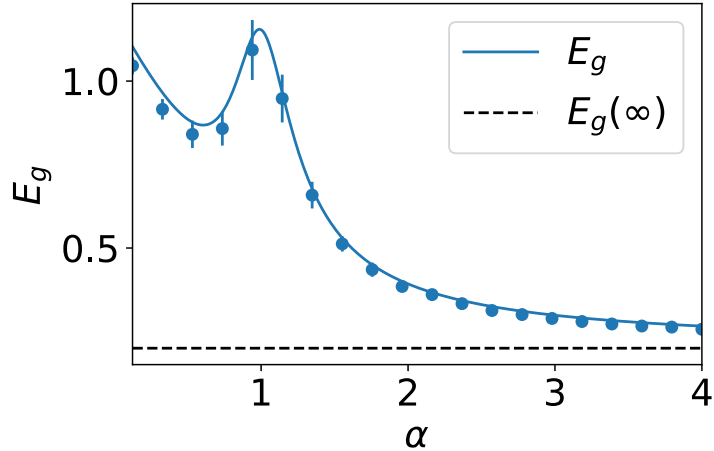

Supplementary Figure 2: Zero noise target function with components outside of the RKHS. We simulated the linear kernel $K(\mathbf{x}, \mathbf{x}') = \sum_{\rho=1}^{N} \eta_\rho x_\rho x'_\rho$ where $\mathbf{x} \sim \mathcal{N}(0, \mathbf{I})$ are $N = 960$ dimensional uncorrelated Gaussian inputs and $\eta_\rho = 0$ for $\rho > 800$ and $\eta_\rho = 1$, otherwise. Target is a linear function $\bar{f}(\mathbf{x}) = \boldsymbol{\beta}^\top \mathbf{x}$ with $\boldsymbol{\beta} \sim \mathcal{N}(0, \frac{1}{N}\mathbf{I})$. This experiment demonstrates that out-of-RKHS components generate an irreducible error $E_g(\infty)$ (here $E_g(\infty) = 0.2$) and acts as effective noise, producing a double descent peak. The ridge parameter is chosen to be $\lambda = 0.01$. Error bars represent the standard deviation over 50 trials.

Although large $\lambda$ regularizes the learning and avoids an overfitting peak, too large $\lambda$ will also slow down the learning as can be seen from the asymptotic limit of Supplementary Equation 82 in $\lambda$:

$$\frac{\partial E_g(\alpha)}{\partial \alpha} = -\frac{2}{\lambda} + \frac{3(2\alpha + 1) + \sigma}{\lambda^2} + \mathcal{O}\left(\frac{1}{\lambda^3}\right) \quad (85)$$

To find an optimal choice of ridge parameter, we study the first derivative of $E_g$ with respect to $\lambda$ and find that there is an optimal $\lambda$ for a given noise level $\sigma^2$ independent of $\alpha$:

$$\frac{\partial E_g(\alpha)}{\partial \lambda} = \frac{2\alpha(\lambda - \sigma^2)}{\left((\alpha + 1 + \lambda)^2 - 4\alpha\right)^{3/2}} = 0, \quad \Rightarrow \quad \lambda^* = \sigma^2 \tag{86}$$

This simple relation holds for all $\alpha$ and also indicates that the optimal choice of regularization leads to a monotonic learning curve as expected (See Fig. 3 in the main text). Note that the error due to the noise term is decreasing, while the noise-independent term is increasing with $\lambda$.

Finally, we numerically plot the $\alpha$ at which learning curve peak occurs as a function of noise with varying $\lambda$ levels in Supplementary Figure 3. We observe that for large noise levels, location of the peak gets closer to $\alpha = 1 + \lambda$.

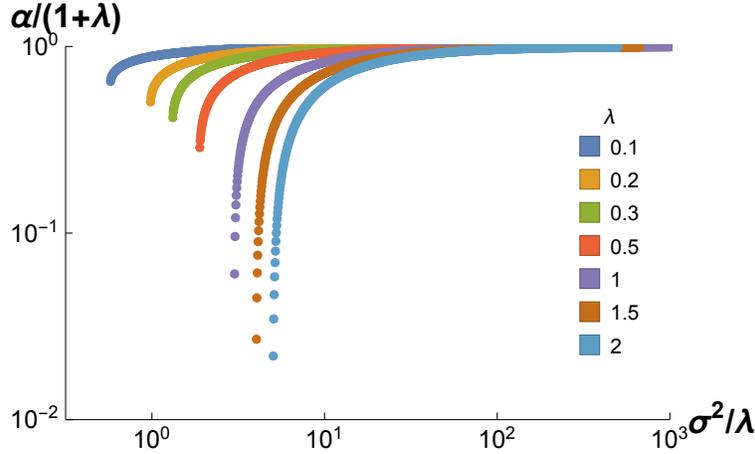

Supplementary Figure 3: Numerical calculation of $\alpha/(1 + \lambda)$ as a function of $\sigma^2/\lambda$ for varying ridge parameter $\lambda$.

## Connection to Random Matrices

The quantities of interest in the generalization formula for the white band-limited spectrum are related to the distribution of random Wishart Matrices. In particular, let $\mathbf{X} \in \mathbb{R}^{N \times P}$ be a random Gaussian matrix with entries $X_{ij} \sim \mathcal{N}(0, 1)$, then the quantity $\kappa_\lambda(\alpha)$ can be directly related to the following Stieltjes transform (resolvent) of the matrix $\frac{1}{N}\mathbf{X}\mathbf{X}^\top + \lambda\mathbf{I}$. The relationship is

$$\kappa_\lambda(\alpha) = \lambda + \frac{\lambda}{N}\text{Tr}\left\langle \left(\frac{1}{N}\mathbf{X}\mathbf{X}^\top + \lambda\mathbf{I}\right)^{-1}\right\rangle_{\{X_{ij}\}} = \frac{1}{2}\left[(1 + \lambda - \alpha) + \sqrt{(1 + \lambda - \alpha)^2 + 4\lambda\alpha}\right], \tag{87}$$

where $\alpha = P/N$ in the $P, N \to \infty$ limit [10, 11]. If one wanted to compute the average case (over design matrices $\mathbf{X}$) generalization error for linear regression, this random matrix arises naturally from the solution to the regularized least-squares regression loss

$$\min_{\mathbf{w}} ||\frac{1}{\sqrt{N}}\mathbf{X}^\top\mathbf{w} - \mathbf{y}||^2 + \lambda||\mathbf{w}||^2 \implies \mathbf{w}^* = \frac{1}{\sqrt{N}}\left(\frac{1}{N}\mathbf{X}\mathbf{X}^\top + \lambda\mathbf{I}\right)^{-1}\mathbf{X}\mathbf{y}. \tag{88}$$

If the target values are produced by a teacher $\mathbf{y} = \frac{1}{\sqrt{N}}\mathbf{X}^\top\overline{\mathbf{w}}$ then

$$E_g = \overline{\mathbf{w}}^\top\left\langle \left(\frac{1}{\lambda}\mathbf{X}\mathbf{X}^\top + N\mathbf{I}\right)^{-2}\right\rangle\overline{\mathbf{w}} = -\frac{\partial}{\partial N}\overline{\mathbf{w}}^\top\left\langle \left(\frac{1}{\lambda}\mathbf{X}\mathbf{X}^\top + N\mathbf{I}\right)^{-1}\right\rangle\overline{\mathbf{w}}. \tag{89}$$

This matrix, and its average over the design $\mathbf{X}$, is also studied intensively in previous work on the generalization error of kernel regression [12, 13, 14]. Differentiation in $N$ of this average matrix gives the desired result in the $N \to \infty$ limit

$$E_g = \frac{\kappa_\lambda(\alpha)^2}{(\kappa_\lambda(\alpha) + \alpha)^2 - \alpha} \lim_{N \to \infty} \left[ ||\overline{\mathbf{w}}||^2 / N \right] \tag{90}$$

Since the distribution of each $\mathbf{x}^\mu$ is isotropic, all covariance eigenvalues are equal and each component $w_\rho$ is learned at an identical rate. This expression is the same as our expression for the white band limited case, Supplementary Equation 79.

Next, we apply our findings to rotation invariant kernels and find that generalization error decomposes into different learning episodes which are individually described by the same formula we derived here in a special setting.

## Supplementary Note 4 - Rotation Invariant Kernels

Here, we consider a widely used class of kernels left invariant under the rotations of the inputs: $K(\mathbf{Ox}, \mathbf{Ox'}) = K(\mathbf{x}, \mathbf{x'})$. We start by decomposing rotation invariant kernels into their spherical and radial directions:

**Lemma 1.** *Let $\mathcal{F}_\mathbf{r}$ be the set of functions that are invariant to all rotations that leave the vector $\mathbf{r} \in \mathbb{S}^{D-1}$ unchanged for all $f \in \mathcal{F}_\mathbf{r}$ and all orthogonal matrices $\mathbf{O} \in \mathbb{R}^{D \times D}$ with $\mathbf{Or} = \mathbf{r}$, $f(\mathbf{Ox}) = f(\mathbf{x})$. Any function $f \in \mathcal{F}_\mathbf{r}$ admits a decomposition*

$$f(\mathbf{x}) = \sum_k a_k(||\mathbf{x}||) Q_k^{(D-1)}(\hat{\mathbf{x}} \cdot \mathbf{r}), \tag{91}$$

*where $Q_k^{(D-1)}(z)$ are the Gegenbauer polynomials (see Supplementary Note 6 for review).*

*Proof.* For $f$ to be invariant under the set of rotations which leave the vector $\mathbf{r}$ invariant, the restriction of $f$ to spherical shells of radius $||\mathbf{x}|| = R$ must also be invariant under rotations. For fixed radius $R$, the set of all functions that are rotation invariant lie in $\mathrm{span}\{Q_k(\mathbf{r}^\top \cdot / || \cdot ||)\}$, since the Gegenbauer polynomials are complete with respect to the measure of inner products on $\mathbb{S}^{D-1}$. Repeating this decomposition for each restriction radius $||\mathbf{x}||$ gives radial dependent coefficients $a_k(||\mathbf{x}||)$. □

Using this lemma, we have the following decomposition for rotation invariant kernels ($K(\mathbf{Ox}, \mathbf{Ox'}) = K(\mathbf{x}, \mathbf{x'})$) by first considering the rotation $\mathbf{O}$'s that leave $\mathbf{x}$ unchanged and then by considering the rotation $\mathbf{O}$'s that leave $\mathbf{x'}$ unchanged:

$$K(\mathbf{x}, \mathbf{x'}) = \sum_k g_k(||\mathbf{x}||, ||\mathbf{x'}||) Q_k(\mathbf{x} \cdot \mathbf{x'}). \tag{92}$$

To calculate the eigenspectrum, we insert an ansatz of the form $\phi_{zkm}(\mathbf{x}) = R_{z,k}(||\mathbf{x}||) Y_{km}(\hat{\mathbf{x}})$ to the eigenvalue problem

$$\int d\mathbf{x} K(\mathbf{x}, \mathbf{x'}) p(\mathbf{x}) \phi_{zkm}(\mathbf{x})$$

$$= \int_0^\infty d||\mathbf{x}|| p(||\mathbf{x}||) \sum_{k'm'} g_{k'}(||\mathbf{x}||, ||\mathbf{x'}||) R_{z,k}(||\mathbf{x}||) Y_{k'm'}(\hat{\mathbf{x}'}) \int_{\mathbb{S}^{D-1}} d\hat{\mathbf{x}} \, Y_{km}(\hat{\mathbf{x}}) Y_{k'm'}(\hat{\mathbf{x}})$$

$$= Y_{km}(\hat{\mathbf{x}'}) \int_0^\infty d||\mathbf{x}|| p(||\mathbf{x}||) g_k(||\mathbf{x}||, ||\mathbf{x'}||) R_{z,k}(||\mathbf{x}||) = \eta_{z,k} R_{z,k}(||\mathbf{x'}||) Y_{km}(\mathbf{x'}), \tag{93}$$

which gives a collection of radial eigenvalue problems (one for each degree $k$ of spherical harmonics)

$$\int_0^\infty d||\mathbf{x}|| p(||\mathbf{x}||) g_k(||\mathbf{x}||, ||\mathbf{x}'||) R_{z,k}(||\mathbf{x}||) = \eta_{z,k} R_{z,k}(||\mathbf{x}'||). \tag{94}$$

For each, $k$, we solve the integral eigenvalue problem for a set of functions $\{R_{z,k}(||\mathbf{x}||)\}_z$ that are orthonormal with respect to $p(||\mathbf{x}||)$. After solving these radial eigenvalue problems, we obtain the following Mercer decomposition of the kernel

$$K(\mathbf{x}, \mathbf{x}') = \sum_{zkm} \eta_{z,k} R_{z,k}(||\mathbf{x}||) R_{z,k}(||\mathbf{x}'||) Y_{km}(\hat{\mathbf{x}}) Y_{km}(\hat{\mathbf{x}}'), \tag{95}$$

where $\eta_{z,k}$ are the eigenvalues of this decomposition, $R_{z,k}(||\mathbf{x}||)$ denotes the radial dependence and $Y_{km}$ are hyper-spherical harmonics in $D$-dimensions. The eigenvalues $\eta_{z,k}$ are the same for every $m$ for each $(z, k)$ mode. There are at least $N(D, k) = \binom{k + D - 1}{D - 1} - \binom{k + D - 3}{D - 1} \sim \mathcal{O}(D^k)$ degeneracy of each kernel mode just due to the rotational symmetry (see Supplementary Note 6). In case of eigenvalue degeneracy, in order to keep $K(\mathbf{x}, \mathbf{x}') \sim \mathcal{O}_D(1)$, each $(z, k)$ term in the Mercer decomposition must be $\mathcal{O}_D(1)$. Since the sum over $N(D, k)$ orders $m$ gives a scaling of each $(z, k)$ term of $\mathcal{O}_D(N(D, k)) \sim \mathcal{O}_D(D^k)$, the eigenvalues must scale like $\eta_{z,k} \sim \mathcal{O}_D(1/N(D, k)) \sim \mathcal{O}_D(D^{-k})$. This observation guides us to consider another possible scaling of $P$ via $P/N(D, k) \sim \mathcal{O}(1)$ leading to a non-trivial generalization error in the limit $P, D \to \infty$.

For the most general case, generalization error for a rotational invariant kernel is given by Supplementary Equation 53 with $\rho$ summation replaced by a sum over $(z, k, m)$. As noted above, eigenvalues $\eta_{z,k,m} = \eta_{z,k}$ are the same for $m = 1, ..., N(D, k)$. For given $k$, $\eta_{z,k}$ for all $z$ might be different with same degeneracy $N(D, k)$. In this case, we choose the scaling $\alpha = P/N(D, l) \sim \mathcal{O}_D(1)$ and take the limit $P, D \to \infty$. We find that the resulting generalization decouples over different $k$ modes for a general target function $\bar{f}(\mathbf{x}) = \sum_{z,k,m} \bar{w}_{z,k,m} \phi_{z,k,m}(\mathbf{x})$:

$$E_g = \frac{\kappa^2}{1 - \gamma} \sum_z \frac{\bar{\eta}_{z,l} \bar{w}_{z,l}^2}{(\kappa + \alpha \bar{\eta}_{z,l})^2} + \frac{1}{1 - \gamma} \sum_{z,k>l} \bar{\eta}_{z,k} \bar{w}_{z,k}^2 + \sigma^2 \frac{\gamma}{1 - \gamma},$$

$$\kappa = \lambda + \sum_z \frac{\kappa \bar{\eta}_{z,l}}{\kappa + \alpha \bar{\eta}_{z,l}} + \sum_{z,k} \bar{\eta}_{z,k}, \quad \gamma = \sum_z \frac{\alpha \bar{\eta}_{z,l}^2}{(\kappa + \alpha \bar{\eta}_{z,l})^2}, \tag{96}$$

where we defined the following $\mathcal{O}(1)$ quantities:

$$\bar{\eta}_{z,k} \equiv N(D, k) \eta_{z,k}, \qquad \bar{w}_{z,k}^2 \equiv \frac{1}{N(D, k)} \sum_{m=1}^{N(D,k)} \bar{w}_{z,k,m}^2. \tag{97}$$

First term corresponds to learning the mode $l$ features while second term corresponds to the higher modes. Note that $\gamma(\alpha = 0) = \gamma(\alpha = \infty) = 0$ meaning that the modes $k > l$ are not being learned in the learning stage $l$. Last term is the noise contribution to $E_g$. Furthermore, self-consistent equation for $\kappa$ simplifies to a polynomial equation of degree $\#(z) + 1$ instead of degree $\#(z) + \#(l) + 1$, where $\#(z)$ and $\#(l)$ denote the total number of $z$ and $l$ modes, respectively.

We found that eigenvalues with different degeneracies $N(D, k)$ decouple as different learning stages for generic rotation invariant kernels in $D \to \infty$ limit. However, kernels with further symmetries such as translational invariance can have eigenvalues with larger degeneracies. To take this case into account, we introduce the following notation: $\eta_K$ denotes the degenerate eigenvalues indexed by an

integer $K$ potentially representing different combinations of $(z, k)$ and $\phi_{K,\rho}$ denotes the corresponding eigenfunctions where $\rho$ denotes collectively the degenerate indices. In this case, the degeneracy of each mode $K$ is denoted by $N(D, K)$ which can be larger than the degeneracy of spherical harmonics. Considering the case where there is only a single eigenvalue for integer mode $K$ with degeneracy $N(D, K)$, self-consistent equation for $\kappa$ for learning stage $L$ in Supplementary Equation 96 becomes a quadratic equation and we obtain the following solution:

$$\tilde{\kappa}(\alpha) \equiv \frac{\kappa}{\bar{\eta}_L} = \frac{1}{2}(1 + \tilde{\lambda}_L - \alpha) + \frac{1}{2}\sqrt{(1 + \tilde{\lambda}_L + \alpha)^2 - 4\alpha},$$

$$\tilde{\lambda}_L = \frac{\lambda + \sum_{K>L} \bar{\eta}_K}{\bar{\eta}_L},$$

$$\tilde{\kappa}(0) = 1 + \tilde{\lambda}_L, \qquad \tilde{\kappa}(\infty) = \tilde{\lambda}_L, \qquad \tilde{\kappa}(\alpha) \geq 0, \quad \forall \alpha \in \mathbb{R}^+, \tag{98}$$

where $\tilde{\kappa}$ is the scaled $\kappa$ by $\bar{\eta}_K$ and $\bar{\eta}_K = N(D, K)\eta_K$. In this case, we choose the scaling $P = \alpha N(D, K)$. This formula is same as the white band-limited example except for a more complicated *effective regularization* $\tilde{\lambda}_L$. Therefore, each learning stage behaves in the same way as white band-limited case, and in the presence of noise, we may observe to see multiple descents associated to each learning episode.

Similar to the discussion for white band-limited case, $\tilde{\kappa}$ is a monotonically decreasing function of $\alpha$. Effective regularization $\tilde{\lambda}_L$ controls the decay rate of $\tilde{\kappa}$ and is completely fixed by kernel eigenspectrum and explicit ridge parameter. For larger $\tilde{\lambda}_L$, the decay of $\kappa(\alpha)$ is slower and for $\tilde{\lambda}_L = 0$, decay is fastest. In fact, for the special case $\tilde{\lambda}_L = 0$ decay rate is discontinuous and the second derivative of $\tilde{\kappa}$ diverges at $\alpha = 1 + \tilde{\lambda}_L = 1$.

With these definitions, $\gamma$ becomes:

$$\gamma = \frac{\alpha}{(\tilde{\kappa} + \alpha)^2}. \tag{99}$$

Similar to the discussion in white band-limited example, the function $\gamma$ has a maximum at $\alpha = 1 + \tilde{\lambda}_L$, and as $\tilde{\lambda}_L \to 0$, its maximum goes to $\gamma \to 1$, while for large $\tilde{\lambda}_L$ its maximum falls like $1/4\tilde{\lambda}_L$. Therefore, for certain cases we expect local maxima or divergences in generalization error due to the factor of $1/(1 - \gamma)$ and for larger $\tilde{\lambda}_L$ we expect the effect of peaks to decrease, acting as an effective regularization [15].

Replacing these definitions in Supplementary Equation 53, we obtain the generalization error for rotation invariant kernels as:

$$\frac{E_g^{(L)}(\alpha) - E_g^{(L)}(\infty)}{\bar{\eta}_L \bar{w}_L^2} = \frac{1}{1 - \gamma}\frac{\tilde{\kappa}^2}{(\tilde{\kappa} + \alpha)^2} + \left(\frac{\sigma^2 + E_g^{(L)}(\infty)}{\bar{\eta}_L \bar{w}_L^2}\right)\frac{\gamma}{1 - \gamma}, \tag{100}$$

where $E_g^{(L)}(\infty) = \sum_{K>L} \bar{\eta}_K \bar{w}_K^2$ is the asymptotic value of the generalization error and superscript $(L)$ indicates that we are considering the scaling $P = N(D, L)\alpha$. The particular form we presented $E_g^{(L)}$ is useful to study $\alpha$ dependence of generalization error across different modes $L$ since the right-hand side of the equation functionally depends only on $\alpha$ and $\tilde{\lambda}_L$ which is completely fixed by the full spectrum of RKHS. Asymptotically, first term is monotonically decreasing with $\frac{1}{\alpha^2}$, while the second term has a maximum at $\alpha = 1 + \tilde{\lambda}_L$ with magnitude:

$$\frac{\gamma(\tilde{\lambda}_L)}{1 - \gamma(\tilde{\lambda}_L)} = \frac{1}{2\sqrt{\tilde{\lambda}_L}}\frac{1}{\sqrt{\tilde{\lambda}_L} + \sqrt{1 + \tilde{\lambda}_L}}, \tag{101}$$

where generalization error might display a peak with increasing training samples. Therefore, we conclude that the "double descent" behavior can only arise due to the noise in target, consistent with the observations of [16]. We also observe that the *effective noise* is given by $\tilde{\sigma}_L^2 \equiv \frac{\sigma^2 + E_g^{(L)}(\infty)}{\bar{\eta}_L \bar{w}_\zeta^2}$ which implies that the errors from higher modes might act like noise in generalization error. Note that effective noise can be scale $N(D, L)$ dependent due to the weight factor in the denominator.

From the particular form of generalization error in Supplementary Equation 96, we observe that there is a trade-off between noiseless and the noisy term and it is not obvious for which combinations of $\tilde{\sigma}_L^2$ and $\tilde{\lambda}_L$ the generalization error has a local maximum. Similar to the discussion in white band-limited case, we obtain the a phase diagram by identifying where on the $(\tilde{\lambda}_L, \tilde{\sigma}_L^2)$ plane the first derivative of Supplementary Equation 96 vanishes defined:

$$\tilde{\sigma}_L^2 \geq g(\tilde{\lambda}_L) \equiv 3\tilde{\lambda}_L \big(3\tilde{\lambda}_L + 2 - 2\sqrt{1 + \tilde{\lambda}_L}\sqrt{9\tilde{\lambda}_L + 1}\cos\theta_L\big),$$
$$\theta_L = \frac{1}{3}\left(\pi + \tan^{-1}\frac{8\sqrt{\tilde{\lambda}_L}}{9\tilde{\lambda}_L(3\tilde{\lambda}_L + 2) - 1}\right) \tag{102}$$

Above this curve where non-monotonicity occurs, we further observe that the curve $\tilde{\sigma}_L^2 = 2\tilde{\lambda}_L + 1$ for $\tilde{\lambda}_L < 1$ separates two regions with a single and double local extrema.

Here, similar to the white band-limited case, we find an optimal $\tilde{\lambda}_L^* = \tilde{\sigma}_L^2$ for each learning episode $L$, achieving the minimum generalization error for all $\alpha$.

This analysis allows us to understand the dependence of non-monotonic learning behavior on the kernel spectrum by studying $\tilde{\lambda}_L$. Let us consider the case where $\bar{\eta}_L \sim \mathcal{O}(s^{-L})$ for some $s > 1$, the case relevant for the Gaussian kernel example. Then in the ridgeless ($\lambda = 0$) limit $\tilde{\lambda}_L$ is given by:

$$\tilde{\lambda}_L = \frac{\sum_{K > L} s^{-K}}{s^{-L}} = \sum_{K=L}^{\infty} s^{-K} = \frac{1}{s-1}. \tag{103}$$

Here $\tilde{\lambda}_L$ is the same for all $L$. We observe that as spectrum decays faster, generalization error might feature larger peaks since the regularization $\tilde{\lambda}_L \to 0$. Therefore, faster decaying spectra are more likely to cause non-monotonic learning curves than the slower decaying ones.

Another example is $\bar{\eta}_L \sim \mathcal{O}(L^{-s})$ which is more relevant for studying neural networks. In this case, we have:

$$\tilde{\lambda}_L = \frac{\lambda}{\bar{\eta}_L} + \frac{\sum_{K=L}^{\infty} \bar{\eta}_K}{\bar{\eta}_L} - 1 = \frac{\lambda}{\bar{\eta}_L} + \sum_{K=0}^{\infty}\left(\frac{L+K}{L}\right)^{-s} - 1$$
$$= L^s\big(\zeta(s, L) + \lambda\big) - 1, \tag{104}$$

where $\zeta(s, L)$ is the Hurwitz zeta function defined as:

$$\zeta(s, L) \equiv \sum_{K=0}^{\infty}\frac{1}{(L+K)^s} = \frac{1}{\Gamma(s)}\int_0^\infty\frac{e^{-Lx}x^{s-1}}{1 - e^{-x}}dx = \frac{l^{-s+1}}{\Gamma(s)}\int_0^\infty\frac{e^{-x}x^{s-1}}{1 - e^{-x/L}}dx, \tag{105}$$

where in the last step, we performed the change of variables $x \to tx$. To understand the spectrum dependence of $\tilde{\lambda}_L$, we approximate $\zeta(s, L) \approx L^{-s+1}/(s-1)$ for large $L$. Then $\tilde{\lambda}_L$ simplifies to:

$$\tilde{\lambda}_L \approx \frac{L}{s-1} + \lambda L^s. \tag{106}$$

Similar to exponential spectrum, again the regularization falls as spectrum decays faster. Furthermore, we can see that regularization $\tilde{\lambda}_L$ increases at least linearly with $L$ (or with power law for $\lambda \neq 0$) meaning that non-monotonicity becomes less visible for higher modes. Another note is that one can think of the quantity $\tilde{\lambda}_L$ as an "effective ridge parameter" which regularizes higher order modes causing them not to fit random noise and therefore stay smoother.

This can be thought of as an example of implicit regularization in learning machines where more complicated features (higher modes) are implicitly chosen not to be learned, since learning rates also slow down with sample complexity as $\tilde{\lambda}_L$ gets larger. This property of the power law spectrum keeps the learned function smoother. It can be also interpreted as explicitly regularizing the learning with a mode dependent ridge parameter $\lambda = L/(s-1)$.

Next, we consider concrete examples of the theory.

## Gaussian kernel

As a popular example, we study Gaussian kernel which further also possesses translational symmetry. Let $p(\mathbf{x}) = \mathcal{N}(0, r^2\mathbf{I})$ be the data distribution on the input space $\mathbb{R}^D$ and $K(\mathbf{x}, \mathbf{x}') = e^{-\frac{1}{2D\omega^2}||x-x'||^2}$ be the Gaussian kernel. For this density and the kernel, the eigenfunctions and eigenvalues can be computed exactly [1]:

$$\eta_{\mathbf{k}} = \left(\frac{2a}{A}\right)^{\frac{D}{2}} \left(\frac{b}{A}\right)^{\sum_i k_i}, \quad \phi_{\mathbf{k}}(\mathbf{x}) = e^{-(c-a)||\mathbf{x}||^2} \prod_{i=1}^{D} H_{k_i}(\sqrt{2c}x_i), \tag{107}$$

where $a = \frac{1}{4r^2}$, $b = \frac{1}{2D\omega^2}$, $c = \sqrt{a^2 + 2ab}$ and $A = a + b + c$. The degeneracy of each mode for fixed $K = \sum_i k_i$ is given by: $\binom{K+D-1}{K} \sim \frac{D^K}{K!}$ in large $D$ limit. We note that this system can also be thought of as a collection of $D$ harmonic oscillators, where the eigenfunctions represent different microstates. The degeneracy in the eigenspectrum is analogous to the number of states with the same total energy: the number of distinguishable macrostates possible when $K$ energy quanta are distributed over $D$ oscillators. The Gaussian can be decomposed in spherical polar coordinates in terms of angular and radial functions as in Supplementary Equation 95, but this decomposition is more complicated than the decomposition in Cartesian coordinates we study here.

An informative limit to study the spectrum is one where $\omega^2 \sim \mathcal{O}_D(1)$ in the $D \to \infty$ limit. In this large $D$ limit, the normalized spectrum converges to

$$\bar{\eta}_K = \eta_K N(D, K) \sim \left(\frac{r^2}{\omega^2 D}\right)^K \frac{D^K}{K!} = \frac{1}{K!} \left(\frac{r^2}{\omega^2}\right)^K \sim \mathcal{O}_D(1). \tag{108}$$

We can also compute the effective regularization at each learning stage

$$\begin{aligned}
\tilde{\lambda}_K &= K! \left(\frac{r^2}{\omega^2}\right)^{-K} \sum_{\ell > K} \frac{1}{\ell!} \left(\frac{r^2}{\omega^2}\right)^{\ell} = \sum_{\ell=1}^{\infty} \frac{K!}{(\ell+K)!} \left(\frac{r^2}{\omega^2}\right)^{\ell} \\
&= K! \left(\frac{\omega^2}{r^2}\right)^K \left[\exp\left(\frac{r^2}{\omega^2}\right) - \sum_{\ell=0}^{K} \frac{1}{\ell!} \left(\frac{r^2}{\omega^2}\right)^{\ell}\right],
\end{aligned} \tag{109}$$

which we see is a monotonically increasing function of $r^2/\omega^2$. Thus, for fixed distribution variance $r^2$, a larger kernel bandwidth $\omega^2$ leads to lower effective regularization. This is associated with larger learning curve peaks in the presence of noise. A smaller kernel bandwidth leads to larger effective regularization, mitigating the peak. The optimal $\omega^2$ for the first learning stage can be determined

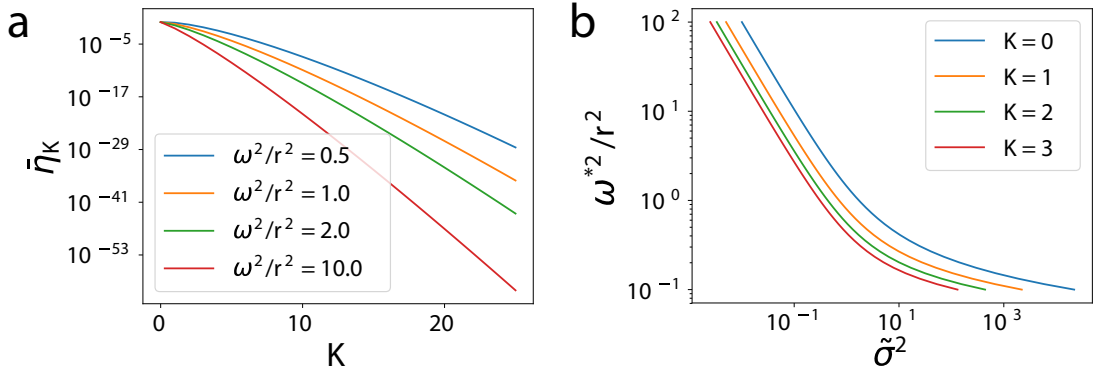

Supplementary Figure 4: Kernel Eigenspectra for the Gaussian RBF on a Gaussian measure in the $D \to \infty$ limit. **a.** Larger bandwidth spectra decay more rapidly with increasing $K$. **b.** The optimal bandwidth $\omega^*$ as a function of the effective noise $\tilde{\sigma}^2$. Small bandwidth kernels are preferred for late learning stages (large $K$) and large effective noise $\tilde{\sigma}^2$. For small $\tilde{\sigma}^2$ the optimal bandwidth satisfies $\omega^{*2} \propto \tilde{\sigma}^{-2}$ as predicted by the approximation obtained in the $r \ll \omega$ limit.

by setting $\tilde{\sigma}_K^2 = \tilde{\lambda}_K$. Under the assumption that the kernel bandwidth is large $\omega^2 \gg r^2$, we find that $\tilde{\lambda}_K \sim \frac{1}{1+K} \frac{r^2}{\omega^2}$ so that the optimal bandwidth for learning stage $K$ and noise level $\tilde{\sigma}_K^2 \ll 1$ is

$$\omega_K^{*2} \sim \frac{r^2}{\tilde{\sigma}^2(K+1)} \ , \ \tilde{\sigma}^2 \to 0. \tag{110}$$

The effective eigenvalues $\overline{\eta}_K$ for each learning stage $K$ and different bandwidths are provided in Supplementary Figure 4a. In the infinite dimension limit $D \to \infty$ the effective regularization is controlled by the bandwidth of the kernel resulting in an optimum $\omega^*$ which is plotted in Supplementary Figure 4b.

Supplementary Figure 5 displays kernel regression on a target function:

$$\bar{f}(\mathbf{x}) = \sum_{i=1}^{P'} \alpha_i K(\mathbf{x}, \bar{\mathbf{x}}_i), \quad \alpha_i \sim \mathcal{B}(1/2), \quad \bar{\mathbf{x}}_i \sim \mathcal{N}(0, \sigma^2 \mathbf{I}), \tag{111}$$

where $K$ is the Gaussian kernel with variance $\omega^2$ and $\alpha_i$ are drawn from a Bernoulli distribution. Generating $P$ noisy labels from this function, we perform kernel regression and calculate generalization error on a randomly generated test data. We repeat this process many times to obtain training and target dataset averaged generalization error (see Supplementary Note 5 for simulation details). Kernel regression experiment fits the theory prediction almost perfectly as can be seen from Supplementary Figure 5.

## Dot-Product Kernels and Neural Tangent Kernel

Here we consider the application of the generalization error Supplementary Equation 53 on dot-product kernels $K(\mathbf{x} \cdot \mathbf{x}') : \mathcal{S}^D \times \mathcal{S}^D \to \mathbb{R}$. Natural orthonormal basis on the input space $\mathcal{S}^D$ are $D$-dimensional hyper-spherical harmonics $\phi_\rho(\mathbf{x}) \equiv Y_{lm}(\mathbf{x})$ with $l = 0, 1, 2, 3, \dots$ and $m = 1, .., N(D, l)$ where $N(D, l) = \binom{n + D - 1}{n} - \binom{n + D - 3}{n - 2}$ is the number of degenerate modes associated to each mode $l$.

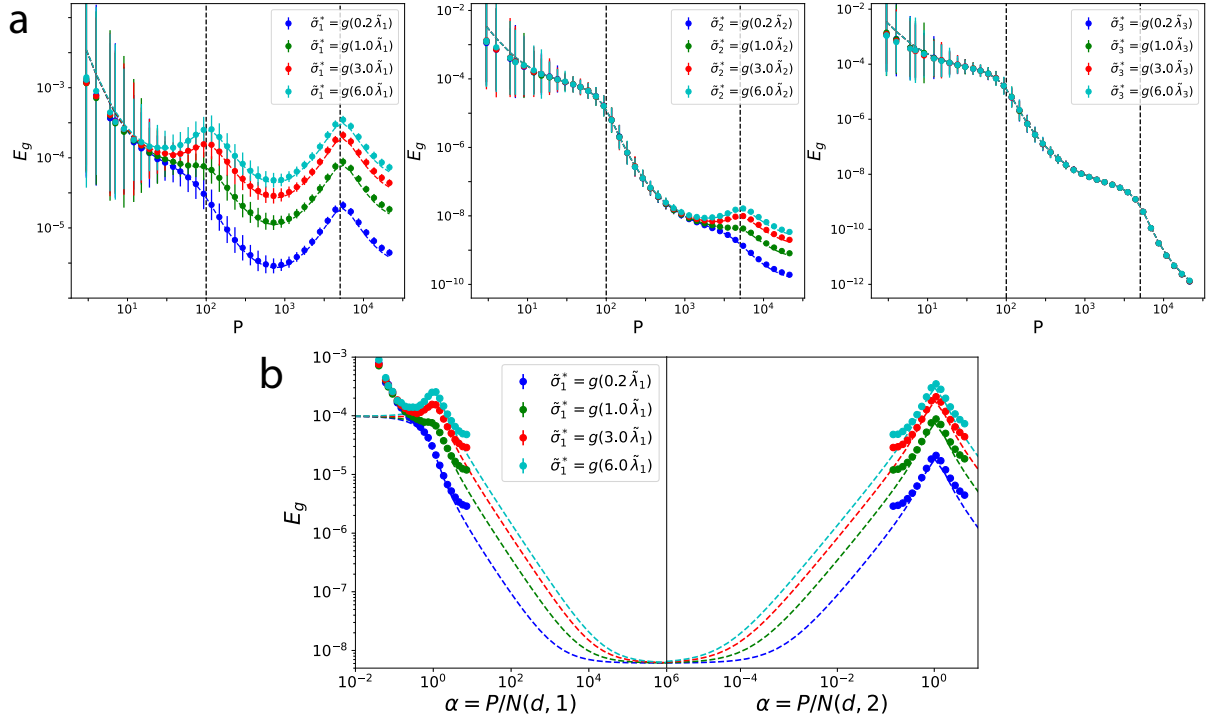

Supplementary Figure 5: Kernel regression with Gaussian RBF kernel with $\omega^2 = 900$, $\sigma = 1$ and $D = 100$ where the target was chosen as described in the Supplementary Note 5. **a.** Shows the theory curves generated by the finite $P$ generalization error formula. Vertical dashed lines indicate different degeneracies $N(D, K)$ corresponding to learning stages $K = 1, 2$. Different panels correspond to noise levels chosen based on $\tilde{\lambda}_L$ of mode $L$. Error bars indicate the standard deviation over 100 trials. **b.** Same experiment (left panel in **a**) for varying $\tilde{\sigma}_1$ compared to the $P, D \to \infty$ version of the generalization error formula. Around $P \sim N(D, L)$ for each learning stage, $E_g$ obtained above still predicts very well except in the middle regions, finite $P, D$ effects dominate. Error bars are removed for visualization.

We consider kernel ridgeless regression ($\lambda = 0$) with a kernel with power law spectrum $\bar{\eta}_k = k^{-s}$. In this case, effective regularization $\tilde{\lambda}_l$ increases with mode $l$ since $\tilde{\lambda}_l \approx l/(s-1)$. With a similar procedure applied in Gaussian RBF example, we set target weights $\bar{w}_k^2 = \eta_k = \bar{\eta}_k/N(D,k)$. A kernel regression experiment and prediction are shown in Supplementary Figure 6a. Note that we use the finite $P$ version of the generalization error to produce this plot meaning that the theory is still perfectly predictive without taking the infinite $P$ limit. In each panel, the label noise variance chosen according to $g(\tilde{\lambda}_k)$ corresponding to the learning stage $k = 1, 2, 3$. When $\tilde{\sigma}_k^2 < g(\tilde{\lambda}_k)$, generalization error falls monotonically at the learning stage $k$ and when $\sigma^2 > g(\tilde{\lambda}_k)$ we observe double-descent. Supplementary Figure 6b, on the other hand, demonstrates how infinite $P$ and $D$ limit of $E_g$, Eq. (9) in main text, predicts the regression experiments at each learning stage (corresponding to different panels) when $P$ is finite. We observe that a later learning stage starts before the previous learning stage saturates to its asymptotic value.

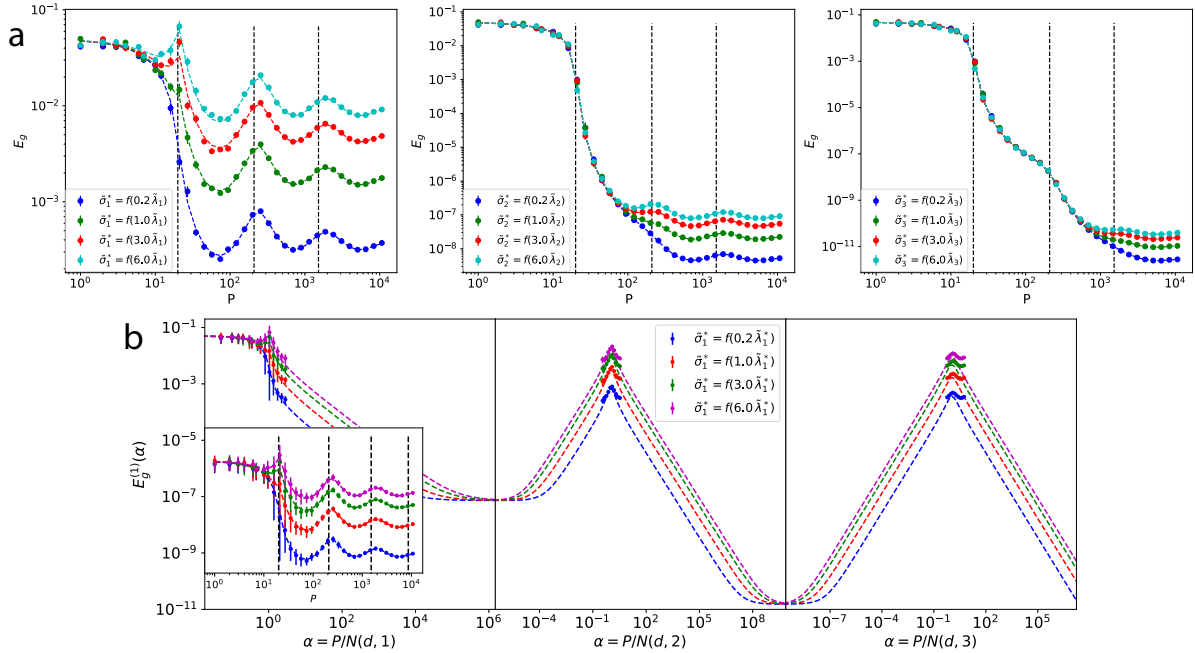

Supplementary Figure 6: Kernel ridgeless regression with power law kernel $\bar{\eta}_k = k^{-8}$ and $D = 20$ where the target was chosen as described in the Supplementary Note 5. **a.** Shows the theory curves generated by the generalization error formula given in Eq. (4) in main text evaluated at finite $P$. Vertical dashed lines indicate different degeneracies $N(D,l)$ corresponding to learning stage $l$. Different panels correspond to noise levels chosen based on $\tilde{\lambda}_l$ of mode $l$. Error bars indicate the standard deviation over 30 trials. **b.** Same experiment for varying $\tilde{\sigma}_1$ compared to the $P, D \to \infty$ version of the generalization error formula, Eq. (9) in main text. $E_g$ still predicts experiments very well around $P \sim N(D,l)$ for each learning stage. Finite $P, D$ effects dominate beyond. Error bars are removed for visualization.

The relevance of power law spectra and dot-product kernels to deep neural networks comes from the correspondence of infinitely wide neural networks and ridgeless kernel regression [17]. Consider a neural network with $L$ hidden layers where $n_\ell = n$ units in each of these layers for $\ell = 1, \ldots, L$ and $n_0 = D$ being the input dimension. We initialize the weights in each layer randomly $W_{ij}^{(\ell)} \sim \mathcal{N}(0, \sigma_W^2)$. With this parameterization and the concatenation of all network parameters in

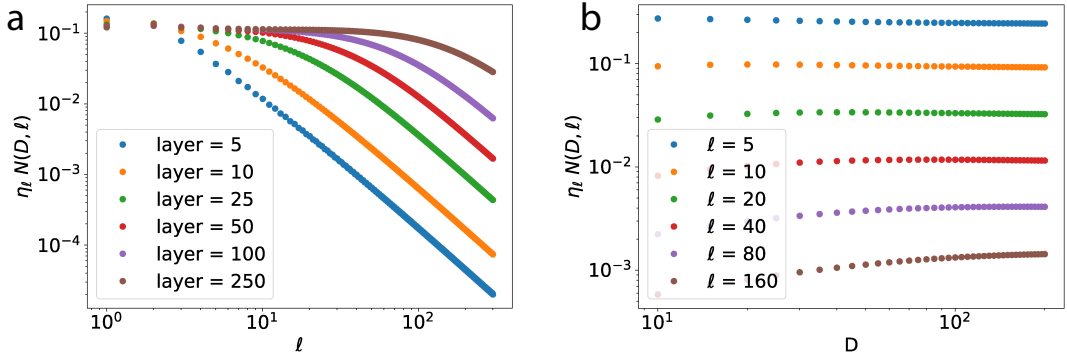

Supplementary Figure 7: Spectrum dependence of NTK to number of layers and input dimension. **a.** Empirically, spectrum $\bar{\eta}_l = \eta_l N(D, l)$ becomes white as more layers added. **b.** Furthermore, we confirm that the spectrum $\bar{\eta}_l$ is independent of input dimension for large $D$.

$\boldsymbol{\theta} = \text{Vec}\{\mathbf{W}^{(\ell)}\}_{\ell=1}^{L+1}$, the network function is:

$$f(\mathbf{x}; \boldsymbol{\theta}) = n^{-1/2} \mathbf{W}^{(L+1)} \sigma \left( n^{-1/2} \mathbf{W}^{(L)} \sigma \left( \dots n^{-1/2} \mathbf{W}^{(2)} \sigma \left( n_0^{-1/2} \mathbf{W}^{(1)} \mathbf{x} \right) \right) \right), \quad (112)$$

where $\sigma$ is a non-linearity. We will only consider the Rectified Linear Unit (ReLU). Training the network parameters $\boldsymbol{\theta}$ with gradient flow on a squared loss to zero training error is equivalent to the function obtained from ridgeless kernel regression with the Neural Tangent Kernel (NTK) [17, 18, 19]. This kernel can be obtained heuristically by linearizing the neural network function $f(\mathbf{x}, \boldsymbol{\theta})$ around its initial set of parameters $\boldsymbol{\theta}_0$, $f(\mathbf{x}, \boldsymbol{\theta}) \approx f(\mathbf{x}, \boldsymbol{\theta}_0) + \nabla_{\boldsymbol{\theta}} f(\mathbf{x}, \boldsymbol{\theta}_0) \cdot (\boldsymbol{\theta} - \boldsymbol{\theta}_0)$. Optimizing a mean squared regression error over $\boldsymbol{\theta}$ is equivalent to solving a linear regression problem for $\boldsymbol{\theta}$ where the feature Gram matrix is formed from initial parameter gradients: $\mathbf{K}_{\text{NTK},ij} = \nabla_{\boldsymbol{\theta}} f(\mathbf{x}_i, \boldsymbol{\theta}_0) \cdot \nabla_{\boldsymbol{\theta}} f(\mathbf{x}_j, \boldsymbol{\theta}_0)$. In the infinite-width limit, this quantity converges to its average over all possible initializations $\boldsymbol{\theta}_0$, giving rise to the deterministic NTK [17]. As an example, the exact form of NTK for ReLU non-linearity and zero bias is given by:

$K_{\text{NTK}}^{(0)}(\mathbf{x}, \mathbf{x}') = \cos^{-1}(\mathbf{x} \cdot \mathbf{x}')$

$K_{\text{NTK}}^{(1)}(\mathbf{x}, \mathbf{x}') = \cos\left[ f\left( \cos^{-1}(\mathbf{x} \cdot \mathbf{x}') \right) \right] + K_{\text{NTK}}^{(0)}(\mathbf{x}, \mathbf{x}') \left( 1 - \frac{\cos^{-1}(\mathbf{x} \cdot \mathbf{x}')}{\pi} \right)$

$K_{\text{NTK}}^{(2)}(\mathbf{x}, \mathbf{x}') = \cos\left[ f\left( f\left( \cos^{-1}(\mathbf{x} \cdot \mathbf{x}') \right) \right) \right] + K_{\text{NTK}}^{(1)}(\mathbf{x}, \mathbf{x}') \left( 1 - \frac{f\left( \cos^{-1}(\mathbf{x} \cdot \mathbf{x}') \right)}{\pi} \right)$

$\quad \dots$

$$K_{\text{NTK}}^{(L)}(\mathbf{x}, \mathbf{x}') = \cos\Big[ \underbrace{f(f(f(\dots f(\cos^{-1}(\mathbf{x} \cdot \mathbf{x}')))))}_{L \text{ times}} \Big] + K_{\text{NTK}}^{(L-1)}(\mathbf{x}, \mathbf{x}') \left( 1 - \frac{\overbrace{f(f(f(\dots f(\cos^{-1}(\mathbf{x} \cdot \mathbf{x}')))))}^{L-1 \text{ times}}}{\pi} \right),$$

$$(113)$$

where $f(\theta) = \cos^{-1} \left[ \frac{1}{\pi} \left( \sin(\theta) + (\pi - \theta) \cos(\theta) \right) \right]$. By projecting this function onto the Gegenbauer polynomials, we can obtain the spectrum of NTK for any layer [12]. We empirically observe that the eigenvalues obey power-law for large modes as seen from Supplementary Figure 7.

Having obtained the kernel and its spectrum, we perform kernel regression with the exact infinite-width limit NTK and train the corresponding finite width neural network. In Supplementary

Figure 8a, we demonstrate the results for fitting a pure mode target function $\bar{f}(\mathbf{x}) = a_k Q_k^{(D-1)}(\boldsymbol{\beta} \cdot \mathbf{x})$ which has vanishing weights except for a single mode $k$. $\boldsymbol{\beta}$ is randomly generated. We find that our theory describes NTK regression perfectly while neural network experiments show deviation from the theory at large $P$, possibly due to finite size effects. Indeed, increasing the width leads to a better match between experimental neural network risk and our theory, as shown in Supplementary Figure 8b.

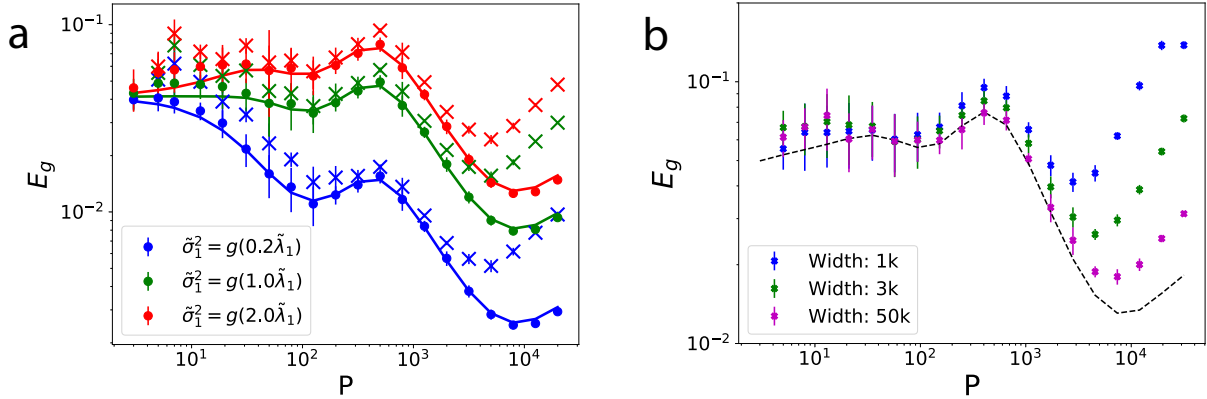

Supplementary Figure 8: **a.** 2-layer NTK regression and corresponding neural network training with 50000 hidden units for $D = 25$ with varying noise levels. Target function is the same as Fig. 6 in main text and is explained in detail in Supplementary Note 5. Solid lines are the theory predicted learning curves, dots represent NTK regression and × represents $E_g$ after neural network training. **b.** Generalization error for 2-layer NN with varying hidden units. We observe that increasing the width brings the learning curve closer to the NTK regression theory (dashed lines). Error bars represent standard deviation for kernel regression and neural network experiment over 15 and 5 trials, respectively.

We further observe a discrepancy between NTK regression and neural network training at low-$P$ where the generalization error is systematically biased towards higher values than its kernel regression counterpart. While this effect requires a further study of the correspondence between finite width neural networks and NTK regression, and remains to be fully understood, we empirically observe that the method of neural network ensembling [20, 16, 9] removes this discrepancy (Supplementary Figure 9b). Neural network ensembling is known to reduce the variance due to random parameter initializations [16] and hence also prevents the high-$P$ mismatch we observe in Fig. 6 in main text and Supplementary Figure 9a.

## Approximate Scaling of Learning Curves

A coarse approximation to our derived learning curves can be deduced provided that the kernel eigenspectrum obeys a certain property. The theoretical mode errors equations reveal that mode errors decay at different rates which are set by the kernel eigenvalues $\{\eta_\rho\}$. We exploit this fact to identify the number of eigenmodes that have been estimated once $P$ samples are taken. This is equivalent to identifying the mode $\rho^*(P)$ which satisfies

$$\kappa(P) \approx P\eta_{\rho^*(P)} \tag{114}$$

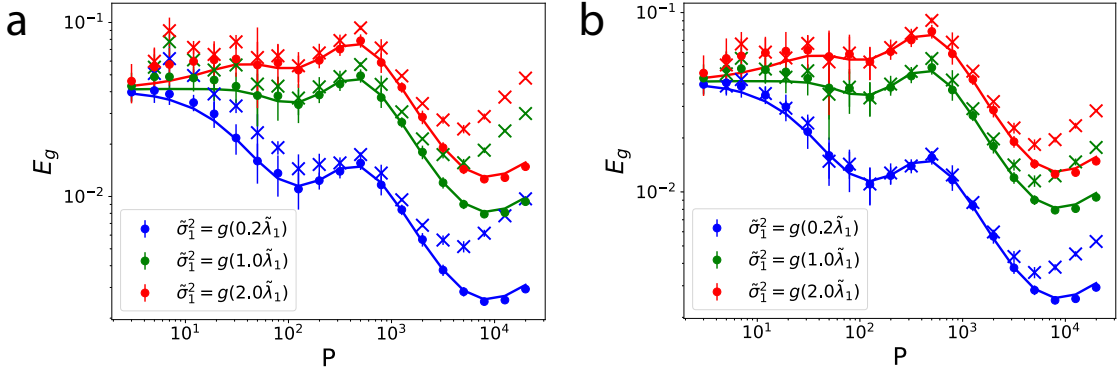

Supplementary Figure 9: **a.** 2-layer NTK regression and corresponding neural network training with 4000 hidden units for $D = 25$ with varying noise levels. Target function is the same as Fig. 6 in main text and is explained in detail in Supplementary Note 5. Solid lines are the theory predicted learning curves, dots represent NTK regression and × represents $E_g$ after neural network training. Notice the mismatch between regression and neural network experiment for low $P$ regime. **b.** Same experiment with 10 times ensembling. We observe that neural network ensembling reduces the mismatch for both low and high $P$ regimes. Error bars represent standard deviation for kernel regression and neural network experiment over 15 and 5 trials, respectively.

If $\kappa \gg P\eta_\rho$ then mode $\rho$ is not yet being learned while $P\eta_\rho \gg \kappa$ implies that the mode $\rho$ has already been estimated accurately. The self-consistency condition for $\kappa$ in the $\lambda \to 0$ limit is

$$1 = \sum_\rho \frac{\eta_\rho}{\eta_\rho P + \kappa} \approx \frac{1}{P} \sum_{\rho < \rho^*} \frac{\eta_\rho}{\eta_\rho} + \frac{1}{\kappa} \sum_{\rho > \rho^*} \eta_\rho = \frac{\rho^*(P)}{P} + \frac{1}{\kappa} \sum_{\rho > \rho^*} \eta_\rho. \tag{115}$$

Using the assumption that $\kappa(P) \approx P\eta_{\rho^*(P)}$, we find

$$P = \rho^* + \frac{1}{\eta_{\rho^*}} \sum_{\rho > \rho^*} \eta_\rho. \tag{116}$$

The solution $\rho^* \sim P$ is self-consistent provided that the effective regularization grows sublinearly

$$\tilde{\lambda}_\rho = \frac{1}{\eta_\rho} \sum_{\rho' > \rho} \eta_{\rho'} \sim \mathcal{O}(\rho). \tag{117}$$

This condition is not guaranteed to be satisfied, but if it is, then the scaling $\rho^*(P) \sim P$ is valid and we can approximate the remaining error with the power in all modes $\rho > \rho^* \approx P$

$$E_g(P) \sim \sum_{\rho > p} \eta_\rho \overline{w}_\rho^2, \tag{118}$$

which indicates that, at $P$, samples the expected generalization error can be approximated by a tail sum of the power in the target function. Examples of spectral decay rates that satisfy the sufficient condition above include power law $\eta_\rho \sim \rho^{-b}$ and exponential decays $\eta_\rho \sim u^\rho$, where $b$ and $u$ are constants. In Supplementary Figure 10, we compare this coarse grained theory to the MNIST experiments performed and reported in Fig. 1e in main text.

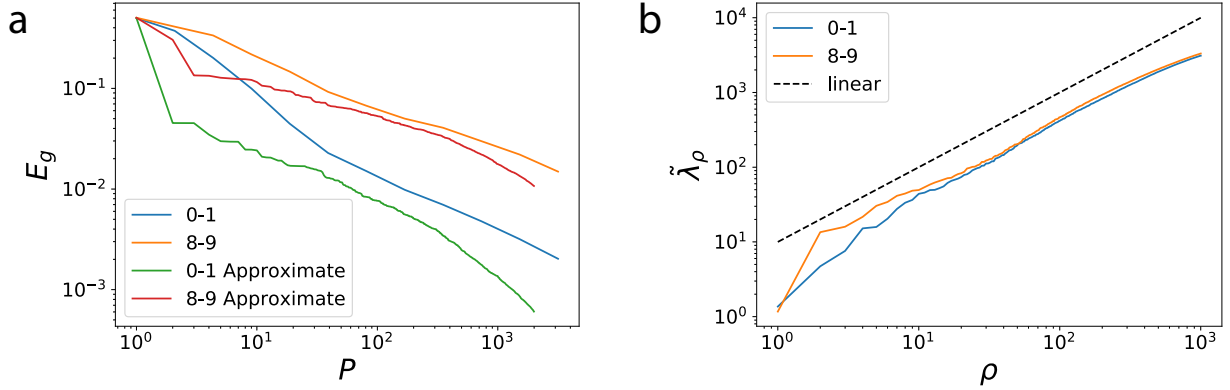

Supplementary Figure 10: **a.** Approximate Learning Curves for the Easy-Hard tasks in Fig. 1 in main text using tail sums. **b.** The effective regularization is only weakly sublinear in $\rho$.

# Supplementary Note 5 - Details of Numerical Experiments

## Calculating Kernel Spectrum for Spherical Data

For a chosen kernel, one can find the eigendecomposition onto spherical harmonics $Y_{kl}$ and Gegenbauer polynomials $Q_k^{(\alpha)}$ [21] using the following formula:

$$K(\mathbf{x}, \mathbf{y}) = \sum_{k=0}^{\infty} \eta_k \sum_{m=1}^{N(D,k)} Y_{k,m}(\mathbf{x}) Y_{k,m}(\mathbf{y}) = \sum_k \eta_k \frac{k+\alpha}{\alpha} Q_k^{(\alpha)}(\mathbf{x} \cdot \mathbf{y}). \tag{119}$$

Here $\alpha = \frac{D-2}{2}$. Projecting the kernel on the Gegenbauer basis, we obtain the eigenvalues:

$$\int_{-1}^{1} dt \, (1-t^2)^{\alpha - \frac{1}{2}} K(t) Q_k^{(\alpha)}(t) = \sum_l \eta_l \frac{l+\alpha}{\alpha} \int_{-1}^{1} dt \, (1-t^2)^{\alpha - \frac{1}{2}} Q_l^{(\alpha)}(t) Q_k^{(\alpha)}(t)$$
$$= \eta_k N(D,k) \frac{\omega_D}{\omega_{D-1}} \frac{\alpha}{k+\alpha}, \tag{120}$$

where $\omega_D = \frac{2\pi^{D/2}}{\Gamma(D/2)}$ is the surface area of a unit $D$-sphere and $N(D,k)$ is the degeneracy (see Supplementary Note 6). Hence, the eigenvalues of the kernel are found by the integral:

$$\eta_k N(D,k) = \frac{\omega_{D-1}}{\omega_D} \frac{k+\alpha}{\alpha} \int_{-1}^{1} dt \, (1-t^2)^{\alpha - \frac{1}{2}} K(t) Q_k^{(\alpha)}(t) \tag{121}$$

We compute the eigenvalues $\eta_l$ of a kernel $K(t)$ by numerically performing the integral in Supplementary Equation 121 with Gauss-Gegenbauer quadrature rule [22] for the measure $(1-t^2)^{\alpha-1/2}$. This formula is used to find the eigenspectrum of NTK in Supplementary Figure 7 and calculate the NTK learning curves in Fig. 6 in main text and in Supplementary Figure 8.

## Synthetic Data Experiments on Unit Sphere

We generally consider a scalar target functions of the form:

$$\bar{f}(\mathbf{x}) = \sum_k a_k Q_k^{(\alpha)}(\boldsymbol{\omega}_k \cdot \mathbf{x}), \quad \boldsymbol{\omega}_k, \mathbf{x} \in \mathbb{S}^{D-1}, \tag{122}$$

where $\boldsymbol{\omega}_k$ is a random projection vector on $\mathbb{S}^{D-1}$ and $a_k$ are coefficients. Using the identity $Q_k^{(\alpha)}(\boldsymbol{\omega} \cdot \mathbf{x}) = \frac{\alpha}{k+\alpha} \sum_{m=1}^{N(D,k)} Y_{km}(\boldsymbol{\omega}) Y_{km}(\mathbf{x})$ [21], we can expand this target function in terms of the features $\psi_{km}(\mathbf{x}) = \sqrt{\eta_k} Y_{km}(\mathbf{x})$:

$$\bar{f}(\mathbf{x}) = \sum_k a_k Q_k^{(\alpha)}(\boldsymbol{\omega}_k \cdot \mathbf{x}) = \sum_k \sum_{m=1}^{N(D,k)} a_k \frac{\alpha}{k+\alpha} Y_{km}(\boldsymbol{\omega}_k) Y_{km}(\mathbf{x}), \quad \boldsymbol{\omega}, \mathbf{x} \in \mathbb{S}^{D-1}$$

$$\equiv \sum_k \sum_{m=1}^{N(D,k)} \bar{w}_{km} \psi_{km}(\mathbf{x}), \quad \bar{w}_{km} = \frac{a_k}{\sqrt{\eta_k}} \frac{\alpha}{k+\alpha} Y_{km}(\boldsymbol{\omega}_k) \tag{123}$$

Therefore, we obtain:

$$\bar{w}_k^2 \equiv \frac{1}{N(D,k)} \sum_m \bar{w}_{km}^2 = \frac{a_k^2}{\eta_k} \frac{\alpha^2}{(k+\alpha)^2} \tag{124}$$

where we used $Q_k^{(\alpha)}(1) = \frac{\alpha}{k+\alpha} N(D,k)$ [21]. Having obtained the weights, we can find the generalization error using Supplementary Equation 53:

$$E_g = \frac{\kappa^2}{1-\gamma} \sum_\rho \frac{\eta_\rho \bar{w}_\rho^2}{(\kappa + P \eta_\rho)^2} + \sigma^2 \frac{\gamma}{1-\gamma} = \boxed{\frac{\kappa^2}{1-\gamma} \sum_k N(D,k) \frac{\eta_k \bar{w}_k^2}{(\kappa + P \eta_k)^2} + \sigma^2 \frac{\gamma}{1-\gamma}}, \tag{125}$$

where $\kappa = \lambda + \kappa \sum_k N(D,k) \frac{\eta_k}{\kappa + P \eta_k}$ to be solved self-consistently and $\gamma = \sum_k N(D,k) \frac{P \eta_k^2}{(\kappa + P \eta_k)^2}$. Hence this formula allows us to create a wide range of target functions and calculate the theoretical generalization error for a regression task on it. In our experiments, we consider two types of target functions as detailed below.

## Pure Target Functions

We consider target functions of the form $\bar{f}(\mathbf{x}) = a_k Q_k^{(\alpha)}(\boldsymbol{\omega} \cdot \mathbf{x})$ where $k$ is an integer. These functions are expressed with a single mode and simpler to evaluate numerically. Throughout our experiments, we picked $a_k = \eta_k \frac{k+\alpha}{\alpha}$ so that $\bar{w}_k^2 = \eta_k$. Hence the generalization error is simply:

$$E_g = \frac{\kappa^2}{1-\gamma} N(D,k) \frac{\eta_k^2}{(\kappa + P \eta_k)^2} + \sigma^2 \frac{\gamma}{1-\gamma}. \tag{126}$$

For synthetic neural network experiments in Fig. 6 in main text and Supplementary Figure 8, we use this type of target functions with $k = 1$ (hence linear target function). The code referenced in Methods section can be used for reproducing the results in this paper as well as for experimenting with different target functions including mixed target functions.

## Target Function with Non-zero Weights and Synthetic Kernel Regression Experiments

Generating a target function with all weights being non-zero with the method described above may be computationally expensive because sampling from many high dimensional spherical harmonics is not efficient. Instead, we devise the following method:

- We generate the target function using the representer's theorem. We choose $P'$ target examples $\{\bar{\mathbf{x}}^\mu\}$ (different than the training set) on the sphere and observe that:

$$\bar{f}(\mathbf{x}) = \sum_{\mu=1}^{P'} \bar{\alpha}_\mu K(\mathbf{x}, \bar{\mathbf{x}}^\mu) = \sum_{\rho=1}^{N} \left( \sum_{\mu=1}^{P'} \bar{\alpha}_\mu \sqrt{\eta_\rho} \phi_\rho(\bar{\mathbf{x}}^\mu) \right) \sqrt{\eta_\rho} \phi_\rho(\mathbf{x}) \equiv \sum_{\rho=1}^{N} \bar{w}_\rho \psi_\rho(\mathbf{x}) \qquad (127)$$

We note that $\bar{w}_\rho$ are random variables. To calculate their statistics, suppose we draw $\bar{\alpha}_\mu$ for each example i.i.d. from a distribution with mean 0 and variance $1/P'$. Then averaging over many $\{\mathbf{x}^\mu\}$ with large $P'$, we get the mean and variance of $\bar{w}_\rho$ to be:

$$\langle \bar{w}_\rho \rangle = 0, \quad \langle \bar{w}_\rho \bar{w}_\gamma \rangle = \eta_\rho \delta_{\rho\gamma}. \qquad (128)$$

For large $P'$, $\left| \bar{w}_\rho \right|^2$ concentrates around $\eta_\rho$, which we use in our theoretical calculations.

We also allow sample corruption by a Gaussian noise:

$$y^\mu = \sum_{\mu=1}^{P} \bar{\alpha}^\mu K(\mathbf{x}, \bar{\mathbf{x}}^\mu) + \epsilon^\mu, \qquad (129)$$

where noise for each sample has variance $\langle \epsilon^\mu \epsilon^\nu \rangle = \sigma^2 \delta^{\mu\nu}$.

- To solve the kernel regression problem, we again use the representer's theorem. Given $P$ training samples, $\{\mathbf{x}^\mu\}$, the solution is of the form:

$$f(\mathbf{x}) = \sum_{\mu=1}^{P} \alpha_\mu K(\mathbf{x}, \mathbf{x}^\mu). \qquad (130)$$

Plugging this into the kernel regression problem, with samples $y^\mu = \bar{f}(\mathbf{x}^\mu) + \epsilon^\mu$ generated by the target, we obtain the coefficients:

$$\min_{\boldsymbol{\alpha}} \left( \frac{1}{2} (\mathbf{y} - \mathbf{K}\boldsymbol{\alpha})^\top (\mathbf{y} - \mathbf{K}\boldsymbol{\alpha}) + \frac{\lambda}{2} \boldsymbol{\alpha}^\top \mathbf{K} \boldsymbol{\alpha} \right), \quad \implies \quad \boldsymbol{\alpha} = (\mathbf{K} + \lambda \mathbf{I})^{-1} \mathbf{y}. \qquad (131)$$

- Once we get these coefficients $\boldsymbol{\alpha}$, we can express the total generalization error as a sum of mode wise errors

$$\begin{aligned}
E_g &= \left\langle (f(x) - \bar{f}(x))^2 \right\rangle \\
&= \sum_{\rho\gamma} \eta_\rho \eta_\gamma \Big[ \sum_{j=1}^{P} \alpha_j \phi_\rho(x_j) - \sum_{i=1}^{P'} \overline{\alpha}_i \phi_\rho(\overline{x}_i) \Big] \Big[ \sum_{j=1}^{P} \alpha_j \phi_\gamma(x_j) - \sum_{i=1}^{P'} \overline{\alpha}_i \phi_\gamma(\overline{x}_i) \Big] \left\langle \phi_\rho(x) \phi_\gamma(x) \right\rangle \\
&= \sum_{\rho} \eta_\rho^2 \Big[ \sum_{j,j'} \alpha_j \alpha_{j'} \phi_\rho(x_j) \phi_\rho(x_j) - 2 \sum_{i,j} \alpha_j \overline{\alpha}_i \phi_\rho(x_j) \phi_\rho(\overline{x}_i) + \sum_{i,i'} \overline{\alpha}_i \overline{\alpha}_{i'} \phi(\overline{x}_i) \phi(\overline{x}_{i'}) \Big].
\end{aligned}$$

If we recognize that $\rho$ indexes both $(k, m)$ for the spherical harmonics, we can simplify the mode error to a simple matrix expression

$$\epsilon_k = \eta_k^2 \frac{k + \alpha}{\alpha} \Big[ \boldsymbol{\alpha}^\top Q_k^{(\alpha)} (\mathbf{X}^T \mathbf{X}) \boldsymbol{\alpha} - 2 \boldsymbol{\alpha}^\top Q_k^{(\alpha)} (\mathbf{X}^T \overline{\mathbf{X}}) \overline{\boldsymbol{\alpha}} + \overline{\boldsymbol{\alpha}}^\top Q_k^{(\alpha)} (\overline{\mathbf{X}}^T \overline{\mathbf{X}}) \overline{\boldsymbol{\alpha}} \Big]. \qquad (132)$$

We use this expression to compute experimental mode errors.

The theoretical generalization error can be obtained simply replacing $\bar{w}_\rho^2$ with the corresponding eigenvalue $\eta_\rho$ Supplementary Equation 128 in our expression Supplementary Equation 125. Gaussian RBF regression experiments are performed with the same method.

## Details of Neural Network Experiments

To perform neural network experiments, we use Neural Tangents package [23]. The label generating target functions are picked using the method for pure target functions (Supplementary Note 5). After we pick $P$ labelled training data, we generate a two-layer neural network using NTK initialization [17] with weight variance $\sigma_W^2 = 1$ and bias variance $\sigma_b^2 = 0$ with ReLU activation. We pick hidden layer widths varying between $4,000 - 50,000$. The network is trained with ADAM optimizer with learning rate 0.008. We perform 5 times averaging, and we sample different data points at each trial.

Corresponding NTK ridgeless regression experiments performed with zero regularization and the theory is calculated using the numerically calculated NTK spectrum and picking the weights to be equal to the spectrum.

## White Band-limited Kernel Experiments

The simplest kernel to experimentally probe the white bandlimited setting is the linear kernel $K(\mathbf{x}, \mathbf{x}') = \mathbf{x} \cdot \mathbf{x}'$ where $\mathbf{x} \sim \mathcal{N}(0, \mathbf{I})$. In this setting, the Mercer eigenfunctions are exactly the individual input vector elements $\phi_\rho(\mathbf{x}) = x_\rho$, $\rho = 1, ..., N$:

$$K(\mathbf{x}, \mathbf{x}') = \sum_{\rho=1}^{N} x_\rho x_\rho' \ , \ \int x_\rho K(\mathbf{x}, \mathbf{x}') p(\mathbf{x}) d\mathbf{x} = x_\rho' \tag{133}$$

We used this kernel and a linear target function $f_t(\mathbf{x}) = \boldsymbol{\beta} \cdot \mathbf{x}$ where $\boldsymbol{\beta} \sim \mathcal{N}(0, \mathbf{I})$ in the Fig. 3 experiments in main text. The number of features hence the input dimension is $N = 500$.

The one dimensional plot of the averaged estimator for uniform data on $S^1$ in Fig. 4 in main text used the following kernel

$$K(x, x') = 2 \sum_{k=1}^{N} \cos\big(k(x - x')\big) = 2 \sum_{k=1}^{N} \Big[ \cos(kx) \cos(kx') + \sin(kx) \sin(kx') \Big] \tag{134}$$

which is an RKHS with dimension $2N$ and the corresponding features are $\psi_k^{cos}(x) = \sqrt{2} \cos(kx)$ and $\psi_k^{sin}(x) = \sqrt{2} \cos(kx)$. In this experiment, the target function was chosen to be the (centered) vonMises function

$$\bar{f}(x) = e^{4(\cos(x)-1)} - I_0(4) \tag{135}$$

where $I_0$ is the Bessel function of order 0. The RKHS weights $\bar{a}_k$ were calculated by projecting the function on the features $\psi_k^{cos}(x)$ and $\psi_k^{sin}(x)$. To ensure the target is in the RKHS, we develop a bandlimited version

$$\bar{f}(x) = \sum_{k=1}^{N} \bar{a}_k \cos(kx). \tag{136}$$

Training data is generated using this target function by randomly choosing $x \in [-\pi, \pi]$ and the perform the kernel regression to obtain the estimator's RKHS weights. Finally, we evaluate the estimator on the interval $[-\pi, \pi]$ to compare the learned function to the target.

# Supplementary Note 6 - Notes on Spherical Harmonics

Here we collect some useful results on spherical harmonics. Details can be found in [21]. We are interested in finding a basis for the functions space on $\mathbb{S}^{D-1} \subset \mathbb{R}^D$. Let $\mathcal{P}_k^D$ to be the space of

homogeneous polynomials of degree $k$. Then its dimension is:

$$\dim \mathcal{P}_k^D = \binom{k + D - 1}{k} \tag{137}$$

Spherical harmonics are homogeneous $Y_{km}(t\mathbf{x}) = t^k Y_{km}(\mathbf{x})$, harmonic $\nabla^2 Y_{km}(\mathbf{x}) = 0$ polynomials, restricted to $\mathbb{S}^{D-1}$. They are orthonormal with respect to the uniform measure on the sphere

$$\int_{\mathbb{S}^{D-1}} Y_{km}(\mathbf{x}) Y_{k'm'}(\mathbf{x}) d\mathbf{x} = \delta_{k,k'} \delta_{m,m'} \tag{138}$$

The number of degree $k$ spherical harmonics in dimension $D$ is

$$N(D,k) = \binom{k + D - 1}{k} - \binom{k + D - 3}{k - 2} = \frac{2k + D - 2}{k} \binom{k + D - 3}{k - 1} \tag{139}$$

For large dimension $D \to \infty$ this number of degree $k$ harmonics grows like

$$N(D,k) \sim \frac{D^k}{k!}, D \to \infty \tag{140}$$

The Gegenbauer polynomial of degree $k$, $Q_k^{(\alpha)}$, can be related to all of the degree $k$ spherical harmonics

$$Q_k^{(\alpha)}(\mathbf{x} \cdot \mathbf{y}) = \frac{\alpha}{k + \alpha} \sum_{m=1}^{N(D,k)} Y_{km}(\mathbf{x}) Y_{km}(\mathbf{y}), \quad \mathbf{x}, \mathbf{y} \in \mathbb{R}^D, \tag{141}$$

where $\alpha = \frac{D-2}{2}$. We often refer to $Q_k^{(\alpha)}(\mathbf{x} \cdot \mathbf{y})$ as $Q_k^{(D-1)}(\mathbf{x} \cdot \mathbf{y})$ in the main text to emphasize that it is a polynomial defined on $\mathcal{S}^{D-1}$.

## Supplementary Discussion

### Relation to Kernel Alignment Risk Estimator (KARE)

A closely related study to our theory on the generalization of kernel methods utilizes random matrix theory to arrive at a related, but different, generalization prediction [4]. Their Kernel Alignment Risk Estimator (KARE) was shown to accurately predict generalization on the Higgs and MNIST datasets. The *signal capture threshold* (SCT) which arises in KARE theory is equivalent to the quantity we denote as $\kappa$, which can be interpreted as the resolvent of a generalized Wishart matrix (SI.3.1). The mode independent prefactor in our theory $\frac{1}{1-\gamma}$, which plays an important role in the KARE theory, can be obtained from differentiation $\kappa$ with respect to the ridge $\partial_\lambda \kappa$. Despite these similarities, our theory is not equivalent to KARE, which can be easily seen in the $\lambda \to 0$ limit, in which KARE scales with $P$ in a manner that does not depend on the kernel or task spectra whereas our learning curves still depend on $\eta_\rho$ and $\overline{w}_\rho^2$ even in the ridgeless limit. Our theory works well even in the $\lambda = 0$ simulations, for example in Fig. 3a and Fig. 4a in main text.

KARE may appear easier to implement than the theory we present here, since its only operations involve taking inverses, traces, and quadratic forms of kernel Gram matrices, whereas our theory involves a full eigendecomposition. In terms of time complexity, however, both methods scale as $O(M^3)$ for a sampled dataset of size $M$.

**Relation to Kernel Alignment**

In this paper, we introduced the idea of task-model alignment. A similar notion of kernel compatibility with target was discussed in [24, 25] (latter of which introduces algorithms to learn better aligned kernels), by introducing an "alignment" metric defined by:

$$A(K_1(x, x'), K_2(x, x')) = \frac{\langle K_1(x, x'), K_2(x, x') \rangle_p}{\sqrt{\langle K_1(x, x'), K_1(x, x') \rangle_p \langle K_2(x, x'), K_2(x, x') \rangle_p}}, \tag{142}$$

where $\langle k_1(x, x'), k_2(x, x') \rangle_p = \int dp(x) dp(x') k_1(x, x') k_2(x, x')$ denotes the inner product of two bi-dimensional functions with respect to the data distribution $p(x)$. The so-called "kernel-target alignment" between a kernel $K(x, x')$ and target function $\bar{f}(x)$ is then defined by $A(K(x, x'), \bar{f}(x)\bar{f}(x'))$.

To get further insight about the differences between the kernel alignment metric and our task-model alignment, we can evaluate the kernel alignment metric using the eigendecompositions of the kernel and the target function. We obtain:

$$A(K(\mathbf{x}, \mathbf{x}'), \bar{f}(\mathbf{x})\bar{f}(\mathbf{x}')) = \frac{1}{\sum_\rho \eta_\rho \bar{w}_\rho^2} \frac{\sum_\rho \eta_\rho^2 \bar{w}_\rho^2}{\sqrt{\sum_\rho \eta_\rho^2}}, \tag{143}$$

which is a scalar between $[0, 1]$. In contrast, the task-model alignment states that when the cumulative power $C(\rho) = \frac{\sum_{\rho' \le \rho} \eta_{\rho'} \bar{w}_{\rho'}^2}{\sum_{\rho'} \eta_{\rho'} \bar{w}_{\rho'}^2}$ at each mode $\rho$ for a target functions is entry-wise larger than the $C(\rho)$ for another target function, the target function with larger $C(\rho)$ is learned with less training data then the other. While the cumulative power contains information about each learning stage $\rho$, kernel alignment is an aggregate measure of how the kernel and task is aligned.

To study the similarities between our prediction and [24] in a simple context, let us consider the regression task with the kernel $K(x, x') = \bar{f}(x)\bar{f}(x')$ and target $\bar{f}(x)$. This kernel and the target are perfectly aligned under the metric of [24] (A=1). The kernel has only a single eigenfunction $\phi(x) = \bar{f}(x)/\sqrt{\langle \bar{f}^2(x) \rangle_p}$ and hence a single non-zero eigenvalue $\eta = \langle \bar{f}^2(x) \rangle_p$. Furthermore, the eigenspace expansion of the target function gives trivially $\bar{w} = 1$. Hence, all the target power is placed in the first eigenfunction. Our theory implies that the $\bar{f}$ is the target function whose generalization error falls fastest under this kernel ($C(1) = 1$). Interestingly, in this case the generalization error has the same form as the band-limited case.

**Relation to Other Statistical Physics Approaches to Kernel Machines**

In the statistical physics domain, the replica method has been used to calculate classification learning curves for support vector machines when the data distribution is spherically symmetric and high dimensional, and for a specific class of target functions [26, 27]. These works revealed a countably infinite number of consistent thermodynamic limits: $P, D \to \infty$ with $P = \alpha D^L$ for integer $L$, which arise due to the degeneracy of kernel eigenvalues for all spherical harmonics of degree $L$. In the $L$-th learning stage, polynomials of degree $L$ are being estimated as $\alpha$ increases. The rotation invariant kernels we discuss in this paper show the same learning stages, although for kernel regression rather than kernel SVM. However, our theory not only applies to spherically symmetric and infinite dimensional settings but is also shown to work for more general data distributions and a wide range of target functions, including realistic datasets such as MNIST and CIFAR. To aid the study of kernel methods in these more general distributions, we introduce useful metrics such as cumulative power $C(\rho)$ to quantify the alignment of the kernel with the learning task of interest.

The generalization of regularized linear regression with isotropic Gaussian features was analyzed with the replica method in [28] and diagrammatic methods in [29, 30, 31]. These works predicted learning curves equivalent to a special case of our kernel regression theory: the white band-limited spectrum. In this special case, we provide a phase diagram showing how effective noise and regularization can alter the non-monotonic behavior of the learning curve. We further show the universality of this learning curve and phase diagram to any setting where the kernel admits a truncated Mercer decomposition with all eigenvalues equal. We explicitly illustrate such an equivalence in the spherically symmetric data setting. In the $L$-th learning stage ($P \approx D^L$ with $D \to \infty$), any kernel essentially reduces to white band-limited kernel with $N(D, L)$ equal variance modes (spherical harmonics of degree $L$). We show that the higher degree components of the target act as effective noise while higher degree kernel eigenvalues act as effective regularization, allowing the phase diagram from the white band-limited setting to carry over to this interesting case.

Generalization error for Gaussian process regression, where the target function is a random field with covariance kernel $K_t$, was analyzed in the average case by Peter Sollich [32, 14]. Using the Sherman-Morrison inverse formula, he derived a continuous approximation of the learning curves from a partial differential equation for $\kappa$. In his work, he introduced a variety of approximations to learning curves based on this PDE approach. The learning curves obtained in the present work with the replica method agree with his *lower continuous approximation* in the kernel regression limit. Our theory goes beyond this work both in formalism and the results. Our framework allows the computation of other relevant observables (training error, average estimator, bias, and variance) by simply including additional sources in our partition function. We provide a detailed analysis of the generalization error expression to elucidate important heuristics for generalization and further show its applicability to real datasets.

Recent works on the effects of over-parameterization in random feature models have been investigated with the replica method [33, 16] and with random matrix theory tools [8, 34]. In the random features model, the last layer weights in a randomly initialized neural network are trained. In this setting the learning curves depend on the input dimension $D$, the number of random features $N$ (network width) and the number of data points $P$. Typically, the authors consider fitting noisy linear target functions with these networks. In this model, two types of overfitting peaks can occur, one when $P \approx D$ and one when $P \approx N$ [35]. The relation between these peaks and the ones we see are discussed in the main text.

The equivalence between training infinite-width neural networks and kernel methods, allowed Cohen et al. to study the generalization of wide neural networks with a Gaussian field theory [6]. They obtained a different expression for the theoretical generalization error using a Poisson averaging technique, allowing them to expand their theoretical prediction in powers of $1/P$ at finite $\lambda$. We show in Supplementary Equation 64 that their expression for $\langle f^*(\mathbf{x}; P) \rangle_\mathcal{D}$ can be obtained from ours by a series expansion in a certain limit, showing that our result contains corrections not captured in their theory. Further, to the perturbative order they consider, they find that $\langle f^*(\mathbf{x}; P)^2 \rangle_\mathcal{D} = \langle f^*(\mathbf{x}; P) \rangle_\mathcal{D}^2$, missing the variance of the estimator which plays a crucial role in our theory, especially for non-monotonicity. Furthermore, their expressions break down in the ridgeless limit and they resort to a renormalization approach to predict learning curves. Our expression is valid in this limit.

Our recent conference publication on kernel regression and the infinite-width limit of neural networks [12] laid the groundwork for the present investigation. However, the current paper significantly expands on it and brings in many new insights. First, the effect of label noise on generalization and the multiple descent phase diagram were not explored in our previous work which we provide an account of here. Second, we provide comparisons with CIFAR-10, not present in the previous work. Third, we introduced and analyzed the white band limited RKHS model. Fourth,

we emphasize task-model alignment as a heuristic here and provide a metric for it. Fifth, the field theory formalism presented in this paper is more general than the technique of [12], which only used the replica method to calculate the average of an inverse matrix. The flexibility of the new theory allowed us to compute other observables including training error, the expected estimator, and the covariance of the estimated coefficients over different datasets.

# Supplementary References



\end{document}